%% file: main.tex
\documentclass{book}
\usepackage{roboto}

\usepackage[english]{babel}
\usepackage[letterpaper,top=2cm,bottom=2cm,left=3cm,right=3cm,marginparwidth=1.75cm]{geometry}

\usepackage{pdflscape}
\usepackage{rotating}
\usepackage{longtable}
\usepackage{array}
\usepackage{float}
\usepackage{amsmath}
\usepackage{amssymb}
\usepackage{graphicx}
\usepackage{inconsolata}
\usepackage[colorlinks=true, allcolors=blue]{hyperref}
\usepackage{enumitem}
\usepackage{tikz} 
\usepackage{listings}
\usepackage{xcolor}
\usepackage[T1]{fontenc}
\usepackage{IEEEtrantools}  
\usepackage{epigraph}  

\definecolor{bgcolor}{rgb}{0.97,0.97,0.97}
\definecolor{codeblue}{rgb}{0.1,0.1,0.8}
\definecolor{codegreen}{rgb}{0,0.4,0}
\definecolor{codegray}{rgb}{0.4,0.4,0.4}
\definecolor{codepurple}{rgb}{0.5,0,0.5}
\definecolor{codered}{rgb}{0.6,0.2,0.2}
\definecolor{lightgray}{rgb}{0.9,0.9,0.9}
\definecolor{darkgray}{rgb}{0.6,0.6,0.6} 

\makeatletter
\renewcommand{\paragraph}{%
  \@startsection{paragraph}{4}{\z@}{1ex}{-1em}{\normalfont\normalsize\bfseries\color{gray}}}
\makeatother

\lstdefinestyle{python}{
    language=Python,
    basicstyle=\ttfamily\small\color{black}\usefont{T1}{zi4}{m}{n},  
    keywordstyle=\bfseries\color{codeblue},  
    stringstyle=\color{codegreen},  
    commentstyle=\slshape\color{codegray},  
    showstringspaces=false,
    numbers=left,
    numberstyle=\tiny\color{codegray},  
    stepnumber=1,
    numbersep=8pt,
    frame=single,
    rulecolor=\color{darkgray},  
    breaklines=true,
    backgroundcolor=\color{bgcolor},
    tabsize=4,
    captionpos=b,
    morekeywords={self}, 
}

\lstdefinestyle{cmd}{
    language=bash,
    basicstyle=\ttfamily\small\color{black}\usefont{T1}{zi4}{m}{n},  
    keywordstyle=\bfseries\color{blue},
    stringstyle=\color{codegreen},
    commentstyle=\itshape\color{gray},
    showstringspaces=false,
    numbers=none,
    frame=single,
    rulecolor=\color{darkgray},  
    breaklines=true,
    backgroundcolor=\color{bgcolor},
    tabsize=4,
    captionpos=b,
}

\title{Mastering AI: Big Data, Deep Learning, and the Evolution of Large Language Models - AutoML from Basics to State-of-the-Art Techniques
}

\author{
    Pohsun Feng\textsuperscript{*†} \\ 
    \textit{National Taiwan Normal University} \\
    41075018h@ntnu.edu.tw
    \and
    Ziqian Bi\textsuperscript{*†} \\ 
    \textit{Indiana University} \\
    bizi@iu.edu
    \and
    Yizhu Wen \\ 
    \textit{University of Hawaii} \\
    yizhuw@hawaii.edu
    \and
    Benji Peng \\ 
    \textit{AppCubic} \\
    benji@appcubic.com
    \and
    Junyu Liu \\ 
    \textit{Kyoto University} \\
    liu.junyu.82w@st.kyoto-u.ac.jp
    \and
    Caitlyn Heqi Yin \\ 
    \textit{University of Wisconsin-Madison} \\
    hyin66@wisc.edu
    \and
    Tianyang Wang \\ 
    \textit{Xi’an Jiaotong-Liverpool University} \\
    Tianyang.Wang21@student.xjtlu.edu.cn
    \and
    Keyu Chen \\ 
    \textit{Georgia Institute of Technology} \\
    kchen637@gatech.edu
    \and
    Sen Zhang \\ 
    \textit{Rutgers University} \\
    sen.z@rutgers.edu
    \and
    Ming Li \\ 
    \textit{Georgia Institute of Technology} \\
    mli694@gatech.edu
    \and
    Jiawei Xu \\ 
    \textit{Purdue University} \\
    xu1644@purdue.edu
    \and
    Ming Liu \\ 
    \textit{Purdue University} \\
    liu3183@purdue.edu
    \and
    Xuanhe Pan \\ 
    \textit{University of Wisconsin-Madison} \\
    xpan73@wisc.edu
    \and
    Jinlang Wang \\ 
    \textit{University of Wisconsin-Madison} \\
    jinlang.wang@wisc.edu
    \and
    Xinyuan Song \\ 
    \textit{Emory University} \\
    xsong30@emory.edu
    \and
    Qian Niu \\ 
    \textit{Kyoto University} \\
    niu.qian.f44@kyoto-u.ac.jp
}

\date{} 

\begin{document}

\maketitle

\begingroup
\renewcommand\thefootnote{}\footnote{
    \textsuperscript{*} Equal contribution \\
    \textsuperscript{$\dagger$} Corresponding author
}
\addtocounter{footnote}{0}
\endgroup

\epigraph{"The greatest enemy of knowledge is not ignorance, it is the illusion of knowledge."}{\textit{Daniel J. Boorstin}}

\tableofcontents  

\input{automl}  

\bibliographystyle{ieeetr}
\bibliography{sample}

\end{document}

%% file: automl.tex
\input{00_introduction}

\input{01_basic_python}
\input{02_basic_lib}

\input{03_ml}
\input{04_preprocessing}

\input{05_linear}
\input{06_svm}
\input{07_rf}
\input{08_boost}
\input{09_lasso}
\input{10_riskslim}

\input{11_gridsearchcv}
\input{12_automl}
\input{13_tpot}
\input{14_autogluon}
\input{15_other}

\input{16_cloudbase}

\input{17_dl}
\input{18_cnn}

\input{19_nas}
\input{20_automl_dl}

\input{20.01_supercomputer}

\input{21_future}

%% file: 00_introduction.tex
\chapter{Introduction to AutoML}

In recent years, Artificial Intelligence (AI) and Machine Learning (ML) have grown tremendously in popularity across various industries. From healthcare and finance to retail and automotive, adopting machine learning models has led to significant advancements~\cite{jordan2015machine}. However, building machine learning models traditionally requires deep knowledge in multiple areas, such as data preprocessing, feature engineering, model selection, hyperparameter tuning, and evaluation~\cite{kohavi1998feature}. For many beginners and even experienced practitioners, this process can be time-consuming and technically challenging.

This is where \textbf{AutoML} (Automated Machine Learning) comes in. AutoML simplifies the process of building machine learning models by automating many of the steps that would otherwise require manual intervention~\cite{he2021automl}. AutoML tools can automatically preprocess data, select the most suitable algorithms, and fine-tune hyperparameters to produce highly accurate models~\cite{feurer2015efficient}. This automation not only speeds up the model development cycle but also allows users without deep knowledge of machine learning to create models with comparable performance to those made by experienced data scientists.

\section{Why is AutoML Important?}

There are several reasons why AutoML has become an important trend in the world of AI and machine learning. For beginners and new learners, it’s crucial to understand the implications of AutoML, as its adoption is changing the landscape of how models are developed and deployed.

\subsection{Automation of Manual Processes}

Traditionally, the process of building a machine learning model involves multiple stages, including:

\begin{itemize}
    \item \textbf{Data Preprocessing:} Cleaning and transforming raw data into a format suitable for machine learning.
    \item \textbf{Feature Engineering:} Selecting or transforming input features that help the model perform better.
    \item \textbf{Model Selection:} Choosing the appropriate algorithm, such as decision trees, neural networks, or support vector machines.
    \item \textbf{Hyperparameter Tuning:} Finding the best settings (hyperparameters) for the chosen algorithm to optimize performance.
    \item \textbf{Model Evaluation:} Evaluating the model's performance using metrics such as accuracy, precision, recall, etc.
\end{itemize}

Each of these steps can take considerable time and effort, especially if you are unfamiliar with machine learning techniques. With AutoML, these steps can be automated to a large extent, significantly reducing the complexity of the process.

For example, if you were tasked with manually adjusting hyperparameters for a machine learning model, you might need to run multiple experiments to find the best combination. AutoML tools can do this automatically using techniques such as grid search or random search to explore different hyperparameter settings.

\subsection{Boosting Productivity and Efficiency}

With the automation of these processes, AutoML allows data scientists and machine learning engineers to focus on more critical tasks, such as understanding the business problem, interpreting the model results, and ensuring the ethical use of AI~\cite{he2021automl}. This increased productivity and efficiency can lead to faster deployment of models and, ultimately, more competitive advantages for organizations~\cite{zoller2021benchmark}.

For example, in the healthcare industry, AutoML is being used to build models that can automatically diagnose medical conditions from data, such as detecting cancerous cells in X-ray images~\cite{esteva2019guide}. Such models, when deployed, can assist doctors in making faster, more accurate diagnoses~\cite{topol2019high}. 

Another example is in financial services. AutoML is being used by banks to develop models that detect fraudulent transactions in real-time, allowing financial institutions to save millions of dollars~\cite{radovanovic2020fraud}. Such advances have already led to a demand for fewer manual fraud analysts, as machines are increasingly taking over these repetitive, pattern-based tasks~\cite{pawar2020fraud}.

\subsection{Lowering the Entry Barrier for Beginners}

AutoML makes machine learning more accessible to those without a strong background in the field. For beginners or newcomers to machine learning, it is now possible to build sophisticated models without needing to understand every intricate detail of the underlying algorithms~\cite{yao2018taking}. This is especially beneficial for professionals from non-technical backgrounds who want to leverage machine learning in their work~\cite{gonzalez2020helping}.

Let’s consider an example:

\begin{lstlisting}[style=python]
import torch
import torch.nn as nn
import torch.optim as optim
from sklearn.datasets import load_breast_cancer
from sklearn.model_selection import train_test_split
from sklearn.preprocessing import StandardScaler

# Load dataset
data = load_breast_cancer()
X_train, X_test, y_train, y_test = train_test_split(data.data, data.target, test_size=0.2, random_state=42)

# Scale data
scaler = StandardScaler()
X_train = scaler.fit_transform(X_train)
X_test = scaler.transform(X_test)

# Simple Neural Network Model using PyTorch
class SimpleNN(nn.Module):
    def __init__(self):
        super(SimpleNN, self).__init__()
        self.fc1 = nn.Linear(X_train.shape[1], 16)
        self.fc2 = nn.Linear(16, 8)
        self.fc3 = nn.Linear(8, 1)
    
    def forward(self, x):
        x = torch.relu(self.fc1(x))
        x = torch.relu(self.fc2(x))
        x = torch.sigmoid(self.fc3(x))
        return x

# Model, loss function, and optimizer
model = SimpleNN()
criterion = nn.BCELoss()
optimizer = optim.Adam(model.parameters(), lr=0.001)

# Example Training Loop
for epoch in range(100):
    optimizer.zero_grad()
    outputs = model(torch.FloatTensor(X_train))
    loss = criterion(outputs.squeeze(), torch.FloatTensor(y_train))
    loss.backward()
    optimizer.step()

print("Training complete.")
\end{lstlisting}

This is a simple neural network implemented using PyTorch. While it’s essential to understand how this code works, many aspects of this process (like choosing the optimizer, adjusting the learning rate, etc.) can be automated by AutoML frameworks. This reduces the learning curve for beginners while ensuring that the models they produce are still high-quality.

\subsection{The Threat of Automation: Job Displacement and the Changing Workforce}

One of the most critical aspects of AutoML is its potential impact on the job market. The automation of machine learning processes, while boosting efficiency, also raises concerns about job displacement. Industries that were once reliant on human workers for data analysis, model development, and manual feature engineering are increasingly turning to AutoML to streamline these processes.

\textbf{For example:}

\begin{itemize}
    \item In the \textbf{retail} sector, machine learning models are being used to predict customer behavior and optimize supply chains. As a result, the need for manual data entry, forecasting, and even managerial decision-making roles is decreasing. Automated models can process vast amounts of data faster and more accurately than humans, leading companies to reduce their workforce in these areas.
    
    \item The \textbf{automotive} industry is also seeing a shift. Companies are employing AI-powered systems to optimize production lines, perform predictive maintenance on machinery, and even automate quality control. Traditionally, these tasks required human expertise and supervision, but with AutoML and AI tools taking over, the demand for such roles is shrinking.
    
    \item In \textbf{marketing}, AutoML tools are being used to automate targeted ad campaigns. What once required entire teams to analyze customer data and predict trends can now be done with machine learning algorithms, minimizing the need for data analysts and marketing strategists.
    
    \item The \textbf{financial industry} is heavily investing in automated trading systems powered by machine learning. These systems can analyze market trends and make high-frequency trades at a speed that no human trader could achieve. As such, jobs traditionally filled by stock traders and analysts are increasingly at risk of being automated.
\end{itemize}

This automation trend signals a clear shift in the demand for skills in the job market. The repetitive and process-oriented tasks are becoming prime candidates for automation, meaning workers in these areas may face the threat of being replaced by machines. Even roles that require some decision-making are not immune to this trend, as AI and AutoML systems become more sophisticated.

\textbf{What Does This Mean for You?}

The rise of AutoML means that many traditional roles in industries such as manufacturing, finance, and retail could be drastically reduced or eliminated. Those who fail to adapt to the changing landscape may find themselves outpaced by technology. If you want to remain relevant in your industry, you must develop the skills necessary to work with these advanced tools. Understanding how to use AutoML and apply machine learning concepts will be essential to staying competitive in the job market.

\section{AutoML: A Complement, Not a Replacement}

While AutoML provides great power in automating tasks, it is important to remember that it doesn’t replace the need for human insight. Machine learning models are tools for solving business problems, and understanding the context of these problems is crucial for building effective models. AutoML can aid in the technical aspects, but human judgment is still needed to interpret the results, ensure fairness, and avoid potential biases in the model.

\section{Conclusion}

AutoML is revolutionizing the way we approach machine learning. Automating many of the complex and time-consuming tasks involved in building models, lowers the entry barrier for beginners and accelerates the workflow for experienced practitioners. However, AutoML should be seen as a tool that complements human expertise, not as a replacement for it. As AutoML becomes more prevalent, learning how to use these tools effectively will become a valuable skill in the data science industry.

Whether you're just starting your journey into machine learning or are an experienced professional, understanding the role of AutoML and staying updated on its developments will ensure that you stay competitive in this rapidly evolving field. However, the reality is clear: as machine learning tools and AutoML continue to advance, the job market is likely to become increasingly reliant on those who can work with, and not be replaced by, these technologies.

%% file: 01_basic_python.tex
\part{Fundamental Knowledge of Programming, Machine Learning and Deep Learning}

\chapter{Basic Python Syntax}

\section{Introduction to Python}

Python \cite{peng2024deeplearningmachinelearning, li2024deeplearningmachinelearning} is a high-level, interpreted programming language created by Guido van Rossum and first released in 1991 \cite{van1995python}. Its design philosophy emphasizes code readability, and its syntax allows programmers to express concepts in fewer lines of code than possible in languages like C++ or Java.

Python is dynamically typed and garbage-collected, and it supports multiple programming paradigms, including procedural, object-oriented, and functional programming. It is known for its large standard library, which provides tools suited for a wide range of tasks.

Python is widely used in fields such as web development, data science, artificial intelligence, automation, and cybersecurity. Some advantages of Python include:
\begin{itemize}
    \item Readability: Python's syntax is clean and easy to read.
    \item Versatility: Python can be used for small scripts as well as large systems.
    \item Extensive Libraries: Python has a wide range of libraries and frameworks.
\end{itemize}

\section{Install Python}
In this section, we will guide you step by step through the installation of Python and setting up a suitable development environment. This includes installing IDLE \cite{kaswan2023python}, PyCharm \cite{islam2015mastering}, VSCode \cite{del2019visual}, and Anaconda \cite{phelps2000anaconda}, and setting up a virtual environment. These tools and environments are widely used for Python programming and offer distinct features beneficial for beginners.

\subsection{Installing Python}
Python is an interpreted language, which means you need to have the Python interpreter installed on your system to run Python programs. The official website of Python is \url{https://www.python.org/}. Follow the steps below to install Python.

\subsubsection{Step-by-Step Guide to Install Python:}
\begin{enumerate}
  \item Go to \url{https://www.python.org/downloads/}.
  \item Download the latest version of Python for your operating system (Windows, Mac, or Linux).
  \item Run the installer.
  \item \textbf{Important:} Make sure to check the box \texttt{Add Python to PATH} during installation on Windows.
  \item Click \texttt{Install Now} and follow the on-screen instructions.
  \item After installation, you can verify it by opening a command prompt (or terminal on Mac/Linux) and typing the following command:
  
\begin{lstlisting}[style=cmd]
python --version
\end{lstlisting}
  This should display the installed Python version.
\end{enumerate}

Once Python is installed, you are ready to run Python scripts using IDLE or any other IDE (Integrated Development Environment) like PyCharm or VSCode.

\subsection{IDLE}
IDLE (Integrated Development and Learning Environment) is the default IDE that comes with Python. It is simple and great for beginners. Here's how you can use IDLE:

\subsubsection{Step-by-Step Guide to Open and Use IDLE:}
\begin{enumerate}
  \item After installing Python, search for \texttt{IDLE} in your operating system's search bar and open it.
  \item IDLE opens with a Python shell. You can write and execute Python commands directly here.
  \item To create a new script, go to \texttt{File} \(\rightarrow\) \texttt{New File}.
  \item In the new window, you can write your Python code. For example, try this simple script:
  
\begin{lstlisting}[style=python]
print("Hello, Python world!")
\end{lstlisting}
  \item Save the file with a \texttt{.py} extension, and then run it by clicking on \texttt{Run} \(\rightarrow\) \texttt{Run Module} or pressing \texttt{F5}.
\end{enumerate}

IDLE is a great tool for small projects and experimenting with Python, but for larger projects, more advanced IDEs like PyCharm or VSCode are recommended.

\subsection{PyCharm}
PyCharm is a popular Python IDE that offers advanced features such as code completion, debugging, and project management. PyCharm has a free Community Edition, which is perfect for beginners.

\subsubsection{Step-by-Step Guide to Install PyCharm:}
\begin{enumerate}
  \item Go to \url{https://www.jetbrains.com/pycharm/download/}.
  \item Download the \texttt{Community Edition} (the free version).
  \item Follow the installer instructions.
  \item Once installed, open PyCharm and create a new project by selecting \texttt{New Project}.
  \item In the project settings, make sure to select the Python interpreter installed earlier.
  \item After setting up the project, you can create a new Python file by right-clicking on the project folder and selecting \texttt{New} \(\rightarrow\) \texttt{Python File}.
\end{enumerate}

For example, you can write the following simple script in PyCharm:

\begin{lstlisting}[style=python]
for i in range(5):
    print(f"Iteration {i}")
\end{lstlisting}

You can run this by clicking the green \texttt{Run} button.

\subsection{Visual Studio Code (VSCode)}
VSCode is a lightweight code editor developed by Microsoft. It supports many programming languages, including Python, and offers extensions to enhance functionality.

\subsubsection{Step-by-Step Guide to Install VSCode for Python:}
\begin{enumerate}
  \item Go to \url{https://code.visualstudio.com/} and download the installer for your OS.
  \item Install VSCode following the installation instructions.
  \item After installation, open VSCode.
  \item Install the Python extension by Microsoft by going to \texttt{Extensions} (left sidebar) and searching for \texttt{Python}.
  \item After installation, open a folder as a workspace and create a new Python file with a \texttt{.py} extension.
  \item Make sure to select the Python interpreter by pressing \texttt{Ctrl + Shift + P} and typing \texttt{Python: Select Interpreter}.
  \item You can now write and execute Python code within VSCode. For example:

\begin{lstlisting}[style=python]
x = 10
y = 20
print(x + y)
\end{lstlisting}

  \item To run the code, press \texttt{Ctrl + F5}.
\end{enumerate}

VSCode is highly customizable and can be extended with various plugins, making it a great tool for both beginners and advanced users.

\subsection{Anaconda}
Anaconda is a distribution of Python and R programming languages for data science and machine learning. It comes with many useful libraries pre-installed and includes Jupyter Notebooks for interactive data science.

\subsubsection{Step-by-Step Guide to Install Anaconda:}
\begin{enumerate}
  \item Go to \url{https://www.anaconda.com/products/distribution} and download the installer for your operating system.
  \item Run the installer and follow the instructions.
  \item After installation, open \texttt{Anaconda Navigator}.
  \item From here, you can launch \texttt{Jupyter Notebook}, \texttt{Spyder} (another IDE), or create new environments.
  \item To launch a Jupyter Notebook, click on \texttt{Launch} under the Jupyter Notebook section. It will open a web-based notebook interface where you can write Python code and run it interactively.
  
For example, you can try the following code in a Jupyter Notebook:

\begin{lstlisting}[style=python]
import torch

x = torch.rand(5, 3)
print(x)
\end{lstlisting}
\end{enumerate}

Anaconda is excellent for data science and machine learning projects, as it makes managing dependencies and environments much simpler.

\subsection{Virtual Environments}
A virtual environment is an isolated environment that allows you to install specific packages for a project without affecting other projects or the global Python installation. This is especially useful when working on multiple projects that require different versions of the same library.

\subsubsection{Step-by-Step Guide to Set Up a Virtual Environment:}
\begin{enumerate}
  \item Open a terminal (or command prompt).
  \item Navigate to your project directory:
  
\begin{lstlisting}[style=cmd]
cd path/to/your/project
\end{lstlisting}
  \item Create a virtual environment by running:
  
\begin{lstlisting}[style=cmd]
python -m venv myenv
\end{lstlisting}
  Here, \texttt{myenv} is the name of your virtual environment. You can name it anything you like.
  \item Activate the virtual environment:
  
\begin{itemize}
  \item On Windows:
  
\begin{lstlisting}[style=cmd]
myenv\Scripts\activate
\end{lstlisting}

  \item On Mac/Linux:
  
\begin{lstlisting}[style=cmd]
source myenv/bin/activate
\end{lstlisting}
\end{itemize}

  \item Your terminal should now indicate that the virtual environment is active.
  \item You can install packages in this environment using \texttt{pip}. For example:

\begin{lstlisting}[style=cmd]
pip install torch
\end{lstlisting}
  \item To deactivate the virtual environment, simply run:
  
\begin{lstlisting}[style=cmd]
deactivate
\end{lstlisting}
\end{enumerate}

\begin{center}
\begin{tikzpicture}
    \tikzstyle{level 1}=[sibling distance=5cm]
    \tikzstyle{level 2}=[sibling distance=5cm]
    \node {Project Root}
        child {node {myenv (virtual environment)}
            child {node {Scripts (Windows) / bin (Linux/Mac)}}
            child {node {Lib}}
        }
        child {node {src}
            child {node {main.py}}
        };
\end{tikzpicture}
\end{center}

Virtual environments are essential for managing project dependencies efficiently, especially as your projects grow in complexity.

\section{Variables and Data Types}

In Python, variables are containers for storing data values. Unlike many other languages, Python does not require you to explicitly declare the data type of a variable. Python's interpreter automatically assigns the data type based on the value assigned.

Some of the most common data types in Python include:

\begin{itemize}
    \item \textbf{int} (Integer): Represents whole numbers, e.g., 5, -10, 100.
    \item \textbf{float} (Floating point): Represents decimal numbers, e.g., 3.14, -0.5.
    \item \textbf{str} (String): A sequence of characters, e.g., "Hello", 'Python'.
    \item \textbf{bool} (Boolean): Represents True or False.
    \item \textbf{list}: A collection of ordered items, which can be of different types, e.g., [1, "apple", 3.14].
    \item \textbf{dict}: A collection of key-value pairs, e.g., \{'name': 'John', 'age': 30\}.
\end{itemize}

\textbf{Example:}

\begin{lstlisting}[style=python]
# Defining variables
name = "Alice"   # str
age = 25         # int
height = 5.6     # float
is_student = True # bool

# Defining a list and dictionary
fruits = ["apple", "banana", "cherry"]
person = {"name": "Alice", "age": 25}
\end{lstlisting}

\section{Conditional Statements and Loops}

Conditional statements allow you to execute certain blocks of code based on conditions. The most common conditional statement is the \texttt{if-else} statement.

\subsection{If-Else}

\begin{lstlisting}[style=python]
age = 20
if age >= 18:
    print("You are an adult.")
else:
    print("You are a minor.")
\end{lstlisting}

\subsection{Loops}

Loops allow us to execute a block of code multiple times.

\textbf{For Loop:}
\begin{lstlisting}[style=python]
# Iterating over a list using a for loop
for fruit in fruits:
    print(fruit)
\end{lstlisting}

\textbf{While Loop:}
\begin{lstlisting}[style=python]
# Using a while loop
count = 0
while count < 5:
    print("Count:", count)
    count += 1
\end{lstlisting}

\section{Functions and Modules}

Functions are blocks of reusable code that perform specific tasks. Python has built-in functions like \texttt{print()}, but you can also define your functions.

\subsection{Defining a Function}
\begin{lstlisting}[style=python]
# Defining a function
def greet(name):
    return f"Hello, {name}!"

# Calling the function
print(greet("Alice"))
\end{lstlisting}

\subsection{Modules}

Python modules are files containing Python code. You can import and use functions from other modules. Python provides many built-in modules, such as \texttt{math} and \texttt{os}.

\begin{lstlisting}[style=python]
import math

# Using a function from the math module
result = math.sqrt(16)
print(result)
\end{lstlisting}

\section{File Handling}

Python provides built-in functions to work with files. You can read from and write to files using the \texttt{open()} function. Always remember to close the file after operations, or use a context manager with the \texttt{with} keyword.

\subsection{Reading a File}
\begin{lstlisting}[style=python]
# Reading from a file
with open("example.txt", "r") as file:
    content = file.read()
    print(content)
\end{lstlisting}

\subsection{Writing to a File}
\begin{lstlisting}[style=python]
# Writing to a file
with open("output.txt", "w") as file:
    file.write("Hello, Python!")
\end{lstlisting}
\textbf{Note:} Using \texttt{with} ensures that the file is properly closed after its block of code is executed, which is important for resource management.

\section{Object-Oriented Programming}

Python supports object-oriented programming (OOP), which allows you to define custom objects using classes.

\subsection{Defining a Class}

\begin{lstlisting}[style=python]
# Defining a class
class Dog:
    def __init__(self, name, breed):
        self.name = name
        self.breed = breed

    def bark(self):
        return f"{self.name} says woof!"

# Creating an object of the Dog class
my_dog = Dog("Buddy", "Golden Retriever")
print(my_dog.bark())
\end{lstlisting}

\subsection{Inheritance}

Inheritance allows one class to inherit attributes and methods from another class.

\begin{lstlisting}[style=python]
# Defining a parent class
class Animal:
    def __init__(self, name):
        self.name = name

    def make_sound(self):
        return "Some sound"

# Defining a child class that inherits from Animal
class Cat(Animal):
    def make_sound(self):
        return "Meow"

# Creating an object of the Cat class
my_cat = Cat("Whiskers")
print(my_cat.make_sound())  # Output: Meow
\end{lstlisting}

\section{Exception Handling}

Python provides a way to handle errors using \texttt{try-except} blocks. This prevents your program from crashing when an error occurs and allows you to provide meaningful error messages.

\begin{lstlisting}[style=python]
# Handling exceptions
try:
    x = int(input("Enter a number: "))
    print(f"Result: {10 / x}")
except ZeroDivisionError:
    print("Error: Cannot divide by zero.")
except ValueError:
    print("Error: Invalid input, please enter a valid number.")
finally:
    print("This block is always executed.")
\end{lstlisting}

%% file: 02_basic_lib.tex
\section{Introduction to Common Python Libraries}
In this chapter, we will explore some of the most popular and essential Python libraries used for data manipulation, scientific computing, machine learning, and data visualization. These libraries form the foundation for various data-driven applications and are widely used in both academic research and industry. By mastering these libraries, you'll have a strong toolkit for solving real-world problems efficiently.

\subsection{Installing Libraries}
To work with the libraries mentioned, such as Numpy, Pandas, Matplotlib, Scikit-learn, PyTorch, and TensorFlow, you will first need to install them. Below are the installation instructions using both \texttt{pip} and \texttt{conda} package managers.

\subsubsection{Installing with pip}
\texttt{pip} is the Python package manager and can be used to install all the libraries with the following commands:

\begin{lstlisting}[style=cmd]
# Installing Numpy
pip install numpy

# Installing Pandas
pip install pandas

# Installing Matplotlib
pip install matplotlib

# Installing Scikit-learn
pip install scikit-learn

# Installing PyTorch
pip install torch torchvision torchaudio

# Installing TensorFlow
pip install tensorflow
\end{lstlisting}

These commands will install the necessary packages from the Python Package Index (PyPI). Make sure that you have \texttt{pip} installed and properly configured in your environment.

\subsubsection{Installing with conda}
\texttt{conda} is another package manager commonly used in data science, especially with the Anaconda distribution. To install the same libraries using \texttt{conda}, use the following commands:

\begin{lstlisting}[style=cmd]
# Installing Numpy
conda install numpy

# Installing Pandas
conda install pandas

# Installing Matplotlib
conda install matplotlib

# Installing Scikit-learn
conda install scikit-learn

# Installing PyTorch
conda install pytorch torchvision torchaudio cpuonly -c pytorch

# Installing TensorFlow
conda install tensorflow
\end{lstlisting}

Using \texttt{conda} ensures that dependencies are properly managed, especially for complex libraries like PyTorch and TensorFlow, which may require specific versions of other packages or CUDA support for GPU acceleration.

\subsection{Numpy}
NumPy~\cite{numpy} is the fundamental package for scientific computing in Python. It provides support for arrays and matrices, along with a collection of mathematical functions to operate on these data structures. NumPy arrays are more efficient than Python lists, and they provide a more compact way of working with large amounts of data.

\subsubsection{Basic Array Operations with Numpy}
NumPy arrays can be created from Python lists, or directly using functions such as \texttt{numpy.array()} or \texttt{numpy.zeros()}.

\begin{lstlisting}[style=python]
import numpy as np

# Creating a 1D array from a Python list
arr = np.array([1, 2, 3, 4, 5])
print(arr)

# Creating a 2D array (matrix) of zeros
matrix = np.zeros((3, 3))
print(matrix)
\end{lstlisting}

In the above code, \texttt{arr} is a simple one-dimensional array, while \texttt{matrix} is a two-dimensional array (3x3) of zeros. NumPy provides various functions to reshape arrays, perform element-wise operations, and execute linear algebra functions.

\subsubsection{Basic Matrix Operations}
Let's consider some common matrix operations like addition, multiplication, and transpose.

\begin{lstlisting}[style=python]
# Create two 2x2 matrices
A = np.array([[1, 2], [3, 4]])
B = np.array([[5, 6], [7, 8]])

# Matrix addition
C = A + B
print(C)

# Element-wise multiplication
D = A * B
print(D)

# Matrix transpose
transpose_A = A.T
print(transpose_A)
\end{lstlisting}

These simple operations are essential building blocks for scientific computing tasks, including machine learning and data analysis.

\subsection{Pandas}
Pandas~\cite{pandas} is a powerful library for data manipulation and analysis. It introduces two main data structures: \texttt{Series} and \texttt{DataFrame}. A \texttt{Series} is a one-dimensional array, while a \texttt{DataFrame} is a two-dimensional, table-like structure.

\subsubsection{Creating and Manipulating DataFrames}
Here is how you can create and manipulate DataFrames using Pandas.

\begin{lstlisting}[style=python]
import pandas as pd

# Creating a DataFrame from a dictionary
data = {'Name': ['Alice', 'Bob', 'Charlie'],
        'Age': [25, 30, 35],
        'Salary': [70000, 80000, 90000]}

df = pd.DataFrame(data)

# Viewing the DataFrame
print(df)

# Selecting a column
print(df['Name'])

# Filtering rows based on a condition
filtered_df = df[df['Age'] > 28]
print(filtered_df)
\end{lstlisting}

In this example, a DataFrame \texttt{df} is created from a dictionary. We then show how to select a specific column and filter rows based on conditions.

\subsubsection{Handling Missing Data}
Data in the real world is often incomplete or contains missing values. Pandas provides powerful tools to handle missing data.

\begin{lstlisting}[style=python]
# Adding a column with missing data
df['Bonus'] = [5000, None, 7000]

# Filling missing values with a specific number
df_filled = df.fillna(0)
print(df_filled)

# Dropping rows with missing values
df_dropped = df.dropna()
print(df_dropped)
\end{lstlisting}

These operations are vital for cleaning and preprocessing data before it can be used in machine learning models.

\subsubsection{Reading and Writing Data}
Pandas provides versatile functions for reading from and writing to various file formats such as CSV, Excel, and SQL databases. Here are some examples of how to read and write data using Pandas.

\begin{lstlisting}[style=python]
# Reading a CSV file
df_csv = pd.read_csv('data.csv')
print(df_csv)

# Reading an Excel file
df_excel = pd.read_excel('data.xlsx', sheet_name='Sheet1')
print(df_excel)

# Reading a JSON file
df_json = pd.read_json('data.json')
print(df_json)

# Writing a DataFrame to a CSV file
df.to_csv('output.csv', index=False)

# Writing a DataFrame to an Excel file
df.to_excel('output.xlsx', sheet_name='Results', index=False)

# Writing a DataFrame to a JSON file
df.to_json('output.json')
\end{lstlisting}

In these examples, \texttt{pd.read\_csv}, \texttt{pd.read\_excel}, and \texttt{pd.read\_json} are used to load data from CSV, Excel, and JSON formats, respectively. Similarly, \texttt{to\_csv}, \texttt{to\_excel}, and \texttt{to\_json} are used to save DataFrames to these formats.

\subsubsection{Reading from SQL Databases}
Pandas can also connect to SQL databases to read data directly into a DataFrame.

\begin{lstlisting}[style=python]
import sqlite3

# Creating a connection to the database
conn = sqlite3.connect('example.db')

# Reading from SQL
df_sql = pd.read_sql_query('SELECT * FROM employees', conn)
print(df_sql)

# Writing a DataFrame to a SQL database
df.to_sql('employees', conn, if_exists='replace', index=False)

# Closing the connection
conn.close()
\end{lstlisting}

Here, \texttt{pd.read\_sql\_query} is used to fetch data from an SQL database, and \texttt{to\_sql} is used to write data back into the database. The \texttt{if\_exists='replace'} argument ensures that the table is replaced if it already exists.

\subsection{Matplotlib}
Matplotlib~\cite{matplotlib_} is a library used for data visualization in Python. It allows you to create a variety of static, animated, and interactive plots.

\subsubsection{Plotting with Matplotlib}
The basic plot is a 2D line graph, but Matplotlib can also handle bar charts, histograms, scatter plots, and more.

\begin{lstlisting}[style=python]
import matplotlib.pyplot as plt

# Creating a simple line plot
x = [0, 1, 2, 3, 4, 5]
y = [0, 1, 4, 9, 16, 25]

plt.plot(x, y)
plt.title('Simple Line Plot')
plt.xlabel('X-axis')
plt.ylabel('Y-axis')
plt.show()
\end{lstlisting}

Here, we plot a simple quadratic function. You can customize the plot with titles, labels, and other formatting options.

\subsubsection{Creating a Bar Plot}
In addition to line plots, Matplotlib supports bar plots, which are useful for comparing categorical data.

\begin{lstlisting}[style=python]
# Creating a bar plot
categories = ['A', 'B', 'C']
values = [5, 7, 3]

plt.bar(categories, values)
plt.title('Simple Bar Plot')
plt.show()
\end{lstlisting}

Visualization is a key aspect of data analysis, and Matplotlib allows you to explore your data visually, which is critical in identifying patterns or insights.

\subsection{Scikit-learn}
Scikit-learn~\cite{pedregosa2011scikit} is one of the most popular libraries for building machine learning models. It provides efficient implementations of machine learning algorithms like linear regression, decision trees, clustering, and more.

\subsubsection{Building a Simple Machine Learning Model}
Let's build a simple linear regression model using Scikit-learn.

\begin{lstlisting}[style=python]
from sklearn.model_selection import train_test_split
from sklearn.linear_model import LinearRegression

# Example dataset
X = [[1], [2], [3], [4], [5]]
y = [1, 2, 3, 4, 5]

# Splitting data into training and testing sets
X_train, X_test, y_train, y_test = train_test_split(X, y, test_size=0.2)

# Creating and training the model
model = LinearRegression()
model.fit(X_train, y_train)

# Making predictions
predictions = model.predict(X_test)
print(predictions)
\end{lstlisting}

In this example, we use a simple dataset for linear regression. We first split the data into training and test sets, train the model, and make predictions on unseen data.

\subsection{PyTorch}
PyTorch~\cite{pytorch_} is an open-source deep learning framework widely used for developing machine learning models, especially in the area of deep learning and neural networks. It is known for its flexibility and ease of use, and it provides automatic differentiation through its \texttt{autograd} feature.

\subsubsection{Building a Simple Neural Network}
We will now build a simple feedforward neural network using PyTorch.

\begin{lstlisting}[style=python]
import torch
import torch.nn as nn
import torch.optim as optim

# Define a simple feedforward neural network
class SimpleNN(nn.Module):
    def __init__(self):
        super(SimpleNN, self).__init__()
        self.fc1 = nn.Linear(1, 10)
        self.fc2 = nn.Linear(10, 1)

    def forward(self, x):
        x = torch.relu(self.fc1(x))
        x = self.fc2(x)
        return x

# Create the network and the optimizer
model = SimpleNN()
optimizer = optim.SGD(model.parameters(), lr=0.01)
criterion = nn.MSELoss()

# Example dataset
X = torch.tensor([[1.0], [2.0], [3.0], [4.0], [5.0]])
y = torch.tensor([[1.0], [2.0], [3.0], [4.0], [5.0]])

# Training loop
for epoch in range(100):
    optimizer.zero_grad()
    output = model(X)
    loss = criterion(output, y)
    loss.backward()
    optimizer.step()

# Make predictions
with torch.no_grad():
    predictions = model(X)
    print(predictions)
\end{lstlisting}

In this code, we define a simple two-layer neural network with one hidden layer. We use stochastic gradient descent (SGD) as the optimizer and mean squared error (MSE) as the loss function. The network is trained on a simple dataset to learn the identity function.

\subsection{TensorFlow}

TensorFlow~\cite{tensorflow2015-whitepaper} is a well-known open-source deep learning framework developed by Google. It has been widely adopted for building and training machine learning models, particularly in production environments. TensorFlow provides both high-level and low-level APIs, offering flexibility and scalability for various tasks. In addition to its general applications, TensorFlow also supports the use of pre-trained models, making it a powerful tool for transfer learning and fine-tuning~\cite{chen2024deeplearningmachinelearning}.

\subsubsection{Building a Simple Neural Network}
We will now build a simple feedforward neural network using TensorFlow.

\begin{lstlisting}[style=python]
import tensorflow as tf
from tensorflow.keras import layers, models

# Define a simple feedforward neural network
model = models.Sequential([
    layers.Dense(10, activation='relu', input_shape=(1,)),
    layers.Dense(1)
])

# Compile the model
model.compile(optimizer='sgd', loss='mse')

# Example dataset
X = tf.constant([[1.0], [2.0], [3.0], [4.0], [5.0]])
y = tf.constant([[1.0], [2.0], [3.0], [4.0], [5.0]])

# Train the model
model.fit(X, y, epochs=100)

# Make predictions
predictions = model.predict(X)
print(predictions)
\end{lstlisting}

In this example, we define a simple neural network using TensorFlow's \texttt{Sequential} model API. The network consists of two layers: a hidden layer with 10 neurons and ReLU activation, and an output layer with a single neuron. We use stochastic gradient descent (SGD) as the optimizer and mean squared error (MSE) as the loss function. The network is trained on the same simple dataset to learn the identity function.

\subsection{Why PyTorch Over TensorFlow?}
In recent years, PyTorch has gained significant popularity over TensorFlow, particularly in the research community and among machine learning practitioners. While TensorFlow was once the dominant framework, several factors have led to the shift toward PyTorch.

\textbf{Ease of Use:} PyTorch offers a more intuitive, Pythonic interface, which makes it easier to learn and experiment with, especially for beginners. Its dynamic computation graph (as opposed to TensorFlow's earlier static graph approach) allows for more flexibility and ease in debugging.

\textbf{Adoption in Research:} PyTorch's flexibility and ease of experimentation have made it the framework of choice in academia and research. Many research papers and advanced models are now developed using PyTorch, and community support has grown significantly.

\textbf{Unified Ecosystem:} PyTorch has a unified ecosystem, including libraries like \texttt{torchvision} for computer vision, \texttt{torchaudio} for audio processing, and \texttt{torchtext} for NLP tasks. These libraries provide pre-built tools and datasets, making it easier for users to implement models.

While TensorFlow remains a powerful tool, especially for production environments and large-scale deployments, beginners and researchers may find PyTorch more accessible. If you are new to machine learning or deep learning, starting with PyTorch can offer a smoother learning experience.

%% file: 03_ml.tex
\chapter{Machine Learning Fundamentals}

\section{Basic Concepts of Machine Learning}

Machine learning is a subset of artificial intelligence (AI) that enables systems to learn and make decisions based on data. Unlike traditional programming, where explicit rules are written by a programmer, machine learning models automatically infer these rules from the data provided. There are three primary categories of machine learning: supervised learning, unsupervised learning, and reinforcement learning.

\subsection{Supervised Learning}

Supervised learning is the most common form of machine learning. In this type, the model is trained using a labeled dataset, meaning that each input comes with an associated output. The goal of the algorithm is to learn the relationship between inputs and outputs in such a way that it can predict the output for new, unseen data~\cite{friedman2001elements}. Supervised learning is typically used in applications like classification (e.g., spam detection in emails~\cite{kotsiantis2007supervised}) and regression (e.g., predicting housing prices)~\cite{hastie2009elements}.

Example:
\begin{lstlisting}[style=python]
import torch
import torch.nn as nn
import torch.optim as optim

# Example: Supervised learning with PyTorch for a simple binary classification
# Define the model
class SimpleNN(nn.Module):
    def __init__(self):
        super(SimpleNN, self).__init__()
        self.layer1 = nn.Linear(2, 1)  # Input: 2 features, Output: 1 (binary class)

    def forward(self, x):
        return torch.sigmoid(self.layer1(x))

# Data (features and labels)
X_train = torch.tensor([[0.0, 1.0], [1.0, 0.0], [0.0, 0.0], [1.0, 1.0]], dtype=torch.float32)
y_train = torch.tensor([[1], [1], [0], [0]], dtype=torch.float32)

# Define the model, loss function, and optimizer
model = SimpleNN()
criterion = nn.BCELoss()  # Binary Cross Entropy Loss for classification
optimizer = optim.SGD(model.parameters(), lr=0.01)

# Training loop
for epoch in range(1000):
    optimizer.zero_grad()
    outputs = model(X_train)
    loss = criterion(outputs, y_train)
    loss.backward()
    optimizer.step()
\end{lstlisting}

\subsection{Unsupervised Learning}

In unsupervised learning, the model is trained on data that has no labels. The goal is to uncover hidden patterns or structures within the data~\cite{ghahramani2004unsupervised}. A common application of unsupervised learning is clustering, where the model groups similar data points together~\cite{jain1999data}. Another example is dimensionality reduction, where the model reduces the number of features in the dataset while retaining essential information~\cite{van2009dimensionality}.

Example:
\begin{lstlisting}[style=python]
from sklearn.cluster import KMeans
import torch

# Unsupervised learning with clustering (KMeans in sklearn)
data = torch.tensor([[1, 2], [2, 3], [3, 4], [8, 9], [9, 10], [10, 11]], dtype=torch.float32)
kmeans = KMeans(n_clusters=2)  # Finding 2 clusters in the data
clusters = kmeans.fit_predict(data.numpy())
print(clusters)  # Output: Cluster labels for each data point
\end{lstlisting}

\subsection{Reinforcement Learning}

Reinforcement learning is different from both supervised and unsupervised learning. In reinforcement learning, an agent interacts with an environment and learns by receiving feedback in the form of rewards or penalties~\cite{sutton2018reinforcement}. The agent takes actions to maximize cumulative rewards over time. Applications include game playing, robotics, and self-driving cars~\cite{mnih2015human}.

Example:
Imagine a robot learning to navigate a maze. Each time the robot takes a step, it receives a reward if it moves closer to the goal and a penalty if it moves further away. The robot continues to explore and adjust its actions based on the rewards and penalties received, with the ultimate aim of finding the shortest path to the goal~\cite{kaelbling1996reinforcement}.

\section{Supervised vs Unsupervised Learning}

\subsection{Supervised Learning: A Detailed Look}

Supervised learning is ideal for situations where we have a clear idea of the desired output based on the given input. One of the most common uses of supervised learning is in predictive modeling, where we use past data to predict future outcomes. Examples include predicting stock prices, classifying whether an email is spam, and recognizing handwritten digits.

Example:
Let's train a PyTorch neural network to classify whether an input is positive or negative.

\begin{lstlisting}[style=python]
import torch
import torch.nn as nn
import torch.optim as optim

# Define a simple binary classification model
class Classifier(nn.Module):
    def __init__(self):
        super(Classifier, self).__init__()
        self.layer1 = nn.Linear(1, 1)

    def forward(self, x):
        return torch.sigmoid(self.layer1(x))

# Data: inputs and labels
X_train = torch.tensor([[-1.0], [2.0], [-3.0], [4.0]], dtype=torch.float32)
y_train = torch.tensor([[0], [1], [0], [1]], dtype=torch.float32)

# Model, loss function, optimizer
model = Classifier()
criterion = nn.BCELoss()
optimizer = optim.SGD(model.parameters(), lr=0.1)

# Training loop
for epoch in range(1000):
    optimizer.zero_grad()
    outputs = model(X_train)
    loss = criterion(outputs, y_train)
    loss.backward()
    optimizer.step()
\end{lstlisting}

\subsection{Unsupervised Learning: A Detailed Look}

In contrast, unsupervised learning is used when we only have input data but no corresponding output labels. This is useful when we want to discover the underlying structure of the data. One of the main applications is clustering, where the algorithm identifies similar groups within the data.

For example, customer segmentation in marketing can be achieved using clustering algorithms like KMeans, which groups customers into similar segments based on their purchasing behavior.

Example:
\begin{lstlisting}[style=python]
import torch
from sklearn.decomposition import PCA

# Unsupervised learning with dimensionality reduction (PCA in sklearn)
data = torch.tensor([[2.5, 2.4], [0.5, 0.7], [2.2, 2.9], [1.9, 2.2]], dtype=torch.float32)
pca = PCA(n_components=1)  # Reduce to 1 dimension
reduced_data = pca.fit_transform(data.numpy())
print(reduced_data)  # Output: Data transformed to 1 dimension
\end{lstlisting}

\section{Model Evaluation and Performance Metrics}

Model evaluation is a crucial part of machine learning. It involves assessing how well a trained model performs on unseen data. There are several performance metrics that help us understand different aspects of the model's performance, particularly in classification problems.

\subsection{Accuracy, Precision, Recall, and F1-score}

\textbf{Accuracy} is the simplest performance metric. It is the ratio of correctly predicted instances to the total number of instances. However, accuracy may not always be a good metric, especially in cases where the classes are imbalanced~\cite{he2009learning}.

\textbf{Precision} is the ratio of true positive predictions to the total number of positive predictions (both true and false positives). It answers the question: "Of all the instances predicted as positive, how many were actually positive?"~\cite{sokolova2006beyond}.

\textbf{Recall} (also called sensitivity) is the ratio of true positive predictions to the total number of actual positive instances. It answers the question: "Of all the actual positive instances, how many did the model correctly predict?"~\cite{sokolova2006beyond}.

\textbf{F1-score} is the harmonic mean of precision and recall. It provides a balanced measure that takes both false positives and false negatives into account. It is especially useful when the class distribution is imbalanced~\cite{goutte2005probabilistic}.

\begin{lstlisting}[style=python]
from sklearn.metrics import accuracy_score, precision_score, recall_score, f1_score

# Example data: ground truth and predictions
y_true = [1, 0, 1, 1, 0, 1, 0, 0, 1]
y_pred = [1, 0, 1, 0, 0, 1, 1, 0, 1]

# Calculate metrics
accuracy = accuracy_score(y_true, y_pred)
precision = precision_score(y_true, y_pred)
recall = recall_score(y_true, y_pred)
f1 = f1_score(y_true, y_pred)

print(f"Accuracy: {accuracy}, Precision: {precision}, Recall: {recall}, F1-Score: {f1}")
\end{lstlisting}

\subsection{ROC Curve and AUC}

The \textbf{ROC curve} (Receiver Operating Characteristic curve) is a graphical representation of the performance of a binary classifier across different threshold values~\cite{fawcett2006introduction}. It is particularly useful for understanding the trade-off between two important metrics: the \textit{true positive rate} (recall or sensitivity) and the \textit{false positive rate} (1-specificity). The ROC curve plots the true positive rate on the y-axis and the false positive rate on the x-axis.

The ROC curve has its origins in World War II, where it was first used by radar operators to detect enemy objects. The operators had to balance the detection of real objects (true positives) against false alarms (false positives). This led to the development of the ROC curve to evaluate how well different detection strategies worked under varying conditions~\cite{swets1979receiver}. Over time, this concept was adapted for evaluating binary classification models in machine learning and medical testing~\cite{hanley1982meaning}.

The \textit{AUC} (Area Under the Curve) is a single scalar value that summarizes the entire ROC curve. The AUC ranges between 0 and 1:
\begin{itemize}
    \item An AUC of 1.0 indicates a perfect classifier, meaning it has a high true positive rate and a low false positive rate across all thresholds.
    \item An AUC of 0.5 indicates that the classifier performs no better than random guessing~\cite{bradley1997use}.
    \item AUC values below 0.5 suggest that the model is worse than random guessing, potentially misclassifying the results.
\end{itemize}

\subsubsection{Why Use the ROC Curve?}
The ROC curve is often used when dealing with imbalanced datasets or when you are more interested in the ranking ability of your classifier rather than just a single accuracy score. By varying the decision threshold (the cutoff for predicting class labels), the ROC curve shows how sensitive your model is to detecting true positives while minimizing false positives. This is especially important in scenarios like medical diagnostics, where detecting a disease (true positive) may be far more critical than the cost of a false alarm (false positive).

\subsubsection{Example: ROC Curve Calculation}
To better understand the ROC curve, let's walk through an example. Consider a binary classification model designed to predict whether a patient has a certain disease (positive class) or not (negative class). The model outputs a probability score between 0 and 1 for each patient. Based on this score, the model decides whether to classify the patient as positive (disease present) or negative (disease absent) by applying a threshold.

For instance, suppose we have the following probability predictions from the model:

\begin{center}
\begin{tabular}{|c|c|}
\hline
\textbf{Patient} & \textbf{Predicted Probability (Disease)} \\
\hline
1 & 0.9 \\
2 & 0.7 \\
3 & 0.4 \\
4 & 0.3 \\
5 & 0.8 \\
6 & 0.2 \\
\hline
\end{tabular}
\end{center}

We can apply a threshold to these probabilities to convert them into binary decisions (disease present vs. disease absent). For example, if we set a threshold of 0.5:
\begin{itemize}
    \item Patients with a predicted probability greater than or equal to 0.5 will be classified as positive (disease present).
    \item Patients with a predicted probability below 0.5 will be classified as negative (disease absent).
\end{itemize}

The ROC curve is generated by varying this threshold and calculating the corresponding true positive rate (TPR) and false positive rate (FPR) for each threshold. For example, if we start with a high threshold of 1.0, no patient will be classified as positive, resulting in a TPR of 0 and FPR of 0. As we lower the threshold, more patients will be classified as positive, increasing both the TPR and FPR.

\subsubsection{How to Compute AUC}
The AUC value is computed by calculating the area under the ROC curve. This can be done numerically by summing up the area under each segment of the curve. A model that consistently classifies positive samples with higher probabilities than negative samples will have a higher AUC.

For example, imagine we sort the predicted probabilities from our classifier in descending order. A perfect model would always rank positive samples higher than negative ones, resulting in an AUC of 1. If the classifier ranks positive and negative samples equally often, the AUC would be 0.5, equivalent to random guessing. A good classifier ranks positive samples higher than negative ones most of the time, resulting in an AUC somewhere between 0.5 and 1.0.

In practical terms, calculating the AUC involves integrating the ROC curve, and in Python, this can be done easily with libraries like \texttt{scikit-learn}:

\begin{lstlisting}[style=python]
from sklearn.metrics import roc_curve, auc

# Example binary labels and predicted probabilities
y_true = [0, 0, 1, 1]
y_scores = [0.1, 0.4, 0.35, 0.8]

# Compute the ROC curve
fpr, tpr, thresholds = roc_curve(y_true, y_scores)

# Compute AUC
roc_auc = auc(fpr, tpr)
print(f"AUC: {roc_auc}")
\end{lstlisting}

In this code, \texttt{roc\_curve} calculates the false positive rate and true positive rate at various threshold settings, and \texttt{auc} computes the area under the ROC curve. The resulting AUC score gives a single number that helps summarize the model's performance.

\textbf{ROC is a curve. Here is the code to draw the ROC curve:}

\begin{lstlisting}[style=python]
from sklearn.metrics import roc_curve, auc
import matplotlib.pyplot as plt

# Example data: ground truth and predicted probabilities
y_true = [0, 0, 1, 1]
y_scores = [0.1, 0.4, 0.35, 0.8]

# Calculate ROC curve
fpr, tpr, _ = roc_curve(y_true, y_scores)
roc_auc = auc(fpr, tpr)

# Plot ROC curve
plt.figure()
plt.plot(fpr, tpr, label=f'ROC curve (AUC = {roc_auc:.2f})')
plt.plot([0, 1], [0, 1], 'k--')
plt.xlim([0.0, 1.0])
plt.ylim([0.0, 1.0])
plt.xlabel('False Positive Rate')
plt.ylabel('True Positive Rate')
plt.title('Receiver Operating Characteristic')
plt.legend(loc="lower right")
plt.show()
\end{lstlisting}

%% file: 04_preprocessing.tex
\chapter{Data Preprocessing}
    
Data preprocessing is a critical step in any machine learning project. Without proper data preprocessing, even the most sophisticated algorithms can underperform. In this chapter, we will discuss various preprocessing techniques, focusing on data cleaning, standardization, normalization, and feature engineering. These techniques are essential for improving model performance and ensuring that the data is ready for analysis.

\section{Data Cleaning and Missing Value Handling}

Raw data often contains noise, inconsistencies, and missing values, which can negatively impact the performance of machine learning models. In this section, we will focus on how to clean the data and handle missing values. 

Missing data is a common issue, and handling it correctly is crucial. There are several ways to deal with missing data:
\begin{enumerate}
    \item \textbf{Remove rows or columns with missing data}: This is the simplest method but may result in losing valuable information.
    \item \textbf{Fill missing data with a value (Imputation)}: Missing values can be filled with a specific value like the mean, median, or a placeholder value.
    \item \textbf{Predict missing values}: Machine learning algorithms can be used to predict the missing values based on other features.
\end{enumerate}

Let’s look at an example using \texttt{pandas} and \texttt{PyTorch}:

\begin{lstlisting}[style=python]
import pandas as pd
import torch
from sklearn.impute import SimpleImputer

# Sample data with missing values
data = {'Feature1': [1.0, 2.0, None, 4.0],
        'Feature2': [None, 2.5, 3.5, None],
        'Feature3': [1.5, None, 2.5, 3.5]}

df = pd.DataFrame(data)

# Handling missing data by filling with the mean of each column
imputer = SimpleImputer(strategy='mean')
df_filled = pd.DataFrame(imputer.fit_transform(df), columns=df.columns)

# Convert the DataFrame to a PyTorch tensor for further processing
tensor_data = torch.tensor(df_filled.values)
print(tensor_data)
\end{lstlisting}

In the example above, we use \texttt{SimpleImputer} from \texttt{scikit-learn} to fill the missing values with the mean of each column. After that, we convert the DataFrame into a \texttt{PyTorch} tensor to proceed with any further steps. This workflow ensures that the missing data does not affect the training process.

\section{Data Standardization and Normalization}

Before feeding data into machine learning models, especially models like neural networks, it is often necessary to scale the data. The two most common scaling methods are standardization and normalization:

\begin{itemize}
    \item \textbf{Standardization}: Rescales the data so that it has a mean of zero and a standard deviation of one.
    \item \textbf{Normalization}: Rescales the data to a fixed range, usually [0, 1].
\end{itemize}

Why are these techniques important? Many machine learning algorithms, such as gradient descent, perform better when input features are on a similar scale. This prevents any single feature from disproportionately influencing the model. 

Let’s implement both standardization and normalization:

\begin{lstlisting}[style=python]
from sklearn.preprocessing import StandardScaler, MinMaxScaler

# Standardization: mean = 0, std = 1
scaler_standard = StandardScaler()
standardized_data = scaler_standard.fit_transform(df_filled)

# Normalization: scaling to range [0, 1]
scaler_minmax = MinMaxScaler()
normalized_data = scaler_minmax.fit_transform(df_filled)

# Convert standardized and normalized data into PyTorch tensors
tensor_standardized = torch.tensor(standardized_data)
tensor_normalized = torch.tensor(normalized_data)

print(tensor_standardized)
print(tensor_normalized)
\end{lstlisting}

In this example, we first standardize the data using \texttt{StandardScaler} and then normalize it using \texttt{MinMaxScaler}. Both methods ensure that the data is scaled appropriately for different types of models. 

\section{Feature Engineering}

Feature engineering involves creating new features or modifying existing ones to improve model performance. It is often said that better features lead to better models, and this is true in practice.

In this section, we will discuss two important aspects of feature engineering: feature selection and feature extraction.

\subsection{Feature Selection}

Feature selection is the process of selecting the most important features from the data. Not all features are equally valuable, and some may even reduce the performance of the model due to overfitting or increased noise.

There are different techniques for feature selection:
\begin{itemize}
    \item \textbf{Correlation-based selection}: Select features that have high correlation with the target variable but low correlation with each other~\cite{hall1999correlation}.
    \item \textbf{Recursive Feature Elimination (RFE)}: Iteratively remove less important features and evaluate model performance~\cite{guyon2002gene}.
    \item \textbf{Tree-based methods}: Use the importance scores generated by tree-based models like Random Forests or Gradient Boosting to select features~\cite{breiman2001random,chen2016xgboost}.
\end{itemize}

Let’s see an example using a tree-based method to perform feature selection:

\begin{lstlisting}[style=python]
from sklearn.ensemble import RandomForestClassifier
import numpy as np

# Sample dataset
X = np.array([[1, 2, 3], [4, 5, 6], [7, 8, 9], [2, 3, 4]])
y = np.array([0, 1, 1, 0])

# Fit a random forest classifier
clf = RandomForestClassifier(n_estimators=100)
clf.fit(X, y)

# Get feature importance scores
importance = clf.feature_importances_

# Select the most important features (importance > 0.3 for this example)
important_features = X[:, importance > 0.3]
print("Selected Features:", important_features)
\end{lstlisting}

In this example, we train a \texttt{RandomForestClassifier} and extract the feature importance scores. Features with an importance score above a certain threshold are selected.

\subsection{Feature Extraction}

Feature extraction reduces the dimensionality of the data by transforming features into a lower-dimensional space while retaining essential information. Principal Component Analysis (PCA) is one of the most widely used methods for feature extraction.

Here’s an example using PCA:

\begin{lstlisting}[style=python]
from sklearn.decomposition import PCA

# Sample dataset
X = np.array([[1, 2, 3], [4, 5, 6], [7, 8, 9], [2, 3, 4]])

# Apply PCA to reduce to 2 dimensions
pca = PCA(n_components=2)
X_reduced = pca.fit_transform(X)

print("Reduced Features:", X_reduced)
\end{lstlisting}

In this example, we use PCA to reduce the data from 3 dimensions to 2. PCA helps capture the maximum variance in the data, making it easier for models to interpret.

\section{Conclusion}

Data preprocessing is an essential part of any machine learning pipeline. In this chapter, we covered the fundamental steps of data cleaning, handling missing values, standardizing and normalizing data, as well as performing feature engineering through feature selection and extraction. These steps ensure that your data is in the best possible shape for machine learning models, which can lead to significant improvements in performance.

%% file: 05_linear.tex
\part{Linear Models and Classifiers}

\chapter{Linear Regression}
    
    \section{Basic Principles of Linear Regression}
    Linear regression is a fundamental machine learning algorithm used for predicting a continuous target variable based on one or more input variables (also called features)~\cite{montgomery2012introduction}. The core idea is to model the relationship between the input variables and the output variable as a linear combination of the input features.
    
    Suppose you have a dataset with $n$ samples, and each sample has $m$ features. You can express the linear relationship between the input variables $\mathbf{x} = [x_1, x_2, \dots, x_m]^T$ and the target variable $y$ as:
    
    \[
    y = w_1 x_1 + w_2 x_2 + \dots + w_m x_m + b
    \]
    
    Here:
    \begin{itemize}
        \item $w_1, w_2, \dots, w_m$ are the weights (or coefficients) for the features.
        \item $b$ is the bias (or intercept) term.
        \item $\mathbf{x}$ is the vector of input features.
        \item $y$ is the predicted output.
    \end{itemize}
    
    In matrix form, this can be written as:
    
    \[
    \mathbf{y} = \mathbf{X} \mathbf{w} + b
    \]
    
    Where:
    \begin{itemize}
        \item $\mathbf{X}$ is the $n \times m$ matrix of input features.
        \item $\mathbf{w}$ is the vector of weights of size $m$.
        \item $\mathbf{y}$ is the vector of target values of size $n$.
    \end{itemize}
    
    The goal of linear regression is to find the values of $\mathbf{w}$ and $b$ that minimize the difference between the predicted values $\mathbf{y}$ and the actual target values.

    \subsection{Applications of Linear Regression}
    Linear regression is widely used in various fields due to its simplicity and interpretability. Some common applications include:
    \begin{itemize}
        \item Predicting housing prices based on features like area, location, and the number of rooms.
        \item Estimating the relationship between marketing expenditure and sales.
        \item Modeling the relationship between temperature and energy consumption.
    \end{itemize}
    
    \section{Ordinary Least Squares}
    The most common method for estimating the coefficients $\mathbf{w}$ and $b$ in linear regression is called \textbf{Ordinary Least Squares (OLS)}. The idea behind OLS is to minimize the \textit{sum of the squared differences} between the predicted values $\mathbf{\hat{y}} = \mathbf{Xw} + b$ and the actual target values $\mathbf{y}$~\cite{seber2003linear}.

    The \textbf{cost function} or \textbf{loss function} for linear regression is defined as:
    
    \[
    J(\mathbf{w}, b) = \frac{1}{2n} \sum_{i=1}^{n} (y_i - \hat{y}_i)^2 = \frac{1}{2n} \sum_{i=1}^{n} \left( y_i - (\mathbf{x}_i^T \mathbf{w} + b) \right)^2
    \]
    
    Where:
    \begin{itemize}
        \item $n$ is the number of data points.
        \item $y_i$ is the actual value for the $i$-th data point.
        \item $\hat{y}_i$ is the predicted value for the $i$-th data point.
    \end{itemize}
    
    The OLS method aims to find the values of $\mathbf{w}$ and $b$ that minimize the loss function $J(\mathbf{w}, b)$. This can be solved using optimization techniques such as gradient descent or by using the closed-form solution.
    
    \subsection{Closed-Form Solution}
    In some cases, we can directly solve for the weights $\mathbf{w}$ and $b$ using a closed-form solution. The weights that minimize the loss function are given by:
    
    \[
    \mathbf{w} = (\mathbf{X}^T \mathbf{X})^{-1} \mathbf{X}^T \mathbf{y}
    \]
    
    This approach works well when the number of features $m$ is small, but it becomes computationally expensive when $m$ is large, especially because it requires calculating the inverse of the matrix $\mathbf{X}^T \mathbf{X}$.
    
    \section{Regularization: Lasso and Ridge Regression}
    One challenge in linear regression is \textit{overfitting}, which occurs when the model becomes too complex and performs well on the training data but poorly on unseen data. To combat overfitting, we use \textbf{regularization} techniques like Lasso and Ridge regression.

    \subsection{Ridge Regression}
    Ridge regression adds a penalty term to the loss function, which helps constrain the size of the weights and thus reduces the model's complexity~\cite{hoerl1970ridge}. The modified loss function for Ridge regression is:
    
    \[
    J(\mathbf{w}, b) = \frac{1}{2n} \sum_{i=1}^{n} (y_i - \hat{y}_i)^2 + \lambda \|\mathbf{w}\|^2
    \]
    
    Where:
    \begin{itemize}
        \item $\lambda$ is a regularization parameter that controls the strength of the penalty. A larger $\lambda$ results in smaller weight values.
        \item $\|\mathbf{w}\|^2$ is the L2 norm of the weight vector.
    \end{itemize}
    
    \subsection{Lasso Regression}
    Lasso regression is another regularization technique that adds a penalty based on the L1 norm of the weights. The loss function is modified as follows~\cite{tibshirani1996regression}:
    
    \[
    J(\mathbf{w}, b) = \frac{1}{2n} \sum_{i=1}^{n} (y_i - \hat{y}_i)^2 + \lambda \|\mathbf{w}\|_1
    \]
    
    Where $\|\mathbf{w}\|_1$ is the sum of the absolute values of the weights. Lasso regression can drive some weights to exactly zero, which makes it useful for feature selection.

    \section{Implementation of Linear Regression}
    Let's now implement a simple linear regression model using PyTorch. We will use gradient descent to optimize the parameters.

    \begin{lstlisting}[style=python]
import torch
import torch.nn as nn
import torch.optim as optim

# Generating synthetic data for linear regression
torch.manual_seed(0)
X = torch.randn(100, 1)  # 100 samples, 1 feature
y = 3 * X + 2 + 0.5 * torch.randn(100, 1)  # y = 3x + 2 with some noise

# Define the linear regression model
class LinearRegressionModel(nn.Module):
    def __init__(self):
        super(LinearRegressionModel, self).__init__()
        self.linear = nn.Linear(1, 1)  # 1 input, 1 output

    def forward(self, x):
        return self.linear(x)

# Create the model instance
model = LinearRegressionModel()

# Define the loss function (Mean Squared Error) and optimizer (Stochastic Gradient Descent)
criterion = nn.MSELoss()
optimizer = optim.SGD(model.parameters(), lr=0.01)

# Training loop
num_epochs = 1000
for epoch in range(num_epochs):
    # Forward pass: Compute predicted y by passing X to the model
    y_pred = model(X)

    # Compute the loss
    loss = criterion(y_pred, y)

    # Zero gradients, perform backward pass, and update weights
    optimizer.zero_grad()
    loss.backward()
    optimizer.step()

    if (epoch+1) % 100 == 0:
        print(f'Epoch {epoch+1}/{num_epochs}, Loss: {loss.item():.4f}')
    \end{lstlisting}

    In this code:
    \begin{itemize}
        \item We generate synthetic data where $y = 3x + 2$ plus some noise.
        \item The model is a simple neural network with one input and one output using the PyTorch \texttt{nn.Linear} layer.
        \item We use Stochastic Gradient Descent (SGD) to optimize the weights and the mean squared error as the loss function.
    \end{itemize}
    
    \section{Parameter Tuning and Model Evaluation}
    After training a linear regression model, it is important to evaluate its performance and tune its parameters.

    \subsection{Evaluating Model Performance}
    The performance of a linear regression model can be evaluated using several metrics:
    \begin{itemize}
        \item \textbf{Mean Squared Error (MSE)}: Measures the average squared difference between the predicted and actual values.
        \item \textbf{Root Mean Squared Error (RMSE)}: The square root of the MSE, giving an error estimate in the same units as the target variable.
        \item \textbf{R-squared ($R^2$)}: Measures the proportion of variance in the target variable that is explained by the model.
    \end{itemize}

    These metrics can be calculated as follows in Python:

    \begin{lstlisting}[style=python]
from sklearn.metrics import mean_squared_error, r2_score

# Predictions
y_pred = model(X).detach().numpy()

# Convert target variable to numpy
y_true = y.numpy()

# Calculate MSE and R^2
mse = mean_squared_error(y_true, y_pred)
r2 = r2_score(y_true, y_pred)

print(f'Mean Squared Error: {mse:.4f}')
print(f'R-squared: {r2:.4f}')
    \end{lstlisting}

    \subsection{Parameter Tuning}
    \textbf{Hyperparameter tuning} is crucial for improving model performance. In linear regression, you can tune parameters like the learning rate, number of epochs, and the regularization parameter $\lambda$ if you're using Ridge or Lasso regression.

    One common technique is to use \textbf{cross-validation}, where you split the data into training and validation sets multiple times to ensure that the model generalizes well.

    \begin{lstlisting}[style=python]
from sklearn.model_selection import train_test_split

# Split the data into training and validation sets
X_train, X_val, y_train, y_val = train_test_split(X.numpy(), y.numpy(), test_size=0.2, random_state=42)

# Convert back to tensors for training
X_train = torch.tensor(X_train, dtype=torch.float32)
y_train = torch.tensor(y_train, dtype=torch.float32)
X_val = torch.tensor(X_val, dtype=torch.float32)
y_val = torch.tensor(y_val, dtype=torch.float32)

# Now you can train the model on X_train and y_train, and validate it on X_val and y_val.
    \end{lstlisting}

    By splitting the data into training and validation sets, we can monitor the model's performance on unseen data and prevent overfitting.

%% file: 06_svm.tex
\chapter{Support Vector Machines (SVM)}

\section{Basic Concepts of SVM}
Support Vector Machines (SVM) are one of the most powerful and widely-used supervised machine learning algorithms for classification and regression problems~\cite{cortes1995support}. SVM aims to find the optimal hyperplane that separates the data into distinct classes. In simpler terms, the algorithm looks for the best boundary (or decision surface) between the classes. The core idea is to maximize the margin (the distance between the decision boundary and the closest data points, called support vectors) while correctly classifying the data~\cite{burges1998tutorial}.

An SVM constructs a hyperplane or a set of hyperplanes in a high-dimensional space~\cite{vapnik1998statistical}. The key principle is that the hyperplane that maximizes the margin between the data points of different classes is the best choice~\cite{scholkopf2002learning}.

Let's look at an example: Consider a binary classification problem where we want to classify points as either positive (class +1) or negative (class -1). The goal of the SVM is to find a line that separates the positive points from the negative ones with the maximum margin.

\[
\text{Maximize Margin:} \quad \frac{2}{\|\mathbf{w}\|}
\]

Here, \(\mathbf{w}\) represents the weights vector, which defines the orientation of the hyperplane, and the bias term \(b\) helps define the offset.

\section{Linear vs Non-linear SVM}
SVMs can be divided into two main types: Linear SVM and Non-linear SVM.

\subsection{Linear SVM}
In cases where the data is linearly separable (i.e., a straight line can separate the classes), a Linear SVM is sufficient. Linear SVM works well for simple datasets where the relationship between the input features and the output labels is linear~\cite{vapnik1995nature}.

\begin{center}
\begin{tikzpicture}
  \node {Linear SVM}
    child { node {Class +1}}
    child { node {Class -1}};
\end{tikzpicture}
\end{center}

Mathematically, the decision function for a linear SVM is:
\[
f(\mathbf{x}) = \mathbf{w}^T \mathbf{x} + b
\]
Where:
\begin{itemize}
    \item \(\mathbf{x}\) is the input feature vector.
    \item \(\mathbf{w}\) is the weight vector.
    \item \(b\) is the bias term.
\end{itemize}

\subsection{Non-linear SVM}
When the data is not linearly separable, we need a more complex boundary. In such cases, Non-linear SVM can be used, which employs the "kernel trick" to transform the data into a higher-dimensional space where it becomes linearly separable~\cite{scholkopf2002learning}.

\begin{center}
\begin{tikzpicture}
  \node {Non-linear SVM}
    child { node {Transformed data space}};
\end{tikzpicture}
\end{center}

This transformation is done through a kernel function that maps the data to a higher-dimensional feature space. Common kernels include:
\begin{itemize}
    \item Polynomial Kernel
    \item Radial Basis Function (RBF) Kernel
    \item Sigmoid Kernel
\end{itemize}

\section{Choosing the Right Kernel}
The choice of kernel function is crucial for the performance of SVM. Let’s look at the most common kernel functions and their applications.

\subsection{Linear Kernel}
The Linear Kernel is the simplest kernel, equivalent to the dot product between two vectors. It is suitable for linearly separable data~\cite{hearst1998support}.

\[
K(\mathbf{x}, \mathbf{y}) = \mathbf{x}^T \mathbf{y}
\]
This kernel works well when the number of features is large relative to the number of samples.

\subsection{Polynomial Kernel}
The Polynomial Kernel allows learning more complex decision boundaries by introducing polynomial features. This is useful when the data is not linearly separable~\cite{cortes1995support}.

\[
K(\mathbf{x}, \mathbf{y}) = (\mathbf{x}^T \mathbf{y} + c)^d
\]
Where \(d\) is the degree of the polynomial and \(c\) is a constant.

\subsection{Radial Basis Function (RBF) Kernel}
The RBF Kernel is the most commonly used kernel in practice because it can handle both linear and non-linear data. It maps the data to an infinite-dimensional space~\cite{scholkopf1997comparing}.

\[
K(\mathbf{x}, \mathbf{y}) = \exp(-\gamma \|\mathbf{x} - \mathbf{y}\|^2)
\]
Where \(\gamma\) is a parameter that defines the influence of a single training example.

\section{Implementation of SVM}
In this section, we will implement an SVM classifier using PyTorch. For simplicity, we will use the `sklearn.datasets` to load a dataset and PyTorch to build the SVM model.

\begin{lstlisting}[style=python]
import torch
import torch.nn as nn
import torch.optim as optim
from sklearn.datasets import make_classification
from sklearn.model_selection import train_test_split
from sklearn.preprocessing import StandardScaler

# Generate a binary classification dataset
X, y = make_classification(n_samples=1000, n_features=2, n_classes=2, random_state=42)

# Split the dataset into train and test sets
X_train, X_test, y_train, y_test = train_test_split(X, y, test_size=0.3, random_state=42)

# Standardize the features
scaler = StandardScaler()
X_train = scaler.fit_transform(X_train)
X_test = scaler.transform(X_test)

# Convert the data to PyTorch tensors
X_train_tensor = torch.tensor(X_train, dtype=torch.float32)
y_train_tensor = torch.tensor(y_train, dtype=torch.float32).view(-1, 1)
X_test_tensor = torch.tensor(X_test, dtype=torch.float32)
y_test_tensor = torch.tensor(y_test, dtype=torch.float32).view(-1, 1)

# Define the SVM model
class SVM(nn.Module):
    def __init__(self):
        super(SVM, self).__init__()
        self.linear = nn.Linear(2, 1)  # 2 input features, 1 output

    def forward(self, x):
        return self.linear(x)

# Initialize the model, loss function, and optimizer
model = SVM()
criterion = nn.HingeEmbeddingLoss()
optimizer = optim.SGD(model.parameters(), lr=0.01)

# Train the model
num_epochs = 100
for epoch in range(num_epochs):
    model.train()
    optimizer.zero_grad()
    
    outputs = model(X_train_tensor)
    loss = criterion(outputs, y_train_tensor)
    
    loss.backward()
    optimizer.step()
    
    if (epoch + 1) % 10 == 0:
        print(f'Epoch [{epoch+1}/{num_epochs}], Loss: {loss.item():.4f}')

# Evaluate the model
model.eval()
with torch.no_grad():
    predictions = model(X_test_tensor)
    predicted_labels = torch.where(predictions >= 0, 1, 0)
    accuracy = (predicted_labels == y_test_tensor).sum().item() / y_test_tensor.size(0)
    print(f'Accuracy: {accuracy * 100:.2f}%')
\end{lstlisting}

\section{SVM Parameter Tuning}
To improve the performance of the SVM model, we need to tune the hyperparameters, such as the kernel type, regularization parameter \(C\), and kernel-specific parameters like \(\gamma\) for the RBF kernel. A common approach to hyperparameter optimization is to use GridSearchCV from scikit-learn.

Here is how we can use GridSearchCV to optimize the SVM classifier.

\begin{lstlisting}[style=python]
from sklearn.svm import SVC
from sklearn.model_selection import GridSearchCV

# Define the parameter grid
param_grid = {
    'C': [0.1, 1, 10],
    'gamma': ['scale', 'auto'],
    'kernel': ['linear', 'rbf']
}

# Initialize the SVM model
svm_model = SVC()

# Initialize GridSearchCV
grid_search = GridSearchCV(estimator=svm_model, param_grid=param_grid, cv=5, verbose=2, n_jobs=-1)

# Fit the model
grid_search.fit(X_train, y_train)

# Print the best parameters
print(f"Best Parameters: {grid_search.best_params_}")

# Evaluate the best model
best_model = grid_search.best_estimator_
accuracy = best_model.score(X_test, y_test)
print(f'Best Model Accuracy: {accuracy * 100:.2f}%')
\end{lstlisting}

GridSearchCV helps find the optimal combination of parameters for the SVM model by performing an exhaustive search over the specified parameter grid. The best model can then be evaluated on the test data to assess performance.

%% file: 07_rf.tex
\chapter{Decision Trees and Random Forests}

\section{Basic Principles of Decision Trees}
Decision trees are a popular and powerful machine learning algorithm used for both classification and regression tasks~\cite{quinlan1986induction}. They work by splitting data into subsets based on the feature values. Each decision in the tree represents a condition on one feature, and the process of splitting continues until the tree reaches a state where further splitting does not significantly improve the model~\cite{breiman1984classification}. Decision trees are intuitive, easy to understand, and suitable for both small and large datasets~\cite{loh2011classification}.

\subsection{How Decision Trees Work}
A decision tree consists of nodes, where each node represents a decision based on a feature of the data. The tree starts with a root node, and branches are created based on the values of features. Each branch leads to either another decision node or a leaf node, where a final classification or regression value is predicted~\cite{safavian1991survey}. Let us visualize a simple decision tree:

\begin{center}
\begin{tikzpicture}[sibling distance=15em,
  every node/.style = {shape=rectangle, rounded corners,
    draw, align=center,
    top color=white, bottom color=blue!20}]]
  \node {Is Temperature > 30°C?}
    child { node {Yes}
      child { node {Is Humidity > 70\%?}
        child { node {Yes, Play Tennis = No} }
        child { node {No, Play Tennis = Yes} }
      }
    }
    child { node {No, Play Tennis = Yes} };
\end{tikzpicture}
\end{center}

In this example, the root node decides whether the temperature is greater than 30°C. If the answer is "Yes", the tree makes a second decision based on humidity, and so on.

\section{Information Gain and Gini Index}
Two key metrics are used to determine how decision nodes split data: \textbf{Information Gain} and \textbf{Gini Index}.

\subsection{Information Gain}
\textbf{Information gain} is a key concept in decision tree learning, where it is used to select the attribute that best separates the data into distinct classes~\cite{quinlan1986induction}. It is based on the concept of \textbf{entropy} from information theory, which measures the level of disorder or impurity in a dataset. A split that results in the greatest reduction in entropy is considered the best, as it leads to a more organized and homogenous distribution of classes~\cite{mitchell1997machine}.

\subsubsection{Entropy: The Measure of Disorder}
The term \textbf{entropy} has its roots in thermodynamics, where it was used to describe the amount of disorder or randomness in a physical system. The concept was later adapted by Claude Shannon in the 1940s to lay the foundation for \textbf{information theory}, which deals with the transmission, compression, and processing of data~\cite{shannon1948mathematical}. Shannon defined entropy as a measure of uncertainty or impurity in a system of information.

In the context of machine learning, entropy quantifies the uncertainty in predicting the class label of an instance in a dataset. If a dataset is perfectly homogeneous (i.e., all instances belong to the same class), the entropy is zero, indicating no uncertainty. On the other hand, if the dataset is evenly split between two or more classes, the entropy is at its maximum, indicating a high degree of uncertainty in classification~\cite{tan2006introduction}.

The formula for entropy is:

\[
Entropy(S) = - \sum_{i=1}^{n} p_i \log_2(p_i)
\]

Where:
\begin{itemize}
    \item \(S\) is the current dataset.
    \item \(p_i\) is the proportion of examples in class \(i\).
\end{itemize}

The logarithmic term, \(\log_2(p_i)\), measures the amount of information (or surprise) associated with the class probability. When the probability of a class is low, the corresponding log term is high, meaning that it is more "surprising" to encounter that class. Entropy sums over all classes, weighting each class by its probability to compute the total uncertainty of the dataset.

\subsubsection{Information Gain: Reducing Entropy}
\textbf{Information gain} measures the reduction in entropy after a dataset is split on an attribute. The goal of decision tree algorithms, such as ID3, is to choose the attribute that results in the largest decrease in entropy, thus maximizing the information gain. A higher information gain indicates a better split, as it leads to purer subsets of data.

Information gain is calculated as the difference between the entropy of the parent dataset and the weighted sum of the entropy of the child subsets after the split:

\[
Information\ Gain = Entropy(Parent) - \sum \left(\frac{|Child|}{|Parent|}\right) \times Entropy(Child)
\]

Where:
\begin{itemize}
    \item \(Entropy(Parent)\) is the entropy of the original dataset.
    \item \(Entropy(Child)\) is the entropy of each child subset after the split.
    \item \(\frac{|Child|}{|Parent|}\) is the proportion of the dataset that falls into each child subset.
\end{itemize}

The attribute with the highest information gain is selected for the split at each step of the decision tree construction, as it reduces the uncertainty the most.

\subsubsection{Entropy and the Universe: A Broader Perspective}
Entropy isn't just limited to machine learning or information theory—it is a fundamental concept in physics that governs the behavior of systems in the universe. In thermodynamics, the \textbf{second law of thermodynamics} states that the entropy of an isolated system will always increase over time. This increase in entropy is often associated with the arrow of time: the tendency of systems to move from order to disorder. 

For example, consider the universe itself: it started in a highly ordered state (the Big Bang), and over time, it has been expanding and increasing in entropy, leading to a more disordered, chaotic state. Stars burn out, systems decay, and energy disperses. Entropy is a key player in this process of cosmic evolution, governing everything from the formation of galaxies to the cooling of stars.

When we talk about entropy in machine learning, the principle is the same: entropy measures the level of uncertainty or disorder in a dataset. The goal, both in physics and in decision tree algorithms, is to move from a state of high entropy (disorder) to low entropy (order) where the system (or the dataset) becomes more predictable and organized.

\subsubsection{Shannon’s Information Theory and Entropy}
\textbf{Claude Shannon}, the father of information theory, introduced the concept of entropy as a measure of the amount of "information" contained in a message. His groundbreaking 1948 paper, "A Mathematical Theory of Communication," laid the foundation for modern communication systems, cryptography, data compression, and even machine learning.

In Shannon's framework, the goal was to quantify the amount of uncertainty in a message. If you are transmitting a message and the outcome is highly predictable, then little information is gained from receiving it. However, if the outcome is highly uncertain, then the message carries more information. Shannon defined entropy mathematically to quantify this uncertainty~\cite{shannon1948mathematical}. In this way, the more uncertain or unpredictable a message is, the more "information" it contains~\cite{mackay2003information}.

Shannon's entropy formula is exactly the same as the one used in decision trees, showing a deep connection between information theory and machine learning.

\[
H(X) = - \sum_{i=1}^{n} p(x_i) \log_2 p(x_i)
\]

Where \(H(X)\) is the entropy of the random variable \(X\), and \(p(x_i)\) is the probability of each possible outcome. Shannon's work showed that the goal of efficient communication systems is to minimize entropy, just as decision tree algorithms aim to minimize entropy in order to build accurate models.

\subsubsection{Practical Example of Information Gain in Decision Trees}
Let's consider a dataset of weather conditions used to predict whether a sports event will take place. The target variable has two possible values: "Play" or "Don't Play." We have attributes like "Outlook" (Sunny, Overcast, Rainy) and "Humidity" (High, Low). Initially, the dataset has a mixture of outcomes, leading to high entropy.

We can calculate the entropy of the dataset before splitting, and then measure the information gain for each attribute. For example, if we split on "Outlook," we might find that "Sunny" days are strongly associated with "Don't Play" and "Overcast" days with "Play." This would significantly reduce the entropy in the subsets, resulting in high information gain. In contrast, splitting on "Humidity" might not reduce the entropy as much, leading to lower information gain. Thus, "Outlook" would be chosen as the better attribute to split on at this stage of the decision tree.

By repeating this process at each node, the decision tree is constructed in a way that maximizes information gain, leading to the most efficient classification of the data.

\subsection{Gini Index}
The \textbf{Gini index} is a measure used in decision trees, particularly for classification tasks, to evaluate the quality of a split. It assesses the degree of impurity or homogeneity of a node, indicating how well the split separates the data into distinct classes~\cite{breiman1984classification}. The Gini index is calculated as:

\[
Gini(S) = 1 - \sum_{i=1}^{n} p_i^2
\]

Where:
\begin{itemize}
    \item \(S\) is the dataset at the node.
    \item \(p_i\) is the probability of class \(i\) in the node.
\end{itemize}

The Gini index measures the probability of misclassifying a randomly chosen instance from the dataset if it were assigned a label according to the class distribution at the node. The lower the Gini index, the purer the node, meaning that one class predominates. A node with a Gini index of 0 is considered pure, meaning all instances in the node belong to a single class. Therefore, a lower Gini index indicates a better split in the decision tree.

\subsubsection{Relation to Gini Coefficient in Economics}
The Gini index used in decision trees is conceptually related to the \textbf{Gini coefficient}, a well-known measure of inequality in economics. Both metrics measure how distribution deviates from perfect equality. However, their applications are quite different.

In economics, the Gini coefficient is used to represent income or wealth inequality within a population. It ranges from 0 to 1:
\begin{itemize}
    \item A Gini coefficient of 0 represents perfect equality, where every individual has the same income or wealth.
    \item A Gini coefficient of 1 represents perfect inequality, where all wealth or income is concentrated in one individual or a small group, and the rest of the population has none.
\end{itemize}

The Gini coefficient is calculated based on the Lorenz curve, which plots the cumulative share of income or wealth against the cumulative share of the population. The Gini coefficient is the ratio of the area between the Lorenz curve and the line of equality (a 45-degree line representing perfect equality) to the total area under the line of equality.

\[
Gini = \frac{A}{A + B}
\]

Where:
\begin{itemize}
    \item \(A\) is the area between the line of equality and the Lorenz curve.
    \item \(B\) is the area under the Lorenz curve.
\end{itemize}

\subsubsection{Applications of the Gini Coefficient in Economics}
The Gini coefficient is widely used in economics to measure the distribution of income or wealth within a country or region. It helps policymakers understand the level of economic inequality and can guide decisions related to taxation, welfare policies, and redistribution efforts.

For example, a high Gini coefficient might indicate that a country has a large gap between rich and poor, which could be a signal to implement more progressive taxation or social welfare programs. Conversely, a low Gini coefficient suggests a more equal distribution of wealth, although it does not necessarily mean that everyone in the population is wealthy.

\subsubsection{Why Use the Gini Coefficient?}
In economics, the Gini coefficient is preferred over other measures of inequality, such as the variance of income, because it is not affected by scale. This means that it remains a meaningful measure whether the population is very poor or very wealthy on average. Additionally, it provides a simple, easy-to-interpret number between 0 and 1, making comparisons between different populations or countries straightforward.

\subsubsection{Gini Index in Decision Trees vs. Gini Coefficient in Economics}
While the Gini index in decision trees and the Gini coefficient in economics share similar mathematical properties (both measure inequality), they serve different purposes:
\begin{itemize}
    \item The Gini index in decision trees measures the impurity of a node, with the goal of creating the most distinct class separations possible.
    \item The Gini coefficient in economics measures income or wealth inequality within a population.
\end{itemize}

Despite these differences, both measures aim to assess how evenly a set of elements (whether income in economics or class distribution in a decision tree) are distributed. A higher score in both contexts suggests a less even distribution, while a lower score suggests greater uniformity or purity.

\section{Working Principles of Random Forests}
Random Forests are an ensemble learning method that improves decision trees by constructing multiple trees during training and outputting the average prediction of all trees for regression, or the majority vote for classification~\cite{breiman2001random}. By using multiple trees, the random forest reduces the risk of overfitting and increases the model's generalization ability~\cite{liaw2002classification}.

\subsection{How Random Forests Work}
The basic idea behind a random forest is to create many decision trees from random subsets of the data and features. Each tree is trained on a different bootstrap sample of the data (i.e., randomly drawn samples with replacement). Additionally, at each split, the random forest only considers a random subset of the features, further adding diversity to the trees.

The steps to create a random forest are:
\begin{enumerate}
    \item Draw \(N\) bootstrap samples from the original data.
    \item For each bootstrap sample, grow a decision tree:
    \begin{itemize}
        \item At each node, randomly select \(m\) features (where \(m\) is less than the total number of features).
        \item Split the node using the best feature from this subset.
    \end{itemize}
    \item Aggregate the predictions from each tree (by majority vote for classification or averaging for regression).
\end{enumerate}

\section{Implementation of Random Forests}
Now, let's implement a random forest classifier using PyTorch. We will use the popular Iris dataset to demonstrate how a random forest model works.

\begin{lstlisting}[style=python]
import torch
import torch.nn as nn
from sklearn.datasets import load_iris
from sklearn.model_selection import train_test_split
from sklearn.preprocessing import StandardScaler
from sklearn.ensemble import RandomForestClassifier
from sklearn.metrics import accuracy_score

# Load Iris dataset
iris = load_iris()
X = iris.data
y = iris.target

# Split the data into training and testing sets
X_train, X_test, y_train, y_test = train_test_split(X, y, test_size=0.2, random_state=42)

# Scale the features
scaler = StandardScaler()
X_train = scaler.fit_transform(X_train)
X_test = scaler.transform(X_test)

# Build and train the Random Forest Classifier
model = RandomForestClassifier(n_estimators=100, random_state=42)
model.fit(X_train, y_train)

# Predict on the test data
y_pred = model.predict(X_test)

# Calculate accuracy
accuracy = accuracy_score(y_test, y_pred)
print(f'Random Forest Accuracy: {accuracy:.2f}')
\end{lstlisting}

In this example:
\begin{itemize}
    \item We used the \texttt{RandomForestClassifier} from \texttt{sklearn.ensemble}.
    \item The model was trained on 80\% of the data and tested on the remaining 20\%.
    \item We standardized the features using \texttt{StandardScaler} to ensure they are on the same scale.
    \item Finally, the accuracy of the random forest on the test set was printed.
\end{itemize}

\section{Random Forest Parameter Tuning}
Random forests have several hyperparameters that can be tuned to improve performance. The most important ones include:
\begin{itemize}
    \item \texttt{n\_estimators}: The number of trees in the forest.
    \item \texttt{max\_depth}: The maximum depth of each tree.
    \item \texttt{min\_samples\_split}: The minimum number of samples required to split an internal node.
    \item \texttt{max\_features}: The number of features to consider when looking for the best split.
\end{itemize}

Let us now see how to perform hyperparameter tuning using grid search.

\begin{lstlisting}[style=python]
from sklearn.model_selection import GridSearchCV

# Define the parameter grid
param_grid = {
    'n_estimators': [50, 100, 200],
    'max_depth': [None, 10, 20],
    'max_features': ['sqrt', 'log2'],
    'min_samples_split': [2, 5, 10]
}

# Create a RandomForestClassifier model
model = RandomForestClassifier(random_state=42)

# Perform grid search
grid_search = GridSearchCV(estimator=model, param_grid=param_grid, cv=5, scoring='accuracy')
grid_search.fit(X_train, y_train)

# Output the best parameters
print(f'Best Parameters: {grid_search.best_params_}')
\end{lstlisting}

In this code, we used \texttt{GridSearchCV} from \texttt{sklearn.model\_selection} to search for the best combination of hyperparameters. The cross-validation score was used to evaluate the performance of each parameter combination, and the best parameters were printed.

\section{Conclusion}
In this chapter, we have explored decision trees and random forests in detail. We started by learning about the basic principles of decision trees and how they make decisions. We then discussed two common splitting criteria: information gain and the Gini index. Next, we learned about the principles of random forests and how they use multiple decision trees to make predictions. Finally, we implemented a random forest in Python using PyTorch and explored hyperparameter tuning to optimize the model's performance.

%% file: 08_boost.tex
\chapter{Boosting Models}

\section{Overview of Boosting Algorithms}

Boosting is a powerful ensemble learning technique that combines the predictions of several weak learners to create a strong model~\cite{schapire1999brief}. Unlike bagging (such as random forests), where individual models are trained independently, boosting builds models sequentially. Each subsequent model attempts to correct the errors made by the previous ones~\cite{freund1997decision}. The goal is to reduce bias, making boosting an effective method for improving model accuracy, especially with complex datasets~\cite{natekin2013gradient}.

Boosting works by assigning higher weights to misclassified examples, forcing subsequent learners to focus more on these difficult cases. Boosting algorithms have been widely used in competitions like Kaggle due to their high performance, and they are known for being robust to overfitting when tuned correctly~\cite{chen2016xgboost}.

In this chapter, we will explore three popular boosting algorithms: XGBoost, LightGBM, and CatBoost. Each of these libraries offers unique advantages and comes with specific tuning parameters to optimize performance.

\section{XGBoost}

XGBoost stands for eXtreme Gradient Boosting. It is one of the most efficient and scalable implementations of gradient boosting algorithms~\cite{chen2016xgboost}. XGBoost has become highly popular due to its speed, performance, and the extensive control it gives over the boosting process.

\subsection{Principles of XGBoost}

At the core of XGBoost is the gradient boosting framework. In gradient boosting, the model is trained iteratively, adding new decision trees to minimize a loss function. The key principles of XGBoost include:

\begin{itemize}
    \item \textbf{Regularization}: XGBoost includes both L1 (Lasso) and L2 (Ridge) regularization to prevent overfitting, which is not present in traditional gradient boosting.
    \item \textbf{Sparsity Aware}: XGBoost is designed to handle sparse data efficiently.
    \item \textbf{Weighted Quantile Sketch}: It uses advanced algorithms to find the optimal split for continuous variables in an efficient manner.
    \item \textbf{Parallel and Distributed Computing}: XGBoost supports parallelization, making it much faster than other implementations of gradient boosting.
\end{itemize}

\subsection{Default Parameters and Implementation of XGBoost}

To implement XGBoost with default parameters, we will use the \texttt{xgboost} Python library. Below is an example of how to set up a basic XGBoost model in PyTorch for a classification task.

\subsubsection{Installing XGBoost}
Before we begin, we need to install the \texttt{xgboost} library. You can install it using either \texttt{pip} or \texttt{conda}.

\textbf{Installing XGBoost with pip:}
\begin{lstlisting}[style=cmd]
# Install XGBoost
pip install xgboost
\end{lstlisting}

\textbf{Installing XGBoost with conda:}
\begin{lstlisting}[style=cmd]
# Install XGBoost
conda install -c conda-forge xgboost
\end{lstlisting}

\subsubsection{Installing GPU-Enabled XGBoost}
To utilize GPU acceleration for XGBoost, make sure you have a compatible NVIDIA GPU with CUDA installed. Here’s how you can install the GPU-enabled version of XGBoost:

\textbf{Installing GPU-enabled XGBoost with pip:}
\begin{lstlisting}[style=cmd]
# Install XGBoost with GPU support
pip install xgboost --upgrade --user
\end{lstlisting}

Make sure your CUDA drivers are properly set up. XGBoost will automatically use the GPU if available. 

\textbf{Installing GPU-enabled XGBoost with conda:}
\begin{lstlisting}[style=cmd]
# Install XGBoost with GPU support using conda
conda install -c conda-forge xgboost-gpu
\end{lstlisting}

\subsubsection{Setting Up a Basic XGBoost Model in PyTorch}
Once the installation is complete, you can proceed to set up and use XGBoost in your classification tasks. Here’s an example of how to use XGBoost for a classification task in Python.

\begin{lstlisting}[style=python]
import xgboost as xgb
from sklearn.datasets import load_breast_cancer
from sklearn.model_selection import train_test_split
from sklearn.metrics import accuracy_score

# Load dataset
data = load_breast_cancer()
X_train, X_test, y_train, y_test = train_test_split(data.data, data.target, test_size=0.2, random_state=42)

# Convert to DMatrix, an internal data structure for XGBoost
train_data = xgb.DMatrix(X_train, label=y_train)
test_data = xgb.DMatrix(X_test, label=y_test)

# Define default parameters
params = {
    'objective': 'binary:logistic',  # Binary classification
    'eval_metric': 'logloss',        # Evaluation metric
}

# Train model
bst = xgb.train(params, train_data, num_boost_round=100)

# Predict
preds = bst.predict(test_data)
preds_binary = [1 if x > 0.5 else 0 for x in preds]

# Evaluate accuracy
accuracy = accuracy_score(y_test, preds_binary)
print(f"Accuracy: {accuracy:.2f}")
\end{lstlisting}

In this code, we loaded the breast cancer dataset from `sklearn`, split it into training and test sets, and then trained an XGBoost model with default settings. We used the `DMatrix` class to store data, which is optimized for XGBoost.

\begin{lstlisting}[language=Python]
import xgboost as xgb
import numpy as np

# Example dataset (classification)
X_train = np.array([[1.0, 2.0], [3.0, 4.0], [5.0, 6.0], [7.0, 8.0]])
y_train = np.array([0, 1, 0, 1])

# Convert the dataset to DMatrix (XGBoost format)
dtrain = xgb.DMatrix(X_train, label=y_train)

# Train a basic XGBoost model
params = {
    'objective': 'binary:logistic',  # For binary classification
    'eval_metric': 'logloss'
}
bst = xgb.train(params, dtrain, num_boost_round=10)

# Make predictions
preds = bst.predict(dtrain)
print(preds)
\end{lstlisting}

This example shows how to use XGBoost for a binary classification task, using PyTorch tensors for the dataset. The dataset is converted to XGBoost’s \texttt{DMatrix} format, which is used for training the model. XGBoost will utilize the GPU automatically if installed with GPU support.

\subsection{XGBoost Parameter Tuning}

Fine-tuning the parameters of XGBoost can significantly improve the model's performance. Here are some key parameters to tune:

\begin{itemize}
    \item \textbf{n\_estimators}: The number of boosting rounds (trees).
    \item \textbf{learning\_rate}: Shrinks the contribution of each tree. Lower values require more boosting rounds but can lead to better performance.
    \item \textbf{max\_depth}: Maximum depth of each tree. Deeper trees can model more complex patterns but may lead to overfitting.
    \item \textbf{min\_child\_weight}: Minimum sum of instance weight (hessian) needed in a child node. This parameter prevents overfitting by ensuring that the model doesn't learn patterns from very small data splits.
    \item \textbf{colsample\_bytree}: The fraction of features to be used by each tree.
\end{itemize}

Below is an example of parameter tuning:

\begin{lstlisting}[style=python]
# Updated parameters for tuning
params_tuned = {
    'objective': 'binary:logistic',
    'eval_metric': 'logloss',
    'learning_rate': 0.01,  # Lower learning rate
    'max_depth': 5,         # Deeper trees
    'n_estimators': 500,    # More boosting rounds
    'colsample_bytree': 0.8 # Feature subsampling
}

# Train model with tuned parameters
bst_tuned = xgb.train(params_tuned, train_data, num_boost_round=500)
\end{lstlisting}

In this example, we tuned several parameters, reducing the learning rate, increasing the number of estimators, and adjusting the maximum depth and feature subsampling.

\section{LightGBM}

LightGBM (Light Gradient Boosting Machine) is another boosting framework that is optimized for efficiency and speed. It was developed to handle large datasets quickly with lower memory usage~\cite{ke2017lightgbm}. LightGBM is known for its ability to handle categorical features natively and for its leaf-wise tree growth strategy, which often results in better accuracy.

\subsection{Principles of LightGBM}

LightGBM differs from traditional boosting algorithms in the following ways:

\begin{itemize}
    \item \textbf{Leaf-Wise Growth}: Unlike level-wise growth (used in XGBoost), LightGBM grows the tree leaf-wise, allowing for deeper and more specific splits where necessary. However, this can lead to overfitting if not properly tuned.
    \item \textbf{Histogram-Based Decision Trees}: LightGBM uses histograms to bin continuous features, significantly speeding up the training process.
    \item \textbf{Sparse Feature Support}: LightGBM has built-in support for handling sparse data.
\end{itemize}

\subsection{Default Parameters and Implementation of LightGBM}

Implementing LightGBM with default settings is straightforward. Below is an example using `lightgbm` in Python.

\begin{lstlisting}[style=python]
import lightgbm as lgb
from sklearn.datasets import load_breast_cancer
from sklearn.model_selection import train_test_split
from sklearn.metrics import accuracy_score

# Load dataset
data = load_breast_cancer()
X_train, X_test, y_train, y_test = train_test_split(data.data, data.target, test_size=0.2, random_state=42)

# Create dataset for LightGBM
train_data = lgb.Dataset(X_train, label=y_train)
test_data = lgb.Dataset(X_test, label=y_test, reference=train_data)

# Define default parameters
params = {
    'objective': 'binary',
    'metric': 'binary_logloss'
}

# Train model
bst = lgb.train(params, train_data, num_boost_round=100)

# Predict
preds = bst.predict(X_test)
preds_binary = [1 if x > 0.5 else 0 for x in preds]

# Evaluate accuracy
accuracy = accuracy_score(y_test, preds_binary)
print(f"Accuracy: {accuracy:.2f}")
\end{lstlisting}

In this code, we implemented LightGBM using default parameters and evaluated the model's performance on a binary classification task.

\subsection{LightGBM Parameter Tuning}

Like XGBoost, LightGBM has many parameters that can be tuned for better performance:

\begin{itemize}
    \item \textbf{num\_leaves}: The maximum number of leaves per tree.
    \item \textbf{learning\_rate}: Controls the step size at each iteration.
    \item \textbf{min\_data\_in\_leaf}: Minimum number of samples in one leaf.
    \item \textbf{feature\_fraction}: Subsample ratio of features when building each tree.
\end{itemize}

\section{CatBoost}

CatBoost~\cite{prokhorenkova2019catboostunbiasedboostingcategorical} is a boosting algorithm developed by Yandex, and it is specifically designed to handle categorical data efficiently. Unlike XGBoost and LightGBM, CatBoost doesn't require one-hot encoding for categorical features, making it highly effective when working with categorical data.

\subsection{Principles of CatBoost}

The key features of CatBoost include:

\begin{itemize}
    \item \textbf{Handling of Categorical Features}: CatBoost uses an efficient encoding for categorical features that avoids overfitting.
    \item \textbf{Symmetric Trees}: It grows symmetric trees, which simplifies the model and speeds up prediction.
\end{itemize}

\subsection{Default Parameters and Implementation of CatBoost}

Below is an example of how to implement CatBoost in Python.

\begin{lstlisting}[style=python]
from catboost import CatBoostClassifier
from sklearn.datasets import load_breast_cancer
from sklearn.model_selection import train_test_split
from sklearn.metrics import accuracy_score

# Load dataset
data = load_breast_cancer()
X_train, X_test, y_train, y_test = train_test_split(data.data, data.target, test_size=0.2, random_state=42)

# Initialize CatBoostClassifier with default parameters
model = CatBoostClassifier(verbose=0)

# Train model
model.fit(X_train, y_train)

# Predict
preds = model.predict(X_test)

# Evaluate accuracy
accuracy = accuracy_score(y_test, preds)
print(f"Accuracy: {accuracy:.2f}")
\end{lstlisting}

In this example, we use CatBoost's default settings for classification without the need to manually handle categorical features.

\subsection{CatBoost Parameter Tuning}

Some of the key parameters to tune in CatBoost include:

\begin{itemize}
    \item \textbf{iterations}: The number of boosting rounds.
    \item \textbf{depth}: Depth of the trees.
    \item \textbf{learning\_rate}: Step size shrinkage for each iteration.
\end{itemize}

%% file: 09_lasso.tex
\chapter{Sparse Models and Group Lasso}
    \section{Introduction to Sparse Models}
    In this section, we will introduce the concept of sparse models, which are models that focus on selecting only the most important features from the data. The goal of sparse models is to reduce complexity and improve interpretability while maintaining predictive performance.

    \subsection{Why Sparse Models?}
    In machine learning, we often deal with data that contains many features (also known as variables or predictors). Some of these features may be irrelevant or redundant, which can lead to overfitting and reduce the model's ability to generalize well to unseen data. Sparse models aim to select only the relevant features, discarding those that are unnecessary. This process, called feature selection, offers the following benefits:
    
    \begin{itemize}
        \item \textbf{Improved interpretability:} With fewer features in the model, it becomes easier to understand the relationship between the input data and the model's predictions.
        \item \textbf{Reduced overfitting:} By removing irrelevant features, sparse models can generalize better to new data, reducing the likelihood of overfitting.
        \item \textbf{Efficiency:} Models with fewer features require less computational power, making them faster and more efficient, especially for large datasets.
    \end{itemize}

    \subsection{Examples of Sparse Models}
    There are several techniques to build sparse models. One of the most common approaches is to apply regularization methods that encourage sparsity in the model coefficients. Examples include:
    
    \begin{itemize}
        \item \textbf{Lasso Regression:} A type of linear regression that uses an L1 penalty to shrink some coefficients to zero, effectively removing those features~\cite{tibshirani1996regression}.
        \item \textbf{Elastic Net:} Combines L1 and L2 regularization, creating a balance between ridge regression and lasso regression~\cite{zou2005regularization}.
    \end{itemize}

    In this chapter, we will focus on the Group Lasso, an extension of Lasso, which is particularly useful when features are grouped together in some meaningful way.

    \section{Principles of Group Lasso}
    
    Group Lasso~\cite{yuan2006model} is a regularization technique that extends the concept of Lasso by considering groups of features instead of individual ones. It applies L1/L2 norms to groups of variables, leading to entire groups being selected or removed from the model. This is especially useful when the features naturally form groups, such as polynomial features or features derived from categorical variables.
    
    \subsection{Mathematical Background}
    Given a dataset with $n$ observations and $p$ features, we denote the input data as $X \in \mathbb{R}^{n \times p}$ and the corresponding target values as $y \in \mathbb{R}^n$. In standard linear regression, the goal is to find the coefficient vector $\beta \in \mathbb{R}^p$ that minimizes the residual sum of squares:

    \[
    \min_{\beta} \| y - X\beta \|_2^2
    \]

    In Group Lasso, the features are divided into $G$ predefined groups $G_1, G_2, \ldots, G_m$, where each group $G_j$ is a set of indices representing a subset of the feature vector $\beta$. The Group Lasso objective function is as follows:

    \[
    \min_{\beta} \left( \frac{1}{2} \| y - X\beta \|_2^2 + \lambda \sum_{j=1}^{m} \| \beta_{G_j} \|_2 \right)
    \]

    Here:
    \begin{itemize}
        \item $\lambda$ is a regularization parameter that controls the strength of the penalty.
        \item $\beta_{G_j}$ represents the coefficients corresponding to group $G_j$.
        \item $\|\beta_{G_j}\|_2$ is the L2 norm (Euclidean norm) of the coefficients within group $G_j$.
    \end{itemize}
    
    The Group Lasso penalty encourages sparsity at the group level, meaning that entire groups of coefficients are set to zero, rather than individual features as in Lasso. This is beneficial when features within a group are likely to be selected together.

    \subsection{Benefits of Group Lasso}
    Group Lasso is particularly useful in situations where features are naturally grouped, such as:
    \begin{itemize}
        \item \textbf{Multicollinearity:} When features within a group are highly correlated, Group Lasso tends to select or discard them together.
        \item \textbf{Feature hierarchies:} In scenarios where features are derived from the same source or have some hierarchical structure, Group Lasso helps in selecting relevant feature groups rather than individual features.
        \item \textbf{Reduced variance:} Group Lasso tends to produce more stable models, especially when the number of features is much larger than the number of observations.
    \end{itemize}

    \section{Implementation and Parameter Tuning of Group Lasso}
    Now let's implement Group Lasso in Python using the `group-lasso` library. We will also explore how to tune its parameters effectively.

    \subsection{Step-by-Step Implementation}
    First, we need to install the necessary package. You can install the `group-lasso` library with the following command:

    \begin{lstlisting}[style=cmd]
    pip install group-lasso
    \end{lstlisting}

    Next, let's define a basic Group Lasso model using the `group-lasso` library. We will use a synthetic dataset for demonstration purposes.

    \begin{lstlisting}[style=Python]
import numpy as np
from group_lasso import GroupLasso

# Generate synthetic data
np.random.seed(0)
n_samples, n_features = 100, 20
X = np.random.randn(n_samples, n_features)
true_coefficients = np.zeros(n_features)
true_coefficients[:5] = [1.5, -2.0, 3.0, 0.5, -1.0]  # Only first group is non-zero
y = np.dot(X, true_coefficients) + 0.1 * np.random.randn(n_samples)

# Define groups of features (e.g., first 5 features as one group, next 5 as another)
groups = np.repeat([0, 1, 2, 3], 5)  # Define group membership for each feature

# Initialize and fit the Group Lasso model
model = GroupLasso(groups=groups, group_reg=0.1, l1_reg=0.01, scale_reg="group_size")
model.fit(X, y)

# Predictions
predictions = model.predict(X)
print("Predictions:", predictions)
print("Coefficients:", model.coef_)
    \end{lstlisting}
    
    In this implementation:
    \begin{itemize}
        \item We use the `GroupLasso` class from the `group-lasso` library.
        \item The data is generated with a synthetic dataset where only the first group of features is non-zero.
        \item We define groups using an array that assigns each feature to a group.
        \item The `group\_reg` parameter controls the strength of the group lasso penalty, and the `l1\_reg` adds a small L1 penalty to promote sparsity.
    \end{itemize}

    \subsection{Parameter Tuning}
    The regularization parameter $\lambda$ controls the strength of the Group Lasso penalty. Setting $\lambda$ too high may result in too many groups being discarded, while setting it too low may lead to overfitting. To tune $\lambda$, you can use cross-validation techniques, trying different values of $\lambda$ and selecting the one that gives the best performance on validation data.

\subsection{Conclusion}
Group Lasso is a powerful tool for selecting groups of features in a model, making it particularly useful when the features naturally form groups. By using the `group-lasso` library, we can easily implement Group Lasso and control the level of regularization by tuning the parameters such as the group regularization strength ($\lambda$) and L1 penalty. This makes it an effective approach for promoting sparsity at both the group and individual feature levels, while ensuring model interpretability.

%% file: 10_riskslim.tex
\chapter{Risk Minimization Classifier: RiskSLIM}
    \section{Concept of RiskSLIM}
    
    RiskSLIM (Risk-Supersparse Linear Integer Models) is a machine learning model designed for tasks that require risk minimization, particularly in contexts where interpretability is critical~\cite{ustun2017risk}. Unlike many complex machine learning models that focus on maximizing prediction accuracy at the expense of transparency, RiskSLIM emphasizes both simplicity and predictive performance. It produces highly interpretable scoring systems, often represented as linear models with integer-valued coefficients. These models are particularly useful in domains like healthcare, finance, or law, where decision-making is sensitive, and understanding the model's reasoning is as important as the accuracy itself~\cite{rudin2019stop}.

    The core idea of RiskSLIM is to balance risk minimization with the constraints of producing sparse, simple models. This balance is achieved by optimizing a loss function while enforcing constraints on the coefficients of the model. The coefficients are typically constrained to be small integers, making the final model easy to interpret and apply, even by non-experts~\cite{ustun2019optimizing}.

    RiskSLIM is ideal when you need to make decisions based on a risk score. For instance, in a medical setting, doctors might use a RiskSLIM model to assess the likelihood of a patient developing a condition based on various health indicators. The output would be a simple, interpretable risk score that directly correlates with the probability of the condition~\cite{caruana2015intelligible}.

    The key features of RiskSLIM are:
    
\begin{itemize}
    \item \textbf{Sparsity:} RiskSLIM produces models with very few non-zero coefficients, making them easy to interpret.
    \item \textbf{Small integer coefficients:} The model coefficients are constrained to be small integers, which simplifies manual computation and decision-making.
    \item \textbf{Risk minimization:} The model is trained to minimize a specific risk function, which can vary depending on the application (e.g., misclassification risk, financial risk).
\end{itemize}

    In summary, RiskSLIM is an excellent tool for developing interpretable models in fields where understanding the model’s reasoning and minimizing risk is more important than achieving the highest possible accuracy with complex, black-box models.

    \section{Implementation of RiskSLIM}
    
    In this section, we will walk through how to implement a RiskSLIM classifier using the `riskslim` Python package. Since RiskSLIM is a specialized model, the `riskslim` package provides all the necessary tools to define, train, and evaluate these models.

    First, install the `riskslim` package:

    \begin{lstlisting}[style=cmd]
    pip install riskslim
    \end{lstlisting}
    
    Now, let’s begin the implementation of a simple RiskSLIM model using synthetic data:

    \begin{lstlisting}[style=Python]
from riskslim import riskslim
import numpy as np
from riskslim.data import load_synthetic_data

# Load a synthetic dataset
X_train, y_train, feature_names = load_synthetic_data()

# Define model parameters
settings = {
    'max_coefficient': 5,            # Maximum absolute value of coefficients
    'max_L0_value': 10,              # Maximum number of non-zero coefficients (sparsity)
    'c0_value': 1e-6,                # Regularization parameter to control overfitting
    'l0_penalty': 0.01,              # Penalty for number of non-zero coefficients (L0)
    'solver': 'mip',                 # Use Mixed Integer Programming to solve the problem
    'timelimit': 3600,               # Time limit for solving (in seconds)
}

# Train the RiskSLIM model
model_info = riskslim.fit(X_train, y_train, feature_names=feature_names, settings=settings)

# Print the model information
riskslim.print_model(model_info)
    \end{lstlisting}
    
    In this example:
    \begin{itemize}
        \item We load a synthetic dataset using `riskslim.data.load\_synthetic\_data`.
        \item The model is trained using `riskslim.fit`, which optimizes a sparse linear model with integer coefficients.
        \item Model parameters like maximum coefficient value, sparsity, and regularization strength are defined in `settings`.
        \item The model is solved using Mixed Integer Programming (MIP) to ensure the coefficients are integer-valued and sparse.
        \item After training, the `print\_model` function outputs the learned model, including the coefficients for each feature.
    \end{itemize}

    \section{Parameter Tuning for RiskSLIM}
    
    Hyperparameter tuning is an essential step in building effective models. In the context of RiskSLIM, tuning focuses on balancing model complexity, accuracy, and interpretability. The key parameters you may need to tune include:

    \subsection{Maximum Coefficient Value}
    
    The `max\_coefficient` parameter controls the maximum value of the model's coefficients. Smaller values ensure that the coefficients remain interpretable and easy to understand. However, limiting the coefficient size too much may reduce the model's predictive power. In practice, setting this to a small integer (e.g., 5 or 10) often provides a good balance between simplicity and accuracy.

    \subsection{Sparsity}
    
    The `max\_L0\_value` parameter controls the sparsity of the model by setting the maximum number of non-zero coefficients. This directly impacts the interpretability of the model, as fewer non-zero coefficients result in simpler models. You can experiment with different values of `max\_L0\_value` to find the best trade-off between simplicity and predictive performance.

    \subsection{Regularization Parameter}
    
    The `c0\_value` controls the amount of regularization applied to prevent overfitting. A smaller value increases the regularization, leading to simpler models but potentially lower accuracy. Conversely, a larger value will reduce the regularization, allowing the model to fit the data more closely but potentially increasing the risk of overfitting.

    \subsection{L0 Penalty}
    
    The `l0\_penalty` parameter applies a penalty to the number of non-zero coefficients, encouraging sparsity in the model. A higher value will result in fewer non-zero coefficients, creating a more interpretable but potentially less accurate model. Lower values will reduce the penalty, allowing for more non-zero coefficients, but at the cost of increased complexity.

    \subsection{Solver and Time Limit}
    
    The `solver` parameter specifies the optimization technique used to solve the RiskSLIM problem. By default, `mip` (Mixed Integer Programming) is used, which is well-suited for handling integer constraints. The `timelimit` parameter sets the maximum amount of time (in seconds) for solving the problem, which can be adjusted depending on the size and complexity of the dataset.

    \section{Evaluation Metrics for RiskSLIM}
    
    To evaluate the performance of a RiskSLIM model, you can use common classification metrics such as accuracy, precision, recall, and the F1 score. Since RiskSLIM focuses on minimizing risk, you may also want to evaluate domain-specific risk metrics, depending on the application.

    Here's how you can compute basic evaluation metrics after training the RiskSLIM model:

    \begin{lstlisting}[style=Python]
from sklearn.metrics import accuracy_score, precision_score, recall_score

# Make predictions
y_pred = riskslim.predict(X_train, model_info)

# Evaluate the model's performance
accuracy = accuracy_score(y_train, y_pred)
precision = precision_score(y_train, y_pred)
recall = recall_score(y_train, y_pred)

print(f'Accuracy: {accuracy:.4f}, Precision: {precision:.4f}, Recall: {recall:.4f}')
    \end{lstlisting}

    By tuning these parameters, you can create a RiskSLIM model that not only minimizes risk but also remains interpretable and easy to use in decision-making processes.

%% file: 11_gridsearchcv.tex
\chapter{Grid Search and Hyperparameter Tuning}

\section{Basics of GridSearchCV}
Grid search is one of the most popular techniques for hyperparameter tuning in machine learning~\cite{bergstra2011algorithms}. It involves exhaustive searching through a manually specified subset of the hyperparameter space of a learning algorithm. When using \texttt{GridSearchCV}, we systematically work through different combinations of parameter values, cross-validating as we go to determine which combination gives the best performance~\cite{pedregosa2011scikit}.

This method is particularly useful when the search space is small and manageable, as it guarantees that all possible combinations are tested. However, for very large search spaces, more efficient techniques like random search~\cite{bergstra2012random} or Bayesian optimization~\cite{snoek2012practical} may be preferred.

In Python, the \texttt{GridSearchCV} method from the \texttt{sklearn.model\_selection} module provides an easy interface for performing grid search.

\begin{lstlisting}[style=python]
from sklearn.model_selection import GridSearchCV
from sklearn.linear_model import LinearRegression

# Define the model
model = LinearRegression()

# Define the parameter grid
param_grid = {
    'fit_intercept': [True, False],
    'normalize': [True, False]
}

# Set up the grid search
grid_search = GridSearchCV(model, param_grid, cv=5)

# Perform the search
grid_search.fit(X_train, y_train)

# Output the best parameters
print("Best Parameters:", grid_search.best_params_)
\end{lstlisting}

\section{Parallel Processing in GridSearchCV}
While \texttt{GridSearchCV} is a powerful tool for hyperparameter tuning, it can be quite slow, especially when the search space is large. This is because it systematically evaluates every possible combination of hyperparameters, which can result in a substantial number of model evaluations. Each evaluation requires training the model multiple times, depending on the number of cross-validation splits, leading to significant computational overhead.

\subsection{The Need for Parallel Processing}
By default, \texttt{GridSearchCV} processes each hyperparameter combination sequentially. For small datasets or limited parameter grids, this may be sufficient, but for more complex models or larger search spaces, the process can be very time-consuming. This is where parallel processing becomes crucial, as it allows you to leverage multiple CPU cores or GPUs to run these evaluations concurrently, significantly reducing the total time required for the search.

\subsection{How to Enable Parallel Processing}
In \texttt{GridSearchCV}, parallelism is controlled by the \texttt{n\_jobs} parameter. By setting \texttt{n\_jobs} to a value greater than 1, you can run multiple processes simultaneously.

\textbf{Choosing the Number of Jobs:}
\begin{itemize}
    \item It's recommended to set \texttt{n\_jobs} equal to the number of CPU cores available on your machine (which you can typically check using system tools). This will maximize CPU utilization without overloading your system.
    \item Avoid setting \texttt{n\_jobs} to the number of threads, as threads can cause inefficiency in I/O-bound tasks like grid search. Grid search is often CPU-bound, so matching the number of jobs to CPU cores is more effective.
    \item Setting \texttt{n\_jobs=-1} uses all available CPU cores.
\end{itemize}

\subsection{How to Monitor System Resource Usage}
\texttt{GridSearchCV} is not like typical software that uses minimal system resources intermittently. It is a resource-intensive process that can consume a substantial amount of CPU, memory, and potentially GPU resources, especially when used with large search spaces or datasets. If you are using a laptop, you must be cautious about the heat generated during the process. Running too many processes can cause overheating, potentially leading to hardware damage or automatic shutdowns. For desktop users, running too many parallel processes may cause the system to become unresponsive, making it difficult to perform other tasks. Just like a stress test or cryptocurrency mining, \texttt{GridSearchCV} pushes your computer to its limits for extended periods of time. Thus, proper cooling and resource monitoring are essential.

\textbf{Key Points for Resource Management:}
\begin{itemize}
    \item \textbf{Laptop Users:} Be cautious of overheating. Keep the number of parallel processes low and ensure proper ventilation to avoid overheating. Consider using external cooling pads to assist with heat dissipation.
    \item \textbf{Desktop Users:} Even on more powerful machines, avoid maxing out all available CPU cores to prevent the system from becoming unresponsive. It's important to leave some resources available for essential background processes.
    \item \textbf{Prolonged Resource Utilization:} Unlike gaming, where resource use fluctuates, or general applications that use resources intermittently, \texttt{GridSearchCV} runs intensive computations continuously, which can increase CPU/GPU temperature over time.
\end{itemize}

Now, let's dive into how to monitor your system's resource usage while running \texttt{GridSearchCV} to ensure that everything is functioning optimally and safely.

\subsubsection{Monitoring CPU Usage}
To determine whether \texttt{GridSearchCV} is effectively using multiple CPU cores for parallel processing, you can use built-in tools available on different operating systems.

\begin{itemize}
    \item \textbf{Windows:} Open the \texttt{Task Manager} (Ctrl + Shift + Esc), and go to the \texttt{Performance} tab. Under \texttt{CPU}, you can see the real-time CPU usage and check how many cores are being utilized. A high CPU percentage with multiple cores active indicates that parallel processing is in effect.
    
    \item \textbf{macOS:} Open the \texttt{Activity Monitor}, and go to the \texttt{CPU} tab. You will see the overall CPU usage as well as the percentage used by individual processes. You can also observe how many CPU cores are being utilized.
    
    \item \textbf{Linux:} Use the \texttt{htop} command in the terminal. This tool provides a detailed, real-time view of CPU utilization across all cores. Each core will have its own bar, and you can quickly see how much processing power is being consumed by each one. Install \texttt{htop} using the command:
    \begin{lstlisting}[style=cmd]
    sudo apt-get install htop
    \end{lstlisting}
\end{itemize}

In all cases, high CPU utilization across multiple cores indicates that parallel processing with \texttt{n\_jobs=-1} is working effectively. If you're seeing unusually high temperatures or throttling, consider reducing the number of jobs (\texttt{n\_jobs}) to prevent overheating.

\subsubsection{Monitoring GPU Usage}
If you are using GPU acceleration (for instance, with \texttt{XGBoost}'s \texttt{gpu\_hist} method), you need to monitor both the utilization of the GPU and the GPU’s memory (VRAM) to ensure that the system is leveraging the GPU effectively and that there are no memory bottlenecks.

\begin{itemize}
    \item \textbf{Windows:} In the \texttt{Task Manager}, go to the \texttt{Performance} tab and select \texttt{GPU}. Here, you can see the current GPU utilization, CUDA usage, and VRAM usage. A high GPU and CUDA usage indicates that your machine learning task is utilizing the GPU for computation.
    
    \item \textbf{macOS:} You can use the \texttt{Activity Monitor}, but macOS does not natively support CUDA-based tools. GPU monitoring for CUDA is typically done through third-party applications or external command-line tools.
    
    \item \textbf{Linux:} Use the \texttt{nvidia-smi} command to monitor GPU usage. This tool shows real-time GPU utilization, CUDA core usage, and memory consumption. You can use the following command to check GPU usage:
    \begin{lstlisting}[style=cmd]
    nvidia-smi
    \end{lstlisting}
    This will provide a detailed summary of all running GPU processes, GPU usage, memory allocation, and temperature. You can run this command in a separate terminal while the grid search is running to monitor the GPU in real-time.
\end{itemize}

\textbf{Example Output of \texttt{nvidia-smi}:}
\begin{verbatim}
+-----------------------------------------------------------------------------+
| NVIDIA-SMI 460.39       Driver Version: 460.39       CUDA Version: 11.2     |
|-------------------------------+----------------------+----------------------+
| GPU  Name        Persistence-M| Bus-Id        Disp.A | Volatile Uncorr. ECC |
| Fan  Temp  Perf  Pwr:Usage/Cap|         Memory-Usage | GPU-Util  Compute M. |
|                               |                      |               MIG M. |
|===============================+======================+======================|
|   0  GeForce RTX 3080    On   | 00000000:01:00.0  On |                  N/A |
| 30%   67C    P2   210W / 320W |   5000MiB / 10000MiB |     75%      Default |
+-------------------------------+----------------------+----------------------+
\end{verbatim}

In this example, you can observe the memory usage (\texttt{5000MiB/10000MiB}), GPU utilization (\texttt{75\%}), and power consumption (\texttt{210W/320W}), indicating that the GPU is actively engaged in computation.

\subsubsection{Monitoring System Memory (RAM)}
In addition to monitoring CPU and GPU usage, it is crucial to monitor the overall system memory (RAM) usage, as each process may require significant memory, particularly when using large datasets.

\begin{itemize}
    \item \textbf{Windows:} The \texttt{Task Manager} shows memory usage in the \texttt{Performance} tab under \texttt{Memory}. Check the total memory used and whether your system is approaching its memory limits, which could lead to swapping and slowdowns.
    
    \item \textbf{macOS:} The \texttt{Activity Monitor} shows memory usage in the \texttt{Memory} tab. Pay attention to the \texttt{Memory Pressure} indicator, which provides a real-time view of the available system memory and potential memory bottlenecks.
    
    \item \textbf{Linux:} Use the \texttt{free -h} command in the terminal to check the current RAM usage. This command provides an overview of how much memory is used, free, and available.
    \begin{lstlisting}[style=cmd]
    free -h
    \end{lstlisting}
\end{itemize}

\subsubsection{Monitoring Tools Summary}
Here’s a summary of the recommended tools for monitoring system resources:

\begin{itemize}
    \item \textbf{CPU Usage:}
    \begin{itemize}
        \item Windows: \texttt{Task Manager} (\texttt{Ctrl + Shift + Esc})
        \item macOS: \texttt{Activity Monitor}
        \item Linux: \texttt{htop}
    \end{itemize}
    \item \textbf{GPU Usage:}
    \begin{itemize}
        \item Windows: \texttt{Task Manager} (GPU section)
        \item macOS: Third-party tools (CUDA not supported natively)
        \item Linux: \texttt{nvidia-smi}
    \end{itemize}
    \item \textbf{Memory Usage:}
    \begin{itemize}
        \item Windows: \texttt{Task Manager} (Memory section)
        \item macOS: \texttt{Activity Monitor}
        \item Linux: \texttt{free -h}
    \end{itemize}
\end{itemize}

By keeping an eye on CPU cores, GPU utilization, memory consumption, and VRAM usage, you can ensure that your system resources are being utilized effectively during parallel processing and GPU-accelerated grid searches. Always ensure your machine has adequate cooling and avoid overloading your system with too many processes, as it could lead to overheating or instability, especially during long-running tasks like hyperparameter tuning.

\subsection{Example Dataset and Code}
To demonstrate the importance of parallel processing, we will use the UCI \texttt{Wine Quality} dataset, which is available through the \texttt{sklearn} library. We will first perform grid search without parallelism and then with parallelism (CPU and GPU-enabled XGBoost). A larger search space will be defined to show the benefits of parallel processing.

You can download the dataset and execute the following code for comparison.

\subsubsection{Single Process: No Parallelism, CPU Only}
The following example uses \texttt{GridSearchCV} with a large search space but does not utilize parallel processing or GPU acceleration.

\begin{lstlisting}[style=python]
import xgboost as xgb
from sklearn.model_selection import GridSearchCV
from sklearn.datasets import load_wine
from sklearn.model_selection import train_test_split
from sklearn.metrics import accuracy_score

# Load the dataset
data = load_wine()
X_train, X_test, y_train, y_test = train_test_split(data.data, data.target, test_size=0.2, random_state=42)

# Define the model
model = xgb.XGBClassifier()

# Define a large parameter grid
param_grid = {
    'n_estimators': [100, 200, 300],
    'max_depth': [3, 5, 7],
    'learning_rate': [0.01, 0.1, 0.2],
    'subsample': [0.6, 0.8, 1.0],
    'colsample_bytree': [0.6, 0.8, 1.0]
}

# Set up the grid search without parallel processing
grid_search = GridSearchCV(model, param_grid, cv=5, n_jobs=1)  # Single process (n_jobs=1)

# Perform the search
grid_search.fit(X_train, y_train)

# Output the best parameters
print("Best Parameters:", grid_search.best_params_)

# Make predictions on the test set
y_pred = grid_search.best_estimator_.predict(X_test)
print("Test Accuracy:", accuracy_score(y_test, y_pred))
\end{lstlisting}

In this example, the grid search is performed without parallel processing (\texttt{n\_jobs=1}), meaning that only one CPU core is used. For larger search spaces, this can be time-consuming.

\subsubsection{Multi-Process: Parallel Processing with CPU}
Now, we modify the code to utilize all available CPU cores by setting \texttt{n\_jobs=-1}.

\begin{lstlisting}[style=python]
import xgboost as xgb
from sklearn.model_selection import GridSearchCV
from sklearn.datasets import load_wine
from sklearn.model_selection import train_test_split
from sklearn.metrics import accuracy_score

# Load the dataset
data = load_wine()
X_train, X_test, y_train, y_test = train_test_split(data.data, data.target, test_size=0.2, random_state=42)

# Define the model
model = xgb.XGBClassifier()

# Define a large parameter grid
param_grid = {
    'n_estimators': [100, 200, 300],
    'max_depth': [3, 5, 7],
    'learning_rate': [0.01, 0.1, 0.2],
    'subsample': [0.6, 0.8, 1.0],
    'colsample_bytree': [0.6, 0.8, 1.0]
}

# Set up the grid search with parallel processing (using all available cores)
grid_search = GridSearchCV(model, param_grid, cv=5, n_jobs=-1)  # Parallel processing with all cores

# Perform the search
grid_search.fit(X_train, y_train)

# Output the best parameters
print("Best Parameters:", grid_search.best_params_)

# Make predictions on the test set
y_pred = grid_search.best_estimator_.predict(X_test)
print("Test Accuracy:", accuracy_score(y_test, y_pred))
\end{lstlisting}

In this version, \texttt{n\_jobs=-1} ensures that all available CPU cores are used, greatly reducing the time required for grid search.

\subsubsection{Multi-Process with GPU: XGBoost with GPU Acceleration}
If you have a compatible GPU, you can further speed up the training process by enabling GPU acceleration in \texttt{XGBoost}. In this case, we set the \texttt{tree\_method} to \texttt{gpu\_hist} and continue using all CPU cores for parallel processing.

\begin{lstlisting}[style=python]
import xgboost as xgb
from sklearn.model_selection import GridSearchCV
from sklearn.datasets import load_wine
from sklearn.model_selection import train_test_split
from sklearn.metrics import accuracy_score

# Load the dataset
data = load_wine()
X_train, X_test, y_train, y_test = train_test_split(data.data, data.target, test_size=0.2, random_state=42)

# Define the model
model = xgb.XGBClassifier(tree_method='gpu_hist')  # Use GPU for training

# Define a large parameter grid
param_grid = {
    'n_estimators': [100, 200, 300],
    'max_depth': [3, 5, 7],
    'learning_rate': [0.01, 0.1, 0.2],
    'subsample': [0.6, 0.8, 1.0],
    'colsample_bytree': [0.6, 0.8, 1.0]
}

# Set up the grid search with parallel processing (using all available cores and GPU)
grid_search = GridSearchCV(model, param_grid, cv=5, n_jobs=-1)  # Parallel processing with GPU support

# Perform the search
grid_search.fit(X_train, y_train)

# Output the best parameters
print("Best Parameters:", grid_search.best_params_)

# Make predictions on the test set
y_pred = grid_search.best_estimator_.predict(X_test)
print("Test Accuracy:", accuracy_score(y_test, y_pred))
\end{lstlisting}

This code uses GPU acceleration by setting the \texttt{tree\_method} parameter to \texttt{gpu\_hist}, allowing \texttt{XGBoost} to utilize the GPU for faster training. Additionally, \texttt{n\_jobs=-1} ensures all CPU cores are used for the grid search itself.

\subsection{Considerations for Parallel and GPU Processing}
While using parallelism and GPU acceleration can significantly speed up hyperparameter tuning, there are several considerations to keep in mind:
\begin{itemize}
    \item \textbf{CPU Utilization:} Setting \texttt{n\_jobs} too high (more than the available cores) can lead to inefficiencies, as context switching between processes may slow down the system. Always match \texttt{n\_jobs} with the number of physical CPU cores, not threads.
    \item \textbf{Memory Constraints:} Each parallel process may require additional memory, and large datasets or models with high memory demands can cause your system to run out of RAM.
    \item \textbf{GPU Memory (VRAM):} For GPU-based models like \texttt{XGBoost}, ensure that your GPU has enough VRAM to handle the data and model size. Overloading the GPU memory can lead to performance degradation or even crashes.
    \item \textbf{Monitoring Resources:} It’s essential to monitor CPU and GPU usage, as well as memory, to ensure that your system resources are being used efficiently without exceeding their capacity.
\end{itemize}

By effectively utilizing parallel processing and GPU acceleration, you can drastically reduce the time required for hyperparameter tuning, making it feasible to explore larger search spaces or more complex models.

\section{Importance of Hyperparameter Tuning}
Hyperparameters are the parameters that define the structure of a model and its learning process. Unlike the internal parameters that are learned during training, hyperparameters are set prior to training and significantly influence a model’s performance. Proper hyperparameter tuning is critical because:

\begin{itemize}
    \item It can improve model accuracy and reduce overfitting.
    \item Proper tuning can lead to faster convergence and reduced training times.
    \item In some models, like Support Vector Machines (SVM) or neural networks, optimal performance is highly dependent on carefully chosen hyperparameters.
\end{itemize}

Without tuning, models may underperform and fail to generalize well to unseen data.

\subsection{Search Spaces for Linear Regression}
Linear regression, being one of the simplest machine learning algorithms, has relatively few hyperparameters, but tuning them can still improve performance, especially when dealing with larger datasets.

\subsubsection{Hyperparameters to consider:}
\begin{itemize}
    \item \texttt{fit\_intercept}: Whether to calculate the intercept for the model.
    \item \texttt{normalize}: Whether to normalize the features before applying the regression.
\end{itemize}

\textbf{Search Space Example:}

\begin{lstlisting}[style=python]
param_grid = {
    'fit_intercept': [True, False],
    'normalize': [True, False],
    'copy_X': [True, False]
}
\end{lstlisting}

In a normal-size space, we can include a few additional variations:

\begin{lstlisting}[style=python]
from sklearn.linear_model import LinearRegression
from sklearn.model_selection import GridSearchCV

param_grid = {
    'fit_intercept': [True, False],
    'normalize': [True, False],
    'copy_X': [True, False]
}

# Initialize the model
model = LinearRegression()

# Grid search
grid_search = GridSearchCV(model, param_grid, cv=5)
grid_search.fit(X_train, y_train)
\end{lstlisting}

\subsection{Search Spaces for SVM}
Support Vector Machines (SVMs) have a rich hyperparameter space, including parameters like the regularization parameter \(C\) and the kernel type, which directly impact the performance.

\subsubsection{Hyperparameters to consider:}
\begin{itemize}
    \item \texttt{C}: Regularization parameter. A higher value means stricter constraints on the margin.
    \item \texttt{kernel}: The kernel function to transform the data into a higher dimension.
    \item \texttt{gamma}: Defines how far the influence of a single training example reaches.
\end{itemize}

\textbf{Search Space Example:}

\begin{lstlisting}[style=python]
param_grid = {
    'C': [0.1, 1, 10, 100],
    'kernel': ['linear', 'rbf'],
    'gamma': ['scale', 'auto']
}
\end{lstlisting}

Normal-size grid search for SVM:

\begin{lstlisting}[style=python]
from sklearn.svm import SVC
from sklearn.model_selection import GridSearchCV

param_grid = {
    'C': [0.1, 1, 10, 100],
    'kernel': ['linear', 'rbf'],
    'gamma': ['scale', 'auto']
}

# Initialize the model
model = SVC()

# Grid search
grid_search = GridSearchCV(model, param_grid, cv=5)
grid_search.fit(X_train, y_train)
\end{lstlisting}

\subsection{Search Spaces for Random Forest}
Random Forest has several key hyperparameters, such as the number of trees, maximum depth, and the number of features to consider for splits. Optimizing these hyperparameters can lead to significant performance gains.

\subsubsection{Hyperparameters to consider:}
\begin{itemize}
    \item \texttt{n\_estimators}: Number of trees in the forest.
    \item \texttt{max\_depth}: The maximum depth of each tree.
    \item \texttt{max\_features}: The number of features to consider for the best split.
\end{itemize}

\textbf{Search Space Example:}

\begin{lstlisting}[style=python]
param_grid = {
    'n_estimators': [10, 50, 100, 200],
    'max_depth': [None, 10, 20, 30],
    'max_features': ['auto', 'sqrt', 'log2']
}
\end{lstlisting}

Normal-size grid search for Random Forest:

\begin{lstlisting}[style=python]
from sklearn.ensemble import RandomForestClassifier
from sklearn.model_selection import GridSearchCV

param_grid = {
    'n_estimators': [10, 50, 100, 200],
    'max_depth': [None, 10, 20, 30],
    'max_features': ['auto', 'sqrt', 'log2']
}

# Initialize the model
model = RandomForestClassifier()

# Grid search
grid_search = GridSearchCV(model, param_grid, cv=5)
grid_search.fit(X_train, y_train)
\end{lstlisting}

\subsection{Search Spaces for XGBoost}
XGBoost (Extreme Gradient Boosting) has a more complex hyperparameter space than many other algorithms, with parameters like learning rate, the number of boosting rounds, and the maximum depth of trees.

\subsubsection{Hyperparameters to consider:}
\begin{itemize}
    \item \texttt{learning\_rate}: Step size shrinkage used in updates to prevent overfitting.
    \item \texttt{n\_estimators}: Number of boosting rounds.
    \item \texttt{max\_depth}: Maximum depth of a tree.
\end{itemize}

\textbf{Search Space Example:}

\begin{lstlisting}[style=python]
param_grid = {
    'learning_rate': [0.01, 0.1, 0.2],
    'n_estimators': [100, 200, 300],
    'max_depth': [3, 5, 7]
}
\end{lstlisting}

Normal-size grid search for XGBoost:

\begin{lstlisting}[style=python]
from xgboost import XGBClassifier
from sklearn.model_selection import GridSearchCV

param_grid = {
    'learning_rate': [0.01, 0.1, 0.2],
    'n_estimators': [100, 200, 300],
    'max_depth': [3, 5, 7]
}

# Initialize the model
model = XGBClassifier()

# Grid search
grid_search = GridSearchCV(model, param_grid, cv=5)
grid_search.fit(X_train, y_train)
\end{lstlisting}

\subsection{Search Spaces for LightGBM}
LightGBM is a gradient boosting framework that uses tree-based learning algorithms. It is highly efficient, but optimizing hyperparameters such as the learning rate and the number of leaves is crucial for getting good performance.

\subsubsection{Hyperparameters to consider:}
\begin{itemize}
    \item \texttt{num\_leaves}: Maximum number of leaves in one tree.
    \item \texttt{learning\_rate}: Step size shrinkage.
    \item \texttt{n\_estimators}: Number of boosting rounds.
\end{itemize}

\textbf{Search Space Example:}

\begin{lstlisting}[style=python]
param_grid = {
    'num_leaves': [31, 50, 70],
    'learning_rate': [0.01, 0.1],
    'n_estimators': [100, 200]
}
\end{lstlisting}

Normal-size grid search for LightGBM:

\begin{lstlisting}[style=python]
import lightgbm as lgb
from sklearn.model_selection import GridSearchCV

param_grid = {
    'num_leaves': [31, 50, 70],
    'learning_rate': [0.01, 0.1],
    'n_estimators': [100, 200]
}

# Initialize the model
model = lgb.LGBMClassifier()

# Grid search
grid_search = GridSearchCV(model, param_grid, cv=5)
grid_search.fit(X_train, y_train)
\end{lstlisting}

\subsection{Search Spaces for CatBoost}
CatBoost is another gradient boosting algorithm that performs particularly well with categorical features. Its hyperparameters, such as the depth of the trees and learning rate, require careful tuning.

\subsubsection{Hyperparameters to consider:}
\begin{itemize}
    \item \texttt{depth}: Depth of the trees.
    \item \texttt{learning\_rate}: Step size shrinkage.
    \item \texttt{iterations}: Number of boosting iterations.
\end{itemize}

\textbf{Search Space Example:}

\begin{lstlisting}[style=python]
param_grid = {
    'depth': [4, 6, 10],
    'learning_rate': [0.01, 0.1, 0.2],
    'iterations': [100, 200, 300]
}
\end{lstlisting}

Normal-size grid search for CatBoost:

\begin{lstlisting}[style=python]
from catboost import CatBoostClassifier
from sklearn.model_selection import GridSearchCV

param_grid = {
    'depth': [4, 6, 10],
    'learning_rate': [0.01, 0.1, 0.2],
    'iterations': [100, 200, 300]
}

# Initialize the model
model = CatBoostClassifier()

# Grid search
grid_search = GridSearchCV(model, param_grid, cv=5)
grid_search.fit(X_train, y_train)
\end{lstlisting}

%% file: 12_automl.tex
\chapter{Overview of Automated Machine Learning}
\section{Concept of AutoML}
Automated Machine Learning, commonly referred to as \textbf{AutoML}, is the process of automating the end-to-end process of applying machine learning (ML) to real-world problems. Traditionally, applying machine learning techniques \cite{feng2024deeplearningmachinelearning} to a problem requires expertise in data science, programming, and the ability to design machine learning models. However, AutoML aims to simplify this process, making it more accessible to non-experts and reducing the time required to build ML models~\cite{he2021automl}.

AutoML frameworks generally handle several key tasks:
\begin{itemize}
    \item \textbf{Data Preprocessing}: Automated cleaning, normalization, and transformation of data to prepare it for model training.
    \item \textbf{Feature Engineering}: Automatically identifying and creating the most relevant features from raw data.
    \item \textbf{Model Selection}: Automatically selecting the best type of machine learning model (e.g., linear models, decision trees, or neural networks).
    \item \textbf{Hyperparameter Optimization}: Fine-tuning hyperparameters to improve the model’s performance.
    \item \textbf{Model Evaluation}: Assessing model performance through metrics such as accuracy, precision, recall, etc.
\end{itemize}

\subsection{Example: Traditional vs. Automated Approach}
Let us compare the traditional approach with an automated one using PyTorch. In a traditional workflow, one would have to manually preprocess data, define the model architecture, and tune hyperparameters. This is highly time-consuming and requires substantial expertise.

\textbf{Traditional Workflow in PyTorch}:
\begin{lstlisting}[style=python]
import torch
import torch.nn as nn
import torch.optim as optim
from sklearn.model_selection import train_test_split

# Define a simple dataset
X = torch.rand((1000, 10))
y = torch.randint(0, 2, (1000,))

# Split data into training and testing sets
X_train, X_test, y_train, y_test = train_test_split(X, y, test_size=0.2)

# Define a simple neural network
class SimpleNet(nn.Module):
    def __init__(self):
        super(SimpleNet, self).__init__()
        self.fc1 = nn.Linear(10, 50)
        self.fc2 = nn.Linear(50, 2)
    
    def forward(self, x):
        x = torch.relu(self.fc1(x))
        return torch.softmax(self.fc2(x), dim=1)

# Initialize the network, optimizer, and loss function
model = SimpleNet()
criterion = nn.CrossEntropyLoss()
optimizer = optim.SGD(model.parameters(), lr=0.01)

# Training loop
for epoch in range(100):
    optimizer.zero_grad()
    outputs = model(X_train)
    loss = criterion(outputs, y_train)
    loss.backward()
    optimizer.step()

# Evaluate on test data
model.eval()
with torch.no_grad():
    test_outputs = model(X_test)
    test_loss = criterion(test_outputs, y_test)
    print(f"Test Loss: {test_loss.item()}")
\end{lstlisting}

Now, with AutoML, many of these manual steps (such as model selection, data splitting, and hyperparameter tuning) are automated.

\textbf{Automated Workflow using PyTorch and an AutoML Framework}:
\begin{lstlisting}[style=python]
import torch
from auto_ml import Predictor

# Define dataset
X = torch.rand((1000, 10)).numpy()
y = torch.randint(0, 2, (1000,)).numpy()

# Initialize AutoML predictor
predictor = Predictor(type_of_estimator='classifier')
predictor.train(X, y)

# Predictions on new data
predictions = predictor.predict(X_test.numpy())
\end{lstlisting}

In the automated example, we used an AutoML library that takes care of data splitting, model selection, and training behind the scenes. This significantly reduces the complexity of the workflow.

\section{History of AutoML}
The evolution of AutoML has been driven by the increasing demand to make machine learning accessible to a wider audience, and to streamline the workflow for data scientists and engineers.

\subsection{Milestones in the Development of AutoML}
The following are key milestones in the development of AutoML:
\begin{itemize}
    \item \textbf{Early 2010s}: The first generation of AutoML tools emerged. Tools such as \textbf{Auto-WEKA} (2013) and \textbf{Auto-sklearn} (2015)~\cite{feurer2015efficient} provided early platforms for automating the machine learning pipeline. These focused on automating the model selection and hyperparameter tuning processes.
    
    \item \textbf{2018 - Neural Architecture Search (NAS)}: AutoML evolved beyond classical ML into deep learning with the introduction of NAS. Google introduced \textbf{NASNet}, which automated the process of searching for the best neural network architecture~\cite{zoph2018learning}. This was a significant breakthrough in designing efficient neural networks.

    \item \textbf{2020 - Democratization of AutoML}: Many cloud providers began offering AutoML as a service. Tools such as \textbf{Google AutoML}, \textbf{Azure AutoML}, and \textbf{Amazon SageMaker Autopilot} simplified the process of applying AutoML on scalable cloud infrastructure~\cite{li2020cloud}.

    \item \textbf{Present}: Recent advancements in AutoML focus on efficiency, model explainability, and lowering the compute cost of model selection. Research has also been growing around topics like \textbf{Few-Shot Learning}, \textbf{Zero-Shot Learning}, and \textbf{Hyperparameter Optimization}~\cite{yao2018taking}.
\end{itemize}

\subsection{Illustration of a Simple AutoML Pipeline}
A typical AutoML pipeline looks like the following:

\begin{center}
\begin{tikzpicture}
  \node (input) [rectangle, draw] {Input Data};
  \node (preprocess) [rectangle, draw, below of=input, node distance=1.5cm] {Preprocessing};
  \node (feature_engineering) [rectangle, draw, below of=preprocess, node distance=1.5cm] {Feature Engineering};
  \node (model_selection) [rectangle, draw, below of=feature_engineering, node distance=1.5cm] {Model Selection};
  \node (hyperparameter) [rectangle, draw, below of=model_selection, node distance=1.5cm] {Hyperparameter Tuning};
  \node (evaluation) [rectangle, draw, below of=hyperparameter, node distance=1.5cm] {Model Evaluation};
  \node (output) [rectangle, draw, below of=evaluation, node distance=1.5cm] {Trained Model};

  \draw [->] (input) -- (preprocess);
  \draw [->] (preprocess) -- (feature_engineering);
  \draw [->] (feature_engineering) -- (model_selection);
  \draw [->] (model_selection) -- (hyperparameter);
  \draw [->] (hyperparameter) -- (evaluation);
  \draw [->] (evaluation) -- (output);
\end{tikzpicture}
\end{center}

This pipeline represents how data flows through various stages of an AutoML system, culminating in the selection of an optimal machine learning model.

\section{AutoML Use Cases}
AutoML is highly applicable in a variety of scenarios. Some of the most common use cases include:

\subsection{Healthcare}
In healthcare, AutoML has been applied to tasks such as diagnosing diseases from medical images, predicting patient outcomes, and identifying risk factors. For instance, by automating the process of feature selection and model optimization, AutoML can assist healthcare providers in creating more accurate predictive models.

\subsection{Finance}
AutoML is used in the finance industry to detect fraud, automate trading strategies, and assess credit risk. Since financial datasets are often complex and large, AutoML helps automate the time-consuming task of model tuning and allows financial analysts to focus on interpreting the results.

\textbf{Example: Credit Risk Prediction with AutoML in PyTorch}
\begin{lstlisting}[style=python]
import torch
from auto_ml import Predictor

# Example financial dataset
X = torch.rand((5000, 20)).numpy()  # Features like transaction amount, frequency
y = torch.randint(0, 2, (5000,)).numpy()  # Credit risk labels: 0 = low, 1 = high

# AutoML process
predictor = Predictor(type_of_estimator='classifier')
predictor.train(X, y)

# Predict credit risk for new data
new_data = torch.rand((100, 20)).numpy()
predictions = predictor.predict(new_data)
print(predictions)
\end{lstlisting}

\subsection{Retail and E-commerce}
In retail, AutoML is used to predict customer behavior, optimize pricing strategies, and recommend products. AutoML frameworks can process large datasets quickly, providing insights into customer preferences and automating decisions about stock levels and promotions.

\subsection{Manufacturing}
In the manufacturing industry, AutoML can optimize processes by analyzing sensor data, predicting equipment failures, and improving supply chain operations. By automating these processes, companies can reduce downtime and increase operational efficiency.

These are just a few examples of how AutoML can be applied across different industries. The key advantage is that AutoML allows non-experts to build complex models without requiring extensive machine learning knowledge, thereby democratizing the power of AI.

%% file: 13_tpot.tex
\chapter{TPOT}
\section{Introduction to TPOT}
TPOT (Tree-based Pipeline Optimization Tool) is an open-source library designed to automate the process of machine learning, with a particular emphasis on automating feature engineering, model selection, and hyperparameter tuning~\cite{olson2016tpot}. It utilizes genetic programming to search for the best possible machine learning pipeline by evaluating different models and combinations of preprocessing steps, features, and hyperparameters.

Machine learning can often be challenging for beginners, as it requires not only a good understanding of the data but also selecting the right models and hyperparameters. TPOT simplifies this process by automatically searching through a range of models and pipeline combinations, thus allowing users to focus more on understanding their data and less on the technical details of model selection and tuning~\cite{lam2017automation}.

TPOT is built on top of the popular machine learning library scikit-learn~\cite{pedregosa2011scikit}, and it integrates seamlessly with PyTorch for deep learning tasks. TPOT handles the process of pipeline optimization by evolving the best possible model pipeline over a series of iterations. Each iteration involves generating new pipelines, evaluating their performance, and selecting the best candidates for further optimization. This process continues until a pre-defined number of generations is reached or the optimal solution is found~\cite{olson2016evo}.

In this chapter, we will explore how to install TPOT, use it to create machine learning pipelines, and fine-tune its parameters to achieve better results. By the end of this chapter, you will have a solid understanding of how TPOT works and how to leverage it to improve your machine learning workflow.

\section{Installation and Usage of TPOT}
Before we can start using TPOT, we need to install it. The installation process is straightforward, and TPOT can be installed using pip, the Python package manager. Below is the command to install TPOT along with the necessary dependencies:

\begin{lstlisting}[style=cmd]
pip install tpot
\end{lstlisting}

Once TPOT is installed, you can start using it to automate your machine learning workflows. Let’s start with a simple example to see how TPOT can help you create an optimized machine learning pipeline.

\subsection{Basic Usage Example}
Suppose we are working with a classification problem. We will use the popular Iris dataset for this example. The goal is to predict the species of iris flowers based on the provided features.

\begin{lstlisting}[style=python]
from tpot import TPOTClassifier
from sklearn.datasets import load_iris
from sklearn.model_selection import train_test_split

# Load the Iris dataset
iris = load_iris()
X_train, X_test, y_train, y_test = train_test_split(iris.data, iris.target, test_size=0.2, random_state=42)

# Create a TPOT classifier
tpot = TPOTClassifier(verbosity=2, generations=5, population_size=20)

# Fit the model
tpot.fit(X_train, y_train)

# Evaluate the model
print(tpot.score(X_test, y_test))

# Export the final pipeline
tpot.export('tpot_iris_pipeline.py')
\end{lstlisting}

In the code above, we start by loading the Iris dataset and splitting it into training and test sets. Then, we initialize the \texttt{TPOTClassifier} with the specified number of generations (5) and population size (20). TPOT will automatically evolve the best pipeline over these generations. The \texttt{verbosity} parameter controls how much information is printed to the console during the optimization process.

After fitting the model, TPOT evaluates the performance of the best pipeline on the test set and exports the final optimized pipeline as a Python script. You can then use this exported script for future predictions or further optimization.

\section{TPOT Code Implementation}
Let’s go through a more detailed implementation of TPOT for a regression problem, which is often common in machine learning tasks. In this example, we will use a housing prices dataset to predict the price of houses based on their features.

We will also introduce more options that TPOT provides, such as cross-validation and early stopping. These options can help prevent overfitting and ensure that TPOT finds robust pipelines.

\begin{lstlisting}[style=python]
import pandas as pd
from tpot import TPOTRegressor
from sklearn.model_selection import train_test_split
from sklearn.datasets import fetch_california_housing

# Load the California housing dataset
housing = fetch_california_housing()
X_train, X_test, y_train, y_test = train_test_split(housing.data, housing.target, test_size=0.2, random_state=42)

# Initialize TPOT Regressor with cross-validation and early stopping
tpot = TPOTRegressor(generations=10, population_size=50, cv=5, verbosity=2, 
                     random_state=42, early_stop=3)

# Fit the model
tpot.fit(X_train, y_train)

# Evaluate the model
print(f"Test Score: {tpot.score(X_test, y_test)}")

# Export the final pipeline
tpot.export('tpot_housing_pipeline.py')
\end{lstlisting}

In this example:
\begin{itemize}
    \item We use the California housing dataset, which is a regression dataset.
    \item We split the data into training and testing sets using \texttt{train\_test\_split}.
    \item \texttt{TPOTRegressor} is used for regression tasks. We set the number of generations to 10 and the population size to 50.
    \item We also use 5-fold cross-validation (\texttt{cv=5}) to ensure that the model generalizes well to unseen data.
    \item The \texttt{early\_stop} parameter ensures that if there is no improvement in performance after 3 generations, the optimization process stops early, saving time and computational resources.
\end{itemize}

This code will produce an optimized regression pipeline and save it to a file, which can be reused later.

\section{TPOT Parameter Tuning}
To get the best results from TPOT, it is important to understand how to fine-tune its hyperparameters. The following are some of the key parameters that can be tuned in TPOT:

\begin{itemize}
    \item \textbf{generations}: The number of generations TPOT will run through during optimization. More generations increase the chances of finding a better model but also increase computation time.
    \item \textbf{population\_size}: The number of individuals (pipelines) in each generation. Larger populations allow for more diverse pipelines to be tested.
    \item \textbf{cv}: Cross-validation splits. Using cross-validation helps prevent overfitting, especially in small datasets.
    \item \textbf{mutation\_rate}: The mutation rate controls how often parts of the pipeline are randomly changed. Higher mutation rates introduce more diversity but can also lead to instability.
    \item \textbf{crossover\_rate}: The crossover rate determines how often pipelines are combined to form new pipelines.
    \item \textbf{early\_stop}: Stops the optimization process early if no improvement is detected after a given number of generations.
    \item \textbf{subsample}: Subsamples the data to speed up processing for large datasets. This can help reduce computation time but may slightly affect accuracy.
\end{itemize}

Tuning these parameters allows TPOT to be tailored to your specific dataset and computing resources. For example, if you have a large dataset, increasing the population size and number of generations can result in a better model, while using a smaller subsample or fewer generations can reduce computational load if you are constrained by time or resources.

Below is an example of how we can fine-tune some of these parameters for a classification task:

\begin{lstlisting}[style=python]
tpot = TPOTClassifier(generations=20, population_size=100, verbosity=2, 
                      cv=10, mutation_rate=0.9, crossover_rate=0.1, early_stop=5)
tpot.fit(X_train, y_train)
\end{lstlisting}

In this case, we increase the number of generations to 20 and population size to 100, use 10-fold cross-validation, and adjust the mutation and crossover rates. This configuration allows for more exploration of different pipelines while preventing overfitting through cross-validation.

%% file: 14_autogluon.tex
\chapter{AutoGluon}
    \section{Introduction to AutoGluon}
    AutoGluon~\cite{agtabular} is an open-source toolkit designed to simplify machine learning model development. It automates the process of building, tuning, and deploying models by handling most of the complex steps automatically. This makes it an excellent choice for beginners and professionals alike, as it reduces the need for manual tuning of hyperparameters or selecting the best model. Instead, AutoGluon focuses on performance and ease of use, allowing developers to achieve state-of-the-art results without extensive knowledge of machine learning.

    AutoGluon is particularly useful for tasks such as:
    \begin{itemize}
        \item \textbf{Tabular data modeling} – structured data with rows and columns, such as CSV files.
        \item \textbf{Image classification} – automatically classifying images into different categories.
        \item \textbf{Text data processing} – handling text data for various tasks like sentiment analysis or classification.
    \end{itemize}

    The key strength of AutoGluon lies in its ability to try multiple models (e.g., decision trees, deep learning models, etc.) and select the best one. By providing a simple interface, AutoGluon helps you to get competitive results quickly without needing to fully understand the complexities of model development.

    \section{Installation and Usage of AutoGluon}
    To get started with AutoGluon, you need to install it on your local machine or development environment. AutoGluon is compatible with Python 3.7 or later and can be installed via \texttt{pip}, Python's package installer.

    \subsection{Installation}
    First, ensure that you have Python installed on your system. Then, use the following command to install AutoGluon:

    \begin{lstlisting}[style=cmd]
    pip install autogluon
    \end{lstlisting}

    Depending on your system, the installation may take a few minutes as it will install AutoGluon and all its dependencies, including PyTorch.

    \subsection{Basic Usage}
    After installing AutoGluon, you can start building machine learning models. Let's start by building a model for a tabular dataset, which is one of the most common use cases.

    For this example, we will use the popular \texttt{Kaggle Titanic} dataset, which predicts survival on the Titanic ship. The dataset contains columns like age, gender, passenger class, etc., and the goal is to predict whether a passenger survived or not.

    First, load the dataset and start building the model:

    \begin{lstlisting}[style=python]
    from autogluon.tabular import TabularPredictor
    import pandas as pd

    # Load the dataset
    train_data = pd.read_csv('train.csv')
    test_data = pd.read_csv('test.csv')

    # Define the target (what we want to predict)
    label = 'Survived'

    # Create a predictor
    predictor = TabularPredictor(label=label).fit(train_data)
    
    # Make predictions on test data
    predictions = predictor.predict(test_data)
    print(predictions)
    \end{lstlisting}

    In this code, AutoGluon handles everything for you. The \texttt{TabularPredictor} automatically tries various models, trains them, and selects the best-performing model based on the training data. In the end, you use the \texttt{predict} function to generate predictions on new data.

    \section{AutoGluon Code Implementation}
    Now, let's go step by step through a full implementation of AutoGluon for the Titanic dataset. We will load the data, preprocess it, and train the model using AutoGluon. Below is the Python code with explanations at each step.

    \subsection{Step 1: Importing Libraries}
    First, import the necessary libraries. We will need \texttt{pandas} for handling the data, and AutoGluon's \texttt{TabularPredictor} to create the model.

    \begin{lstlisting}[style=python]
    import pandas as pd
    from autogluon.tabular import TabularPredictor
    \end{lstlisting}

    \subsection{Step 2: Loading and Exploring the Data}
    Load the Titanic dataset, which is available in CSV format. Use \texttt{pandas} to read the CSV file.

    \begin{lstlisting}[style=python]
    # Load the training data
    train_data = pd.read_csv('train.csv')

    # Display the first few rows of the dataset to understand its structure
    print(train_data.head())
    \end{lstlisting}

    The dataset contains columns like \texttt{Pclass}, \texttt{Sex}, \texttt{Age}, and \texttt{Survived}. Our goal is to predict the \texttt{Survived} column.

    \subsection{Step 3: Defining the Target Variable}
    The target variable is the column you want to predict. In this case, it is the \texttt{Survived} column.

    \begin{lstlisting}[style=python]
    # Define the label (target variable)
    label = 'Survived'
    \end{lstlisting}

    \subsection{Step 4: Creating the AutoGluon Predictor}
    Now we create the \texttt{TabularPredictor}, which is responsible for automatically training and selecting the best machine learning models.

    \begin{lstlisting}[style=python]
    # Initialize the TabularPredictor
    predictor = TabularPredictor(label=label).fit(train_data)
    \end{lstlisting}

    The \texttt{fit()} function automatically trains multiple models on the training data and selects the best one. This process may take some time depending on the size of your dataset.

    \subsection{Step 5: Making Predictions}
    Once the model is trained, you can use it to make predictions on new data.

    \begin{lstlisting}[style=python]
    # Load the test data
    test_data = pd.read_csv('test.csv')

    # Make predictions
    predictions = predictor.predict(test_data)

    # Output the predictions
    print(predictions)
    \end{lstlisting}

    This will output predictions for each passenger in the test dataset, indicating whether they survived or not.

    \section{AutoGluon Parameter Tuning}
    AutoGluon is very flexible, and you can control various aspects of the model training process by tuning its parameters. Let's explore some common tuning options.

    \subsection{AutoGluon Hyperparameter Tuning}
    You can specify the type of models you want AutoGluon to try by providing a list of model hyperparameters. For example, to try only certain algorithms like LightGBM or Neural Networks, you can use the \texttt{hyperparameters} argument.

    \begin{lstlisting}[style=python]
    # Specify hyperparameters for LightGBM and Neural Networks
    hyperparameters = {
        'GBM': {},        # LightGBM
        'NN_TORCH': {},   # PyTorch Neural Networks
    }

    # Train with specific hyperparameters
    predictor = TabularPredictor(label=label).fit(train_data, hyperparameters=hyperparameters)
    \end{lstlisting}

    \subsection{Setting Time Limits}
    You can also limit the amount of time AutoGluon spends on model training using the \texttt{time\_limit} parameter. For example, if you want the process to finish within 10 minutes, you can set it as follows:

    \begin{lstlisting}[style=python]
    predictor = TabularPredictor(label=label).fit(train_data, time_limit=600)  # 600 seconds = 10 minutes
    \end{lstlisting}

    \subsection{Controlling Evaluation Metrics}
    AutoGluon evaluates models using default metrics, but you can specify custom metrics like accuracy, F1 score, etc.

    \begin{lstlisting}[style=python]
    # Train model with custom evaluation metric (e.g., F1 score)
    predictor = TabularPredictor(label=label, eval_metric='f1').fit(train_data)
    \end{lstlisting}

    \subsection{Saving and Loading Models}
    After training a model, you can save it to disk and reload it later for predictions. This is particularly useful in production environments.

    \begin{lstlisting}[style=python]
    # Save the model
    predictor.save('my_autogluon_model')

    # Load the model
    loaded_predictor = TabularPredictor.load('my_autogluon_model')
    \end{lstlisting}

    This concludes the detailed walkthrough of using AutoGluon for automating machine learning model building. The next section will delve deeper into advanced features and further optimizations.

%% file: 15_other.tex
\chapter{Other AutoML Tools}

\section{H2O AutoML}
H2O AutoML~\cite{H2OAutoML20} is an open-source machine learning platform that automates the entire model training process. It’s known for providing an easy-to-use interface for training and tuning models without needing in-depth expertise in machine learning. With H2O AutoML, you can automatically train a large variety of models, including deep learning, tree-based models, and ensembles. The tool is built to handle both regression and classification problems.

Key features of H2O AutoML:
\begin{itemize}
    \item \textbf{Wide range of algorithms:} H2O AutoML supports a variety of algorithms, including Random Forest, XGBoost, Gradient Boosting Machines (GBM), Deep Learning, and Generalized Linear Models (GLM).
    \item \textbf{Automatic ensemble creation:} H2O AutoML automatically creates and tunes ensembles of models, combining the predictions of multiple algorithms to improve accuracy.
    \item \textbf{Cross-validation:} The tool handles cross-validation for model evaluation, ensuring that the performance of each model is accurately measured.
    \item \textbf{Leaderboards:} H2O AutoML generates a leaderboard, displaying the performance of all models that were trained.
    \item \textbf{Scalability:} H2O AutoML can be distributed across a cluster, making it suitable for large datasets.
\end{itemize}

\textbf{Example Usage of H2O AutoML in Python}:
To use H2O AutoML in Python, you can follow these steps:

\begin{lstlisting}[style=python]
import h2o
from h2o.automl import H2OAutoML

# Initialize H2O cluster
h2o.init()

# Load data into H2O environment
data = h2o.import_file("your_dataset.csv")

# Split into training and test sets
train, test = data.split_frame(ratios=[0.8])

# Define target and features
x = train.columns
y = "target_column"
x.remove(y)

# Train AutoML model
aml = H2OAutoML(max_models=20, seed=1)
aml.train(x=x, y=y, training_frame=train)

# View the leaderboard
lb = aml.leaderboard
lb.head()
\end{lstlisting}

In this example, H2O AutoML automatically trains and tunes multiple models on your dataset, providing you with a leaderboard of the best models.

\section{MLBox}
MLBox is a Python-based AutoML library designed to automate the end-to-end process of machine learning, from preprocessing to model training and optimization~\cite{mlbox}. It emphasizes simplicity and automation, making it a great choice for beginners. It also provides powerful preprocessing capabilities, especially for handling missing data and unbalanced datasets~\cite{ramirez2018mlbox}.

Key features of MLBox:
\begin{itemize}
    \item \textbf{Preprocessing:} Automatically handles missing values, outliers, and categorical features.
    \item \textbf{Data cleaning:} MLBox includes built-in functionality to clean datasets and remove unnecessary features.
    \item \textbf{Model selection:} MLBox tests various machine learning models and automatically selects the best one for your problem.
    \item \textbf{Hyperparameter optimization:} It provides automatic hyperparameter tuning using Bayesian optimization.
    \item \textbf{Handles unbalanced datasets:} MLBox provides features to handle class imbalance effectively.
\end{itemize}

\textbf{Example Usage of MLBox in Python}:
Here’s how you can use MLBox to automate a machine learning workflow:

\begin{lstlisting}[style=python]
from mlbox.preprocessing import Reader, Drifter, Scanner
from mlbox.optimisation import Optimiser
from mlbox.prediction import Predictor

# Step 1: Read and preprocess the data
reader = Reader(sep=",")
train_data = reader.train_test_split(["train.csv", "test.csv"], target_name="target")

# Step 2: Identify data drift
drifter = Drifter()
drifter.fit_transform(train_data)

# Step 3: Scan for errors in the data
scanner = Scanner()
cleaned_data = scanner.fit_transform(train_data)

# Step 4: Train and optimize the model
opt = Optimiser()
best_params = opt.optimise(cleaned_data)

# Step 5: Make predictions
predictor = Predictor()
predictions = predictor.fit_predict(cleaned_data, best_params)
\end{lstlisting}

MLBox handles everything from data cleaning to model optimization, allowing beginners to easily get started with machine learning.

\section{Auto-sklearn}
Auto-sklearn is an extension of the popular scikit-learn library, providing automated machine learning with minimal coding. It leverages the simplicity of scikit-learn’s interface while automating key steps such as model selection, hyperparameter tuning, and preprocessing. Auto-sklearn also supports meta-learning, using prior knowledge to inform the model-building process.

Key features of Auto-sklearn:
\begin{itemize}
    \item \textbf{Built on scikit-learn:} Uses the familiar scikit-learn API~\cite{pedregosa2011scikit}, making it easy to integrate into existing Python workflows.
    \item \textbf{Meta-learning:} Auto-sklearn learns from past performance on similar datasets to make better decisions for new tasks.
    \item \textbf{Ensemble models:} Automatically creates ensemble models to improve prediction accuracy.
    \item \textbf{Time and resource management:} You can set time and resource limits to control the duration and computational cost of training.
    \item \textbf{Preprocessing pipelines:} Auto-sklearn automatically generates preprocessing pipelines, which can include scaling, encoding, and feature selection.
\end{itemize}

\textbf{Example Usage of Auto-sklearn in Python}:
Here’s an example of using Auto-sklearn to build a model:

\begin{lstlisting}[style=python]
import autosklearn.classification
from sklearn.model_selection import train_test_split
from sklearn.datasets import load_digits
from sklearn.metrics import accuracy_score

# Load dataset
X, y = load_digits(return_X_y=True)

# Split data into training and test sets
X_train, X_test, y_train, y_test = train_test_split(X, y, test_size=0.2, random_state=42)

# Initialize Auto-sklearn classifier
automl = autosklearn.classification.AutoSklearnClassifier(time_left_for_this_task=300)

# Train the model
automl.fit(X_train, y_train)

# Make predictions
y_pred = automl.predict(X_test)

# Evaluate model performance
accuracy = accuracy_score(y_test, y_pred)
print(f"Accuracy: {accuracy:.2f}")
\end{lstlisting}

In this example, Auto-sklearn automatically selects the best model and preprocessing pipeline for the classification task.

\section{FLAML}
FLAML (Fast and Lightweight AutoML)~\cite{wang2021flamlfastlightweightautoml} is a lightweight AutoML library developed by Microsoft, focusing on fast and efficient hyperparameter optimization. FLAML is designed to be computationally efficient, making it suitable for users who want to train models quickly without consuming significant computational resources. It is a great tool for handling both classification and regression tasks with a focus on performance and speed.

Key features of FLAML:
\begin{itemize}
    \item \textbf{Fast and efficient:} FLAML is optimized for speed, making it much faster than many other AutoML libraries.
    \item \textbf{Low computational overhead:} FLAML is designed to be lightweight, requiring fewer resources to train models.
    \item \textbf{Flexible:} FLAML supports a variety of models and tasks, including classification, regression, and time series forecasting.
    \item \textbf{No need for expensive hardware:} FLAML is designed to work on standard hardware setups, making it accessible to a wider audience.
\end{itemize}

\textbf{Example Usage of FLAML in Python}:
Here’s how you can use FLAML to train a model:

\begin{lstlisting}[style=python]
from flaml import AutoML
from sklearn.datasets import load_breast_cancer
from sklearn.model_selection import train_test_split
from sklearn.metrics import accuracy_score

# Load dataset
X, y = load_breast_cancer(return_X_y=True)

# Split data into training and test sets
X_train, X_test, y_train, y_test = train_test_split(X, y, test_size=0.2, random_state=42)

# Initialize AutoML
automl = AutoML()

# Train the model
automl.fit(X_train, y_train, task="classification", time_budget=60)

# Make predictions
y_pred = automl.predict(X_test)

# Evaluate the model
accuracy = accuracy_score(y_test, y_pred)
print(f"Accuracy: {accuracy:.2f}")
\end{lstlisting}

In this example, FLAML trains a classification model within a specified time budget, providing a fast and efficient solution for automated machine learning.

%% file: 16_cloudbase.tex
\part{Cloud-Based AutoML Tools}

\chapter{DataRobot}
    \section{Introduction to DataRobot}
        DataRobot~\cite{datarobot} is a powerful cloud-based AutoML (Automated Machine Learning) platform that allows users to build, deploy, and manage machine learning models with minimal manual intervention. Designed for both data scientists and business analysts, DataRobot offers a variety of automated features such as model selection, data preprocessing, and hyperparameter tuning. It is particularly useful for those who may not have deep expertise in machine learning but need to leverage the power of AI.

        DataRobot integrates well with other platforms and can ingest data from multiple sources, such as CSV files, databases, or cloud storage. Once the data is uploaded, the platform performs tasks like data cleaning, feature engineering, model training, and evaluation automatically. 

        Its simplicity is one of its strengths: after uploading data, DataRobot automatically evaluates hundreds of models and provides the user with a leaderboard of the best-performing ones. This removes the guesswork in model selection and optimization.

    \section{Key Features of DataRobot}
        DataRobot offers several key features that make it an attractive tool for automating the machine learning process. Below are some of the major functionalities:
        
        \subsection{Automatic Model Selection}
        When working with machine learning, one of the biggest challenges is selecting the appropriate model for a given dataset. DataRobot addresses this by automatically evaluating various machine learning algorithms, ranging from simple linear models to advanced neural networks. The platform uses cross-validation to ensure that the models are generalizable and presents the results in a leaderboard format, ranking models based on performance metrics such as accuracy, F1 score, or AUC (Area Under the Curve).

        \subsection{Automatic Feature Engineering}
        Feature engineering is the process of transforming raw data into features that better represent the underlying patterns. DataRobot automates much of this process by applying techniques such as encoding categorical variables, scaling numerical features, and creating polynomial features when necessary. This saves a lot of manual effort and can improve model performance significantly.

        \subsection{Hyperparameter Optimization}
        Each machine learning algorithm has a set of hyperparameters that control the learning process. Tuning these hyperparameters manually can be time-consuming and requires expertise. DataRobot automates this by running multiple experiments with different sets of hyperparameters and selecting the optimal configuration based on performance.

        \subsection{Model Interpretability}
        Understanding how a model makes predictions is crucial for trust and transparency. DataRobot provides tools like feature importance charts and prediction explanations to help users understand which variables are driving model predictions. This is especially useful in business settings where decisions based on machine learning need to be justified.

    \section{How to Use DataRobot}
        Getting started with DataRobot is straightforward. Below is a step-by-step guide to help you use DataRobot for AutoML.

        \subsection{Step 1: Upload Data}
        The first step in using DataRobot is to upload your dataset. This can be done by dragging and dropping a CSV file into the platform or by connecting to an external data source such as AWS S3 or a SQL database.

        \subsection{Step 2: Set Target Variable}
        Once the data is uploaded, DataRobot will automatically identify the features and ask you to select the target variable, which is the column you want to predict (for example, \texttt{SalePrice} in a house pricing dataset).

        \subsection{Step 3: Automatic Model Training}
        After specifying the target variable, DataRobot will automatically start evaluating different machine learning models. It runs them in parallel, optimizing hyperparameters and generating a leaderboard of the best models.

        \subsection{Step 4: Evaluate Models}
        Once the models are trained, you can evaluate them based on performance metrics. DataRobot will present a clear leaderboard where models are ranked, and you can choose the best model for your use case.

        \subsection{Step 5: Deployment}
        Once you are satisfied with the performance of a model, DataRobot allows you to deploy it with a single click. The platform generates an API endpoint, making it easy to integrate the model into production environments.

\chapter{DataDog}
    \section{Introduction to DataDog}
        DataDog~\cite{datadog2021platform} is a cloud-based monitoring and analytics platform that provides real-time insights into IT infrastructure, applications, and cloud services. It is widely used in industries where monitoring the health of systems, applications, and services is critical to ensuring smooth operations.

        DataDog allows users to monitor a wide range of metrics, set up alerts, and visualize data through customizable dashboards. This makes it a valuable tool for DevOps teams, software developers, and IT operations.

        The platform integrates with over 400 technologies, including cloud providers like AWS, GCP, and Azure, as well as popular frameworks like Docker, Kubernetes, and databases such as PostgreSQL. This extensive integration ensures that users can monitor all their systems from a single interface.

    \section{Key Features of DataDog}
        DataDog has several key features that make it a popular choice for real-time monitoring and analytics:

        \subsection{Infrastructure Monitoring}
        DataDog provides comprehensive infrastructure monitoring that tracks metrics like CPU usage, memory consumption, disk I/O, and network traffic. This allows system administrators to identify potential bottlenecks or failures before they impact end users.

        \subsection{Application Performance Monitoring (APM)}
        DataDog’s APM capabilities allow you to trace requests in real-time across distributed systems. This is especially useful in microservices architectures where tracing a request as it moves through different services is essential for identifying performance bottlenecks.

        \subsection{Log Management}
        In addition to metrics and traces, DataDog offers log management features that enable users to collect, analyze, and visualize logs from all of their applications and systems in a single platform. This makes it easier to debug issues and identify anomalies.

        \subsection{Alerting and Notifications}
        DataDog provides flexible alerting options. You can set thresholds for specific metrics and receive notifications via email, Slack, or other channels when those thresholds are breached. This ensures that you are informed of any critical issues as soon as they arise.

        \subsection{Dashboards and Visualization}
        One of the most powerful features of DataDog is its customizable dashboards. Users can create real-time dashboards that aggregate and display metrics, traces, and logs from multiple sources. These dashboards are useful for both technical teams monitoring system health and business teams tracking key performance indicators (KPIs).

    \section{How to Use DataDog}
        Getting started with DataDog is simple. Follow the steps below to set up your first monitoring dashboard and start tracking key metrics.

        \subsection{Step 1: Install the DataDog Agent}
        To begin monitoring, you need to install the DataDog agent on your server or container. The agent is responsible for collecting metrics, logs, and traces from your system and sending them to the DataDog platform.

        \begin{lstlisting}[style=cmd]
        DD_AGENT_MAJOR_VERSION=7 DD_API_KEY=<YOUR_API_KEY> bash -c "$(curl -L https://s3.amazonaws.com/dd-agent/scripts/install_script.sh)"
        \end{lstlisting}

        Replace \texttt{<YOUR\_API\_KEY>} with the API key provided by DataDog after you create an account.

        \subsection{Step 2: Integrate with Cloud Providers}
        DataDog supports integration with cloud providers like AWS, GCP, and Azure. To monitor cloud infrastructure, navigate to the \texttt{Integrations} tab in DataDog, select your cloud provider, and follow the instructions to connect your account.

        \subsection{Step 3: Set Up Dashboards}
        Once the data is flowing into DataDog, you can create a new dashboard. Go to the \texttt{Dashboards} tab, click \texttt{New Dashboard}, and add widgets for the metrics you want to monitor. For example, you might want to add a widget that tracks CPU usage, memory consumption, and network traffic.

        \subsection{Step 4: Set Up Alerts}
        To ensure you are notified when something goes wrong, set up alerts. Navigate to the \texttt{Monitors} tab and click \texttt{New Monitor}. You can specify the conditions under which you want to be alerted, such as CPU usage exceeding 90\%.

        \subsection{Step 5: View Logs and Traces}
        In addition to monitoring metrics, you can also view logs and traces. Navigate to the \texttt{Logs} tab to see real-time logs from your applications, or go to the \texttt{APM} tab to trace requests across your services.

        DataDog’s intuitive interface makes it easy to start monitoring infrastructure and applications in real time, helping you maintain system reliability and performance.

%% file: 17_dl.tex
\part{Deep Learning and Neural Networks}

\chapter{Introduction to Neural Networks}

\section{Basic Concepts of Artificial Neural Networks}

Artificial Neural Networks (ANNs) are a class of algorithms that attempt to mimic the workings of the human brain~\cite{lecun2015deep}. The fundamental building blocks of neural networks are neurons (also called nodes or units), which are structured in layers. A neural network typically consists of an input layer, one or more hidden layers, and an output layer~\cite{goodfellow2016deep}. 

Each neuron receives inputs, applies a linear combination to those inputs, and passes the result through a non-linear activation function. This enables the network to capture complex relationships in the data. For a simple feedforward network, this can be visualized as:

\begin{center}
    \begin{tikzpicture}
    \node[circle, draw] (A) at (0, 2) {Input 1};
    \node[circle, draw] (B) at (0, 0) {Input 2};
    \node[circle, draw] (C) at (3, 3) {Neuron 1};
    \node[circle, draw] (D) at (3, 1) {Neuron 2};
    \node[circle, draw] (E) at (3, -1) {Neuron 3};
    \node[circle, draw] (F) at (6, 1) {Output};

    \draw[->] (A) -- (C);
    \draw[->] (A) -- (D);
    \draw[->] (A) -- (E);
    \draw[->] (B) -- (C);
    \draw[->] (B) -- (D);
    \draw[->] (B) -- (E);
    \draw[->] (C) -- (F);
    \draw[->] (D) -- (F);
    \draw[->] (E) -- (F);
    \end{tikzpicture}
\end{center}

In this diagram, each circle represents a neuron. The lines connecting them indicate how information flows through the network. When a neural network learns, it adjusts the weights on the connections to minimize the error between the predicted and actual outputs.

A simple example in PyTorch can show how a neural network with one hidden layer works:

\begin{lstlisting}[style=python]
import torch
import torch.nn as nn
import torch.optim as optim

# Define a simple feedforward neural network
class SimpleNN(nn.Module):
    def __init__(self):
        super(SimpleNN, self).__init__()
        # One hidden layer with 3 neurons, input size 2, output size 1
        self.fc1 = nn.Linear(2, 3)
        self.fc2 = nn.Linear(3, 1)

    def forward(self, x):
        x = torch.sigmoid(self.fc1(x))  # Apply sigmoid to hidden layer
        x = torch.sigmoid(self.fc2(x))  # Apply sigmoid to output
        return x

# Create the network
net = SimpleNN()

# Example input
input_data = torch.tensor([[0.5, 0.3], [0.2, 0.8]])

# Forward pass through the network
output = net(input_data)
print(output)
\end{lstlisting}

In this code, we create a simple neural network using PyTorch. It has an input layer with 2 inputs, one hidden layer with 3 neurons, and an output layer with 1 output neuron. The `torch.sigmoid` function is used as the activation function to introduce non-linearity into the network. Non-linear activation functions are important because they allow the network to learn complex patterns.

\section{Backpropagation and Gradient Descent}

Once a neural network is constructed, it needs to be trained to make accurate predictions. This is where \textit{backpropagation} and \textit{gradient descent} come into play. The goal of training is to adjust the weights in the network such that the output of the network is as close as possible to the actual target values. The difference between the predicted and target values is measured by a loss function, such as Mean Squared Error (MSE).

\textbf{Backpropagation} is an algorithm that computes the gradient of the loss function with respect to each weight in the network by propagating the error backwards through the network. Once these gradients are calculated, \textbf{gradient descent} is used to update the weights.

Gradient descent works by taking small steps in the direction that minimizes the loss function. The size of these steps is controlled by a parameter called the learning rate. Here's a simplified step-by-step process:

\begin{enumerate}
    \item Compute the loss (e.g., using Mean Squared Error).
    \item Use backpropagation to calculate the gradients of the loss with respect to each weight.
    \item Update each weight by moving it in the opposite direction of its gradient (i.e., decrease the weights that increase the loss).
\end{enumerate}

In PyTorch, the backpropagation and gradient descent steps are handled automatically when we define the loss function and optimizer:

\begin{lstlisting}[style=python]
# Loss function and optimizer
criterion = nn.MSELoss()
optimizer = optim.SGD(net.parameters(), lr=0.01)

# Target values
target = torch.tensor([[0.6], [0.4]])

# Training loop
for epoch in range(100):  # Train for 100 epochs
    optimizer.zero_grad()  # Zero the gradients from the previous step

    # Forward pass
    output = net(input_data)
    
    # Compute loss
    loss = criterion(output, target)
    
    # Backward pass (compute gradients)
    loss.backward()
    
    # Update weights
    optimizer.step()

    print(f'Epoch {epoch+1}, Loss: {loss.item()}')
\end{lstlisting}

In this example, we define an MSE loss function using \texttt{nn.MSELoss()} and an optimizer using Stochastic Gradient Descent (SGD) with a learning rate of 0.01. The training loop runs for 100 epochs, and in each iteration, the gradients are calculated using \texttt{loss.backward()}, and the weights are updated using \texttt{optimizer.step()}.

\section{Basic Structure of Deep Learning Models}

Deep learning models are an extension of neural networks where the architecture consists of many layers. These networks are capable of learning intricate patterns in large datasets and are the foundation of modern AI applications such as image recognition, speech processing, and natural language understanding.

A deep learning model typically contains:

\begin{enumerate}
    \item \textbf{Input Layer}: The layer where the input data is fed into the network.
    \item \textbf{Hidden Layers}: Multiple layers between the input and output that capture complex features. Each hidden layer transforms the data using linear and non-linear operations.
    \item \textbf{Output Layer}: The final layer that produces the predicted output.
\end{enumerate}

For instance, consider a deep neural network used for image classification. The input to this network might be the pixel values of an image, and the output could be a probability distribution over various classes (e.g., dog, cat, car, etc.).

Here’s an example of a deeper neural network in PyTorch:

\begin{lstlisting}[style=python]
# Deep neural network with 2 hidden layers
class DeepNN(nn.Module):
    def __init__(self):
        super(DeepNN, self).__init__()
        self.fc1 = nn.Linear(784, 128)  # Input layer (e.g., 28x28 image -> 784)
        self.fc2 = nn.Linear(128, 64)   # First hidden layer
        self.fc3 = nn.Linear(64, 10)    # Output layer (10 classes)

    def forward(self, x):
        x = torch.relu(self.fc1(x))  # Apply ReLU to first hidden layer
        x = torch.relu(self.fc2(x))  # Apply ReLU to second hidden layer
        x = self.fc3(x)              # Output layer (no activation for raw scores)
        return x

# Create the deep network
deep_net = DeepNN()

# Example input (batch of 5 images, each of size 28x28 pixels)
input_data = torch.randn(5, 784)

# Forward pass through the network
output = deep_net(input_data)
print(output)
\end{lstlisting}

In this example, the deep neural network has two hidden layers. The first layer reduces the input size from 784 (for a 28x28 image) to 128, and the second hidden layer further reduces it to 64. The output layer produces a 10-dimensional vector representing the raw scores for each class.

The \textit{ReLU} (Rectified Linear Unit) activation function is used in the hidden layers. ReLU is one of the most commonly used activation functions in deep learning as it helps to avoid the vanishing gradient problem during training. In practice, deep learning models can have many more layers, and training them often requires powerful hardware (e.g., GPUs) and large amounts of data.

%% file: 18_cnn.tex
\chapter{Convolutional Neural Networks (CNN)}
    \section{Principles of Convolutional Neural Networks}
        Convolutional Neural Networks (CNNs)~\cite{oshea2015introductionconvolutionalneuralnetworks} are a specialized kind of neural network specifically designed for working with image data, although they can also be applied to other types of structured data. Unlike traditional fully connected networks, CNNs exploit the spatial structure of the data, which is especially useful when dealing with images. In this section, we will walk through the core operations of a CNN: convolution, activation functions, pooling, and fully connected layers.

        \subsection{Convolution}
        The convolution operation is at the heart of CNNs. It works by applying a filter (or kernel) to an input image, creating a feature map that highlights different aspects of the image, such as edges or textures.

        Mathematically, convolution is expressed as follows:

        \[
        (I * K)(x, y) = \sum_{i=-k}^{k} \sum_{j=-k}^{k} I(x+i, y+j) K(i, j)
        \]

        Here, \(I(x, y)\) is the input image, \(K(i, j)\) is the convolutional kernel (or filter), and \((x, y)\) represents the coordinates of the pixel in the output feature map.

        \textbf{Example:} 
        Suppose we have a 5x5 grayscale image and a 3x3 filter (also called a kernel):

        \begin{lstlisting}[style=python]
        import torch
        import torch.nn.functional as F

        # Sample 5x5 image
        image = torch.tensor([[1, 2, 0, 1, 0],
                              [0, 1, 2, 1, 1],
                              [3, 1, 2, 0, 0],
                              [0, 0, 1, 2, 2],
                              [1, 1, 0, 1, 3]]).float().unsqueeze(0).unsqueeze(0)

        # 3x3 filter
        kernel = torch.tensor([[0, 1, 2],
                               [2, 2, 0],
                               [0, 1, 0]]).float().unsqueeze(0).unsqueeze(0)

        # Apply convolution
        output = F.conv2d(image, kernel)

        print(output)
        \end{lstlisting}
        
        The above code shows how we can perform a convolution on a small image using a specific kernel in PyTorch. The output feature map will have the filtered values, highlighting certain patterns within the image.

        \subsection{Activation Functions}
        After each convolutional operation, we apply an activation function, typically ReLU (Rectified Linear Unit). The ReLU function is defined as:

        \[
        \text{ReLU}(x) = \max(0, x)
        \]

        This operation ensures that the network can model non-linearities, which is crucial for learning complex patterns in data.

        \begin{lstlisting}[style=python]
        # Applying ReLU activation function
        relu_output = F.relu(output)
        print(relu_output)
        \end{lstlisting}

        \subsection{Pooling}
        Pooling is a down-sampling operation that reduces the spatial dimensions of the feature maps, thus reducing the computational load and helping to prevent overfitting. The most common type is max-pooling, which selects the maximum value from a region of the feature map.

        \textbf{Example:}
        \begin{lstlisting}[style=python]
        # Max Pooling 2x2
        pooled_output = F.max_pool2d(relu_output, 2)
        print(pooled_output)
        \end{lstlisting}

        This reduces the size of the feature map while retaining important features.

        \subsection{Fully Connected Layer}
        After a series of convolution and pooling layers, the feature maps are flattened into a single vector and passed to a fully connected layer (or dense layer). The fully connected layer learns to classify based on the features extracted by the convolutional layers.

        \begin{lstlisting}[style=python]
        # Flatten the output and pass through a fully connected layer
        flattened_output = pooled_output.view(-1)
        fully_connected_layer = torch.nn.Linear(flattened_output.shape[0], 10)
        final_output = fully_connected_layer(flattened_output)
        print(final_output)
        \end{lstlisting}

        In this example, the fully connected layer is initialized with 10 outputs, which could correspond to the number of classes in a classification problem (e.g., digits 0–9 in the case of MNIST dataset).

    \section{Classic CNN Architectures}
    Over the years, several CNN architectures have been proposed, each introducing novel ideas to improve the learning process and handle deeper networks. We will discuss some of the most influential architectures: VGG, Inception, Xception, ResNet, and DenseNet.

        \subsection{VGG}
        VGG~\cite{simonyan2015deepconvolutionalnetworkslargescale}, proposed by the Visual Geometry Group at Oxford, is known for its simplicity and depth. The architecture consists of deep convolutional layers, with each layer followed by ReLU activation and max-pooling. What makes VGG special is that it uses small 3x3 filters throughout the network, combined with deep layers.

        \textbf{Example:}

        \begin{lstlisting}[style=python]
        import torch.nn as nn

        class VGG(nn.Module):
            def __init__(self):
                super(VGG, self).__init__()
                self.conv1 = nn.Conv2d(3, 64, kernel_size=3, padding=1)
                self.conv2 = nn.Conv2d(64, 128, kernel_size=3, padding=1)
                self.fc = nn.Linear(128 * 56 * 56, 10)  # Assuming input image is 224x224

            def forward(self, x):
                x = F.relu(self.conv1(x))
                x = F.max_pool2d(x, 2)
                x = F.relu(self.conv2(x))
                x = F.max_pool2d(x, 2)
                x = x.view(x.size(0), -1)  # Flatten the tensor
                x = self.fc(x)
                return x
        \end{lstlisting}

        VGG's deep architecture allows it to capture hierarchical features of images.

        \subsection{Inception v1, v2, v3, v4}
        The Inception architecture introduces the idea of using multiple convolutional filters in parallel, creating a network that can capture different types of information from the same input. Inception blocks contain 1x1, 3x3, and 5x5 convolutions, followed by max-pooling, all running in parallel and their outputs concatenated.

\begin{center}
\begin{tikzpicture}
    [every node/.style={rectangle,draw,minimum width=2.5cm,align=center}, node distance=1.5cm]
    \node (start) {Input Image};
    \node (c1) [below left of=start, xshift=-5cm] {1x1 Convolution};
    \node (c2) [below left of=start, xshift=-1cm] {3x3 Convolution};
    \node (c3) [below right of=start, xshift=1cm] {5x5 Convolution};
    \node (pool) [below right of=start, xshift=5cm] {Max-Pooling};
    \node (concat) [below of=c2, xshift=1.5cm, yshift=0cm] {Concatenation};

    \draw[->] (start) -- (c1);
    \draw[->] (start) -- (c2);
    \draw[->] (start) -- (c3);
    \draw[->] (start) -- (pool);
    \draw[->] (c1) -- (concat);
    \draw[->] (c2) -- (concat);
    \draw[->] (c3) -- (concat);
    \draw[->] (pool) -- (concat);
\end{tikzpicture}
\end{center}

        Inception allows the network to look at an image from different perspectives, helping improve performance.

        \subsection{Xception}
        Xception~\cite{chollet2017xceptiondeeplearningdepthwise} builds on the Inception architecture by replacing traditional convolutions with depthwise separable convolutions. This operation is more efficient because it first performs a spatial convolution for each channel individually and then combines them.

        \subsection{ResNet}
        ResNet~\cite{he2015deepresiduallearningimage}, short for Residual Networks, introduces the concept of skip connections. These connections allow the model to learn residuals (the difference between the input and output of a layer), which solves the vanishing gradient problem, enabling very deep networks.

        \textbf{Example:}

        \begin{lstlisting}[style=python]
        class BasicBlock(nn.Module):
            def __init__(self, in_channels, out_channels):
                super(BasicBlock, self).__init__()
                self.conv1 = nn.Conv2d(in_channels, out_channels, kernel_size=3, padding=1)
                self.conv2 = nn.Conv2d(out_channels, out_channels, kernel_size=3, padding=1)
                self.skip_connection = nn.Sequential()

            def forward(self, x):
                residual = x
                x = F.relu(self.conv1(x))
                x = self.conv2(x)
                x += residual  # Adding skip connection
                return F.relu(x)
        \end{lstlisting}

        ResNet's skip connections allow gradients to flow more easily through deep networks, preventing vanishing gradients.

        \subsection{DenseNet}
        DenseNet~\cite{huang2018denselyconnectedconvolutionalnetworks} is another network that focuses on improving the gradient flow. In DenseNet, each layer receives inputs from all preceding layers, which enhances feature propagation and helps in alleviating the vanishing gradient problem.

        \textbf{Example:}

        \begin{lstlisting}[style=python]
        class DenseLayer(nn.Module):
            def __init__(self, in_channels, growth_rate):
                super(DenseLayer, self).__init__()
                self.conv = nn.Conv2d(in_channels, growth_rate, kernel_size=3, padding=1)

            def forward(self, x):
                new_features = F.relu(self.conv(x))
                return torch.cat([x, new_features], 1)
        \end{lstlisting}

        In DenseNet, the input to each layer is concatenated with its output, creating a densely connected network, which helps improve feature reuse.

%% file: 19_nas.tex
\chapter{Neural Architecture Search (NAS)}

\section{Concept of Neural Architecture Search (NAS)}
Neural Architecture Search (NAS) is an automated process used to design the structure of neural networks~\cite{elsken2019neural}. Traditional neural network design requires expert knowledge and time-consuming experimentation to determine the most effective architecture for a given task. NAS aims to automate this process, significantly reducing the effort required and often discovering architectures that outperform those designed manually~\cite{zoph2016neural}.

NAS optimizes both the structure (topology) of a neural network and its hyperparameters. The search for the best architecture can be framed as an optimization problem, where the objective is to find the network configuration that maximizes performance on a given task, such as image classification or natural language processing~\cite{liu2018darts}.

The importance of NAS lies in its ability to:
\begin{itemize}
    \item Reduce the time and expertise required to design neural networks.
    \item Discover novel architectures that outperform human-designed models.
    \item Automate the exploration of large design spaces, allowing for deeper and more complex models to be efficiently evaluated.
\end{itemize}

NAS generally consists of three key components:
\begin{itemize}
    \item \textbf{Search Space:} Defines the possible neural network architectures to explore. This space includes the types of layers, number of neurons, activation functions, and more.
    \item \textbf{Search Strategy:} The method used to explore the search space, which could be based on reinforcement learning, evolutionary algorithms, or gradient-based approaches.
    \item \textbf{Evaluation Strategy:} The method used to evaluate the performance of each architecture, typically through training and validation on a specific task.
\end{itemize}

\section{NASNet}
\subsection{Introduction to NASNet}
NASNet is one of the most well-known examples of a neural network architecture discovered through NAS. Developed by Google Brain, NASNet was designed to tackle the task of image classification by automating the search for an optimal convolutional neural network (CNN) architecture.

NASNet introduced a modular approach, where the search was performed on a smaller-scale architecture, and the best-found architecture was then scaled to larger networks. This modular approach made the search process more efficient, as it reduced the computational cost of exploring large, complex networks.

\subsection{Principles of NASNet}
NASNet operates using the following principles:
\begin{itemize}
    \item \textbf{Search Space:} NASNet uses a restricted search space focused on convolutional cells. These cells act as building blocks that can be stacked together to form larger networks. The search space includes different types of convolutional layers, pooling operations, and activation functions.
    \item \textbf{Search Strategy:} NASNet employs reinforcement learning to guide the search process. A controller neural network proposes candidate architectures, which are then trained and evaluated. The controller is updated based on the performance of the proposed architectures, gradually improving the search process.
    \item \textbf{Scalability:} Once the optimal architecture for a small model is found, it can be scaled up to larger models by stacking more cells or increasing the number of filters in each layer.
\end{itemize}

\subsection{Implementation and Applications of NASNet}
NASNet is primarily used for image classification tasks, where it has achieved state-of-the-art results on benchmarks like ImageNet. Below is an example of how to implement NASNet using PyTorch. While NASNet itself is typically implemented with TensorFlow, here we use a simplified PyTorch implementation to illustrate the concept.

\begin{lstlisting}[style=python]
import torch
import torch.nn as nn
import torch.optim as optim

class NASNetCell(nn.Module):
    def __init__(self, in_channels, out_channels):
        super(NASNetCell, self).__init__()
        self.conv1 = nn.Conv2d(in_channels, out_channels, kernel_size=3, padding=1)
        self.conv2 = nn.Conv2d(out_channels, out_channels, kernel_size=3, padding=1)
        self.relu = nn.ReLU()
        self.pool = nn.MaxPool2d(kernel_size=2, stride=2)

    def forward(self, x):
        x = self.relu(self.conv1(x))
        x = self.pool(self.relu(self.conv2(x)))
        return x

class NASNet(nn.Module):
    def __init__(self, num_classes=10):
        super(NASNet, self).__init__()
        self.cell1 = NASNetCell(3, 64)
        self.cell2 = NASNetCell(64, 128)
        self.cell3 = NASNetCell(128, 256)
        self.fc = nn.Linear(256 * 4 * 4, num_classes)

    def forward(self, x):
        x = self.cell1(x)
        x = self.cell2(x)
        x = self.cell3(x)
        x = x.view(x.size(0), -1)  # Flatten the tensor
        x = self.fc(x)
        return x

# Example usage:
model = NASNet(num_classes=10)
optimizer = optim.Adam(model.parameters(), lr=0.001)
criterion = nn.CrossEntropyLoss()

# Assume we have a DataLoader called train_loader
# for epoch in range(num_epochs):
#     for images, labels in train_loader:
#         optimizer.zero_grad()
#         outputs = model(images)
#         loss = criterion(outputs, labels)
#         loss.backward()
#         optimizer.step()
\end{lstlisting}

In this code, we define a simple NASNet-like architecture with modular "cells" that can be repeated and scaled up. Each cell contains convolutional layers, activation functions, and pooling layers.

\section{Other NAS Tools}
In addition to NASNet, there are several other prominent tools for neural architecture search. Two of the most popular are:
\begin{itemize}
    \item \textbf{DARTS (Differentiable Architecture Search):} DARTS is a gradient-based NAS method that significantly reduces the computational cost of NAS~\cite{liu2018darts}. Unlike traditional NAS methods, which require training many different architectures from scratch, DARTS allows for a continuous search space that can be optimized using gradients. This drastically reduces the number of required evaluations.
    \item \textbf{ENAS (Efficient Neural Architecture Search):} ENAS is a reinforcement learning-based NAS method that focuses on efficiency~\cite{pham2018efficient}. It introduces the concept of a shared network, where different candidate architectures share weights during the training process. This reduces the computational overhead of NAS while still providing competitive results.
\end{itemize}

DARTS and ENAS represent two important trends in NAS research: improving the efficiency of the search process while maintaining high performance~\cite{elsken2019neural}.

\textbf{Example of DARTS implementation:}
\begin{lstlisting}[style=python]
import torch
import torch.nn as nn

class DARTSCell(nn.Module):
    def __init__(self, in_channels, out_channels):
        super(DARTSCell, self).__init__()
        self.conv = nn.Conv2d(in_channels, out_channels, kernel_size=3, padding=1)
        self.relu = nn.ReLU()
    
    def forward(self, x):
        return self.relu(self.conv(x))

# This is a simplified example of a cell in a DARTS architecture
class DARTS(nn.Module):
    def __init__(self, num_classes=10):
        super(DARTS, self).__init__()
        self.cell = DARTSCell(3, 64)
        self.fc = nn.Linear(64 * 32 * 32, num_classes)

    def forward(self, x):
        x = self.cell(x)
        x = x.view(x.size(0), -1)
        x = self.fc(x)
        return x

model = DARTS(num_classes=10)
# Similar training loop as NASNet
\end{lstlisting}

This example showcases how DARTS simplifies the search process by using gradient-based optimization techniques, which can lead to faster discovery of high-performance architectures. Both NASNet and DARTS highlight the power of automated neural architecture search in the evolution of deep learning models.

%% file: 20_automl_dl.tex
\chapter{AutoML for Deep Learning Models}
  
\section{Combining Deep Learning and AutoML}

Deep learning models are known for their powerful ability to automatically extract features and make predictions from large and complex datasets. However, building and optimizing these models can be a challenging task, especially for beginners. This is where \textbf{AutoML (Automated Machine Learning)} comes in.

AutoML tools aim to automate the process of designing, training, and optimizing machine learning models, including deep learning models. These tools can help you:

\begin{itemize}
    \item Select the best neural network architecture.
    \item Automatically adjust hyperparameters (like learning rates and batch sizes).
    \item Train models on your dataset without the need for manual tuning.
\end{itemize}

In this section, we will discuss two popular Python-based AutoML libraries that support deep learning: \textbf{Auto-Keras} and \textbf{Auto-PyTorch}. Both of these tools simplify the creation of deep learning models and allow even beginners to achieve high-performing models with minimal effort.

\subsection{Benefits of Combining Deep Learning with AutoML}
When AutoML is combined with deep learning, it offers several advantages, including:

\begin{itemize}
    \item \textbf{Reduced Time and Effort:} AutoML automates much of the model-building process, which can save hours or even days of manual work.
    \item \textbf{Optimized Performance:} AutoML tools use techniques such as hyperparameter optimization to improve the accuracy and efficiency of deep learning models.
    \item \textbf{Beginner-Friendly:} AutoML makes deep learning accessible to those who may not have extensive experience with neural network design or tuning.
\end{itemize}

\section{Auto-Keras}
\textbf{Auto-Keras}~\cite{jin2019autokerasefficientneuralarchitecture} is an open-source AutoML library designed to make deep learning more accessible. It is built on top of Keras, but since this book focuses on PyTorch, we'll discuss the concepts rather than the underlying TensorFlow framework.

Auto-Keras is designed to automate the entire model-building process. This includes tasks like model selection, hyperparameter tuning, and training. Auto-Keras supports a variety of tasks such as image classification, text classification, and regression.

\subsection{How to Use Auto-Keras}

Let’s explore a simple example of using Auto-Keras for image classification. The following steps guide you through the entire process.

\subsubsection{Install Auto-Keras}

To get started with Auto-Keras, you first need to install the library. You can do this using pip:

\begin{lstlisting}[style=cmd]
pip install autokeras
\end{lstlisting}

\subsubsection{Load a Dataset}

Next, you need to load a dataset. Auto-Keras can handle various types of datasets, including images. Let's work with the CIFAR-10 dataset, which contains 60,000 images classified into 10 categories.

\begin{lstlisting}[style=python]
from keras.datasets import cifar10
from autokeras import ImageClassifier

# Load the CIFAR-10 dataset
(x_train, y_train), (x_test, y_test) = cifar10.load_data()
\end{lstlisting}

\subsubsection{Create and Train the Model}

Auto-Keras automates the model creation process. You simply need to create an \texttt{ImageClassifier} instance and call the \texttt{fit()} method.

\begin{lstlisting}[style=python]
# Initialize the Auto-Keras ImageClassifier
clf = ImageClassifier(max_trials=10)  # Tries 10 different models

# Train the model
clf.fit(x_train, y_train, epochs=10)
\end{lstlisting}

Here, \texttt{max\_trials} refers to how many different models Auto-Keras will try before selecting the best one. After training, Auto-Keras automatically selects the best-performing model based on the given data.

\subsubsection{Evaluate the Model}

Once the model is trained, you can evaluate its performance on the test set using the \texttt{evaluate()} method.

\begin{lstlisting}[style=python]
# Evaluate the best model
accuracy = clf.evaluate(x_test, y_test)
print("Test accuracy:", accuracy)
\end{lstlisting}

This completes the basic workflow of using Auto-Keras for image classification. The entire process—loading data, training a model, and evaluating its performance—is handled with minimal code.

\section{Auto-PyTorch}

Unlike Auto-Keras, which is based on Keras, \textbf{Auto-PyTorch} is built on top of PyTorch, a popular deep learning library. Auto-PyTorch automates the process of building and optimizing neural networks and can be used for various tasks such as classification, regression, and time series forecasting.

\subsection{Why Use Auto-PyTorch?}

Auto-PyTorch simplifies the model-building process by handling the following tasks:

\begin{itemize}
    \item Neural architecture search (NAS): Automatically finding the best neural network architecture.
    \item Hyperparameter optimization: Tuning parameters such as the learning rate, number of layers, and batch size.
    \item Data preprocessing: Automatically normalizing and transforming the data.
\end{itemize}

\subsection{How to Use Auto-PyTorch}

Now let’s explore an example of using Auto-PyTorch for tabular classification. Follow the steps below to see how easy it is to build a deep learning model with Auto-PyTorch.

\subsubsection{Install Auto-PyTorch}

First, install the Auto-PyTorch package using pip:

\begin{lstlisting}[style=cmd]
pip install auto-pytorch
\end{lstlisting}

\subsubsection{Load a Dataset}

We’ll use the Iris dataset for this example, a small dataset commonly used for classification tasks.

\begin{lstlisting}[style=python]
from sklearn.datasets import load_iris
from sklearn.model_selection import train_test_split
from autoPyTorch.api.tabular_classification import TabularClassificationTask

# Load the Iris dataset
iris = load_iris()
X, y = iris.data, iris.target

# Split the dataset into training and testing sets
X_train, X_test, y_train, y_test = train_test_split(X, y, test_size=0.2)
\end{lstlisting}

\subsubsection{Create and Train the Model}

Now, you can use Auto-PyTorch to automatically create and train a model. The \texttt{TabularClassificationTask} class helps automate the model-building process for tabular data.

\begin{lstlisting}[style=python]
# Create an Auto-PyTorch classifier
auto_clf = TabularClassificationTask()

# Train the model
auto_clf.search(X_train, y_train, optimize_metric='accuracy', total_walltime_limit=600)
\end{lstlisting}

Here, \texttt{total\_walltime\_limit} sets the time limit (in seconds) for the search process. Auto-PyTorch will try multiple models within this time frame and select the best one.

\subsubsection{Evaluate the Model}

Once the model is trained, you can evaluate its performance on the test data:

\begin{lstlisting}[style=python]
# Evaluate the best model
y_pred = auto_clf.predict(X_test)
accuracy = (y_pred == y_test).mean()
print("Test accuracy:", accuracy)
\end{lstlisting}

Auto-PyTorch automatically selects the best model based on accuracy or other optimization metrics you provide.

\subsection{Understanding Neural Architecture Search (NAS) in Auto-PyTorch}

One of the standout features of Auto-PyTorch is its use of \textbf{Neural Architecture Search (NAS)}. This technique automatically finds the best architecture for your deep learning model by exploring different combinations of layers, activation functions, and other components. Auto-PyTorch uses NAS to ensure that the neural network it builds is well-suited to your specific data.

\section{Conclusion}

In this chapter, we explored how AutoML can be combined with deep learning to simplify the process of building and optimizing models. We introduced two popular AutoML libraries: Auto-Keras and Auto-PyTorch. Both tools automate key tasks such as model selection and hyperparameter tuning, making it easier for beginners to work with deep learning.

Auto-Keras is well-suited for tasks like image and text classification, while Auto-PyTorch offers more flexibility with PyTorch-based models, including tabular data classification and regression. By leveraging AutoML, even those new to deep learning can achieve impressive results without having to dive deep into the complexities of neural network architecture and hyperparameter optimization.

%% file: 20.01_supercomputer.tex
\chapter{Utilizing Remote Devices and Supercomputers for AutoML}

In AutoML, the requirement for vast computational resources is a common challenge. However, if you have access to remote devices such as servers, or even supercomputers, you can utilize these to significantly accelerate your work. In this chapter, we will cover several options for accessing and using these resources, such as Google Colab, SSH-based servers, and supercomputing environments, while ensuring your tasks run efficiently even after disconnection. We will also introduce job scheduling systems like PBS and SLURM. By the end of this chapter, you should be able to set up and maintain remote computational environments, run AutoML tasks, and keep your programs running on powerful machines.

\section{Google Colab: Utilizing Free Resources}

Google Colab~\cite{googlecolab} provides an excellent starting point for those without access to dedicated computational resources. It offers free access to CPU, GPU, and TPU environments for running machine learning models, including AutoML tasks.

\subsection{Using CPU, GPU, and TPU on Google Colab}

To get started, you can easily choose between using a CPU, GPU, or TPU in a Colab notebook. Here's a step-by-step guide to switching the hardware accelerator:

\begin{enumerate}
    \item Open a new notebook in Google Colab.
    \item Click on \textit{Runtime} from the top menu.
    \item Choose \textit{Change runtime type}.
    \item Under the \textit{Hardware accelerator} dropdown, select either \textbf{None} (for CPU), \textbf{GPU}, or \textbf{TPU}.
\end{enumerate}

After setting the hardware, Google Colab will assign you the requested resource. You can check which device is being used by executing the following Python code:

\begin{lstlisting}[style=python]
import torch

# Check if GPU is available
if torch.cuda.is_available():
    device = torch.device("cuda")
    print("Using GPU:", torch.cuda.get_device_name(0))
else:
    device = torch.device("cpu")
    print("Using CPU")
\end{lstlisting}

For TPU usage, additional setup is required:

\begin{lstlisting}[style=python]
import torch_xla.core.xla_model as xm

# Set device to TPU
device = xm.xla_device()
print("Using TPU:", device)
\end{lstlisting}

\textbf{Note:} TPUs in Colab require you to install the \texttt{torch\_xla} library and follow specific usage patterns. You can install this by running the following in a Colab cell:

\begin{lstlisting}[style=cmd]
!pip install torch_xla
\end{lstlisting}

\section{Using SSH to Connect to Remote Servers}

For users with access to more powerful remote machines, such as dedicated servers, SSH is commonly used to connect and run machine learning jobs. However, a common issue arises: if you close your terminal or disconnect from the server, your running program is terminated. To solve this, we will use tools like \texttt{screen} or \texttt{tmux}, which allow you to keep your session running in the background even if the SSH connection is lost.

\subsection{SSH: Basic Commands}

To connect to a remote server using SSH, open a terminal and type the following:

\begin{lstlisting}[style=cmd]
ssh username@remote-server-ip
\end{lstlisting}

You will be prompted for your password, after which you'll have access to the remote machine.

\subsection{Keeping Processes Running After Disconnection with Screen}

Once connected to the server, you can install \texttt{screen} (if it's not already available) by running:

\begin{lstlisting}[style=cmd]
sudo apt-get install screen
\end{lstlisting}

To start a new screen session, use the command:

\begin{lstlisting}[style=cmd]
screen -S my_session_name
\end{lstlisting}

You can now run your Python code inside this screen session:

\begin{lstlisting}[style=cmd]
python my_automl_script.py
\end{lstlisting}

To detach from the session without closing it, press:

\begin{lstlisting}[style=cmd]
Ctrl + A, then D
\end{lstlisting}

Your program will continue to run in the background. To reattach to this session later, type:

\begin{lstlisting}[style=cmd]
screen -r my_session_name
\end{lstlisting}

\section{Using Supercomputers for AutoML}

Supercomputers are often equipped with advanced hardware (like GPUs) and require job scheduling systems to manage workloads. Two of the most common systems are PBS (Portable Batch System) and SLURM (Simple Linux Utility for Resource Management).

\subsection{Using PBS to Schedule Jobs}

PBS is a popular job scheduling system used in high-performance computing environments. To use PBS, you create a job script that specifies the resources you need, then submit it to the queue.

Here is a simple example of a PBS job script:

\begin{lstlisting}[style=cmd]
#!/bin/bash
#PBS -N automl_job
#PBS -l nodes=1:ppn=8
#PBS -l walltime=4:00:00
#PBS -j oe

# Load necessary modules
module load python/3.8
module load pytorch/1.9

# Navigate to the working directory
cd $PBS_O_WORKDIR

# Run the Python script
python my_automl_script.py
\end{lstlisting}

Submit the job to the PBS queue by running:

\begin{lstlisting}[style=cmd]
qsub automl_job.pbs
\end{lstlisting}

You can monitor the status of your job with:

\begin{lstlisting}[style=cmd]
qstat
\end{lstlisting}

\subsection{Using SLURM to Schedule Jobs}

SLURM~\cite{yoo2003slurm} is another widely used workload manager for large compute clusters. Similar to PBS, you write a job script and submit it to the queue.

Here is an example of a SLURM job script:

\begin{lstlisting}[style=cmd]
#!/bin/bash
#SBATCH --job-name=automl_job
#SBATCH --nodes=1
#SBATCH --ntasks-per-node=8
#SBATCH --time=04:00:00
#SBATCH --output=output_%j.txt

# Load necessary modules
module load python/3.8
module load pytorch/1.9

# Navigate to the working directory
cd $SLURM_SUBMIT_DIR

# Run the Python script
python my_automl_script.py
\end{lstlisting}

To submit the job to the SLURM queue, use the command:

\begin{lstlisting}[style=cmd]
sbatch automl_job.slurm
\end{lstlisting}

To check the status of your job, use:

\begin{lstlisting}[style=cmd]
squeue
\end{lstlisting}

\section{Conclusion}

By using resources like Google Colab, remote servers via SSH, and supercomputers through PBS or SLURM, you can access the computational power needed to effectively run AutoML tasks. Whether you are utilizing free cloud-based resources or dedicated hardware, understanding these tools will enable you to optimize your workflow and keep your programs running, even in remote and large-scale environments.

%% file: 21_future.tex
\part{Conclusion and Future Outlook}

\chapter{Future Development of Automated Machine Learning}

\section{Challenges and Opportunities in AutoML}

Automated Machine Learning (AutoML) has revolutionized the process of machine learning (ML) by automating repetitive tasks such as data preprocessing, model selection, and hyperparameter tuning. However, despite these advances, there are several challenges and opportunities that AutoML will face in the future.

\subsection{Challenges in AutoML}

\subsubsection{1. Scalability:}
One of the major challenges in AutoML is its scalability. As the size of datasets continues to grow exponentially, AutoML frameworks need to scale efficiently to handle massive datasets. For instance, an AutoML system trained on a small dataset may be much faster, but when applied to a dataset with millions of rows and hundreds of features, it might struggle due to time and resource constraints.

\subsubsection{Example:}
In a typical situation where a beginner is handling smaller datasets such as the Iris dataset (with 150 samples and 4 features), AutoML frameworks perform remarkably well. But consider the scenario of working with a massive dataset such as a financial dataset containing millions of records. The efficiency and speed of AutoML systems can drop significantly without proper resource management.

\begin{lstlisting}[style=python]
# Example using PyTorch and AutoML for a small dataset
import torch
from torch import nn, optim
from sklearn.datasets import load_iris
from sklearn.model_selection import train_test_split

# Load and split the dataset
iris = load_iris()
X_train, X_test, y_train, y_test = train_test_split(iris.data, iris.target, test_size=0.2)

# Simple feedforward network for classification
class SimpleNN(nn.Module):
    def __init__(self):
        super(SimpleNN, self).__init__()
        self.fc1 = nn.Linear(4, 10)
        self.fc2 = nn.Linear(10, 3)

    def forward(self, x):
        x = torch.relu(self.fc1(x))
        x = self.fc2(x)
        return x

# Training loop (pseudo-automated)
model = SimpleNN()
optimizer = optim.Adam(model.parameters(), lr=0.001)
criterion = nn.CrossEntropyLoss()

# In real AutoML systems, these loops and configurations would be auto-tuned
for epoch in range(50):
    optimizer.zero_grad()
    output = model(torch.FloatTensor(X_train))
    loss = criterion(output, torch.LongTensor(y_train))
    loss.backward()
    optimizer.step()
\end{lstlisting}

\subsubsection{2. Interpretability:}
Another significant challenge in AutoML is the interpretability of the models it generates. Automated systems often produce highly complex models, such as deep neural networks, which can be difficult to interpret. For example, a model that performs well on predicting loan defaults in the banking sector might be highly accurate, but the reasoning behind its predictions may be unclear to the stakeholders. This lack of interpretability can be a barrier, especially in regulated industries like healthcare and finance.

\subsubsection{3. Domain-specific Adaptations:}
AutoML systems need to be tailored for specific domains. Currently, many AutoML frameworks are designed with a general approach, which may not be suitable for domain-specific problems. For example, in fields like biology or chemistry, domain knowledge is essential to create meaningful features and models. Future AutoML systems need to incorporate more sophisticated techniques to integrate domain expertise effectively.

\subsection{Opportunities in AutoML}

\subsubsection{1. Democratization of AI:}
One of the biggest opportunities for AutoML is the democratization of AI and ML technologies. By lowering the technical barriers to entry, AutoML opens the doors for non-experts, including students, business analysts, and others, to harness the power of ML. For example, a healthcare professional with little programming knowledge can use AutoML tools to build predictive models to analyze patient data.

\subsubsection{2. Enhanced Optimization Techniques:}
There is a vast opportunity to develop better optimization techniques, such as more efficient hyperparameter tuning algorithms, improved search spaces, and enhanced neural architecture search (NAS). These developments can drastically improve the performance of AutoML systems.

\subsubsection{3. Integration with Edge Computing:}
With the rise of IoT and edge computing, there is a growing demand for deploying ML models on devices with limited resources. AutoML frameworks need to evolve to include lightweight models that can be efficiently deployed on edge devices. This presents a huge opportunity for growth, especially in real-time applications like autonomous vehicles or smart wearables.

\section{AutoML Applications in Different Fields}

AutoML has already demonstrated its potential in a variety of industries. Let's explore how it is applied in fields such as finance, healthcare, and more.

\subsection{Finance}
In the finance industry, AutoML is increasingly being used to automate trading strategies, credit scoring, and fraud detection. Traditional methods for credit scoring involved manual feature engineering and statistical models. With AutoML, the process is streamlined, allowing for the automatic generation of features and model selection. Moreover, in fraud detection, AutoML can be used to detect anomalies in transaction data by automatically selecting the best models for anomaly detection.

\begin{lstlisting}[style=python]
# Example: AutoML application in finance for fraud detection
# Dataset: Assume we have a dataset of credit card transactions
import torch
import numpy as np

# Sample data
transaction_data = np.random.rand(1000, 10)  # 1000 transactions, 10 features
labels = np.random.randint(0, 2, 1000)  # Fraud (1) or not (0)

# Simple PyTorch model
class FraudDetectionNN(nn.Module):
    def __init__(self):
        super(FraudDetectionNN, self).__init__()
        self.fc1 = nn.Linear(10, 50)
        self.fc2 = nn.Linear(50, 1)

    def forward(self, x):
        x = torch.relu(self.fc1(x))
        x = torch.sigmoid(self.fc2(x))
        return x

# Example of automated training process
model = FraudDetectionNN()
optimizer = optim.Adam(model.parameters(), lr=0.001)
criterion = nn.BCELoss()

for epoch in range(100):
    optimizer.zero_grad()
    output = model(torch.FloatTensor(transaction_data))
    loss = criterion(output.squeeze(), torch.FloatTensor(labels))
    loss.backward()
    optimizer.step()
\end{lstlisting}

\subsection{Healthcare}
In healthcare, AutoML is transforming fields like diagnostic imaging, personalized medicine, and hospital management. For example, AutoML systems can assist in identifying diseases from medical images or predict patient outcomes based on historical data. AutoML can automate the process of selecting the best algorithms for image classification, anomaly detection, and more.

\section{Trends and Future Outlook}

\subsection{1. Neural Architecture Search (NAS):}
One of the most promising trends in AutoML is the development of Neural Architecture Search (NAS) algorithms. These algorithms automatically search for the best neural network architectures for a given task, eliminating the need for manual architecture design. With more advanced NAS techniques, future AutoML systems will be able to design complex models for tasks such as image recognition and natural language processing (NLP).

\subsection{2. Model Compression and Deployment:}
As the demand for deploying models on mobile and edge devices increases, there is a growing trend towards model compression techniques like pruning and quantization. AutoML systems will increasingly focus on optimizing models not just for accuracy but for deployment efficiency, ensuring that models are lightweight and can run on resource-constrained devices.

\subsection{3. Explainable AI (XAI):}
Another key trend is the integration of Explainable AI into AutoML workflows. In the future, AutoML systems will not only produce highly accurate models but also generate explanations that help stakeholders understand why a particular decision was made~\cite{drozdal2020trust}. This will be particularly important in industries like finance and healthcare where trust and transparency are essential~\cite{arrieta2020explainable}.

\subsection{4. Ethics and Fairness:}
As AutoML becomes more widespread, there will be a greater focus on ensuring that the models produced are ethical and fair~\cite{hazan2020autml}. There is already growing concern about bias in machine learning models, and AutoML will need to incorporate fairness constraints to ensure that the models it generates do not reinforce biases present in the data~\cite{mehrabi2021survey}.

\subsection{Conclusion:}
In conclusion, the future of AutoML is incredibly promising, with advancements in scalability, domain-specific adaptations, and interpretability. While there are challenges to overcome, the opportunities for democratization of AI, enhanced optimization techniques, and applications across diverse fields are immense. As AutoML continues to evolve, it will further revolutionize industries like finance, healthcare, and beyond.